\documentclass[conference]{IEEEtran}
\IEEEoverridecommandlockouts

\usepackage{cite}
\usepackage{amsmath,amssymb,amsfonts}
\usepackage{algorithmic}
\usepackage{graphicx}
\usepackage{textcomp}
\usepackage{xcolor}
\def\BibTeX{{\rm B\kern-.05em{\sc i\kern-.025em b}\kern-.08em
    T\kern-.1667em\lower.7ex\hbox{E}\kern-.125emX}}
\usepackage{mathtools}
\usepackage{cite}
\usepackage[normalem]{ulem}
\usepackage{courier}
\usepackage{fixltx2e}
\usepackage{helvet}
\usepackage{tabularx}
\usepackage{framed}
\usepackage{dsfont}
\usepackage{times}
\usepackage{url}
\usepackage{caption}
\DeclareCaptionFont{6pt}{\fontsize{6pt}{8pt}\selectfont}
\usepackage[shortlabels]{enumitem}
\setlist[enumerate, 1]{1\textsuperscript{o}}

\usepackage{verbatim}
\usepackage[linesnumbered,ruled,vlined,onelanguage]{algorithm2e}
\newcommand{\vars}{\texttt}
\newcommand{\func}{\textsl}

\usepackage{multirow}
\usepackage{subfig} 
\usepackage{booktabs} 
\usepackage{color}
\usepackage{float}
\usepackage{graphicx}


\newcommand{\bx}{\mathbf{x}}
\newcommand{\by}{\mathbf{y}}

\DeclareMathOperator{\sign}{sign}

\usepackage{mathtools}

\usepackage{csquotes}

\setlist[itemize]{leftmargin=*}
\setlist[enumerate]{leftmargin=*}

\newcommand{\FGSM}{\mbox{\sc FGSM}}
\newcommand{\BIM}{\mbox{\sc BIM}}
\newcommand{\JSMA}{\mbox{\sc JSMA}}
\newcommand{\CW}{\mbox{\sc CW}}
\newcommand{\PGD}{\mbox{\sc PGD}}
\newcommand{\DF}{\mbox{\sc DeepFool}}
\newcommand{\OP}{\mbox{\sc One-Pixel}}
\newcommand{\MIM}{\mbox{\sc MIM}}

\def \sign {\mathrm{sign}}
\def \clip {\mathrm{clip}}
\DeclareMathOperator*{\argmin}{argmin}
\DeclareMathOperator*{\argmax}{argmax}

\usepackage{lipsum}

\usepackage{xspace}
\newcommand{\ourframework}{\mbox{\sc Athena}\xspace}

\newcommand{\ying}[1]{{\color{teal} Ying: #1}}

\renewcommand\fbox{\fcolorbox{blue}{blue!20}}

\usepackage{hyperref}
\hypersetup{
	colorlinks   = true,
	citecolor    = blue,
	linkcolor    = blue,
}

\begin{document}

\title{\Large \bf \ourframework: A Framework based on Diverse Weak Defenses \\ for Building Adversarial Defense}


\author{
\IEEEauthorblockN{Ying Meng, Jianhai Su, Jason M. O'Kane, Pooyan Jamshidi}
\IEEEauthorblockA{\textit{Departmetn of Computer Science and Engineering} \\
\textit{University of South Carolina}\\
Columbia, SC, USA}
}

\maketitle

\pagestyle{plain}


\begin{abstract}

There has been extensive research on developing defense techniques against adversarial attacks; however, they have been mainly designed for specific model families or application domains, therefore, they cannot be easily extended. Based on the design philosophy of ensemble of diverse weak defenses, we propose \ourframework---a flexible and extensible framework for building generic yet effective defenses against adversarial attacks.
We have conducted a comprehensive empirical study to evaluate several realizations of \ourframework with four threat models including zero-knowledge, black-box, gray-box, and white-box. We also explain (i) why diversity matters, (ii) the generality of the defense framework, and (iii) the overhead costs incurred by \ourframework.

\end{abstract}



\section{Introduction}\label{sec:intro}

To date, machine learning systems continue to be highly susceptible to \emph{adversarial examples} (also known as \emph{wild patterns}~\cite{biggio2018wild}), notwithstanding their attainment of state-of-the-art performance across a wide variety of domains, including speech recognition~\cite{SpeechRecog:Xiong2016a}, object detection~\cite{He_2016_CVPR}, and image classification~\cite{He_2016_CVPR}. Adversarial examples (AEs) are typically crafted by adding small, human-imperceptible perturbations to benign samples in order to induce machine learning systems to make erroneous predictions~\cite{szegedy2013intriguing, goodfellow2014explaining, carlini2017evaluaterobustness, Eykholt2018AttackPhyWorld, finlayson2019adversarial, heaven2019deep}.

There has been extensive research on developing defense techniques against adversarial attacks; however, they have been mainly designed for specific model families, therefore, they cannot be easily extended to new domains. In addition, the current defense techniques in almost all cases assume specific known attack(s) and are tested in weak threat models~\cite{tramer2020adaptive}. As a result, others have shown that these defenses can be circumvented under slightly different conditions, either a stronger adaptive adversary or in some cases even weaker (but different) adversaries~\cite{carlini2017magnet,carlini2019evaluating,tramer2020adaptive,he2017adversarial}.  The ``arms race'' between the attacks and defenses leads us to pose this central question: 

\begin{center}
\fbox{\parbox{0.97\columnwidth}{
How can we, instead, design a defense, not as a technique, but as a framework that one can construct a specific defense considering the niche tradeoff space of robustness one may want to achieve as well as the cost one is willing to pay to achieve that level of robustness?}}
\end{center}

\begin{figure}[t]\captionsetup[subfigure]{font=scriptsize,labelfont=scriptsize}
    \scriptsize
    \centering
    \tiny
    \subfloat[MNIST]{
        \includegraphics[width=\linewidth]{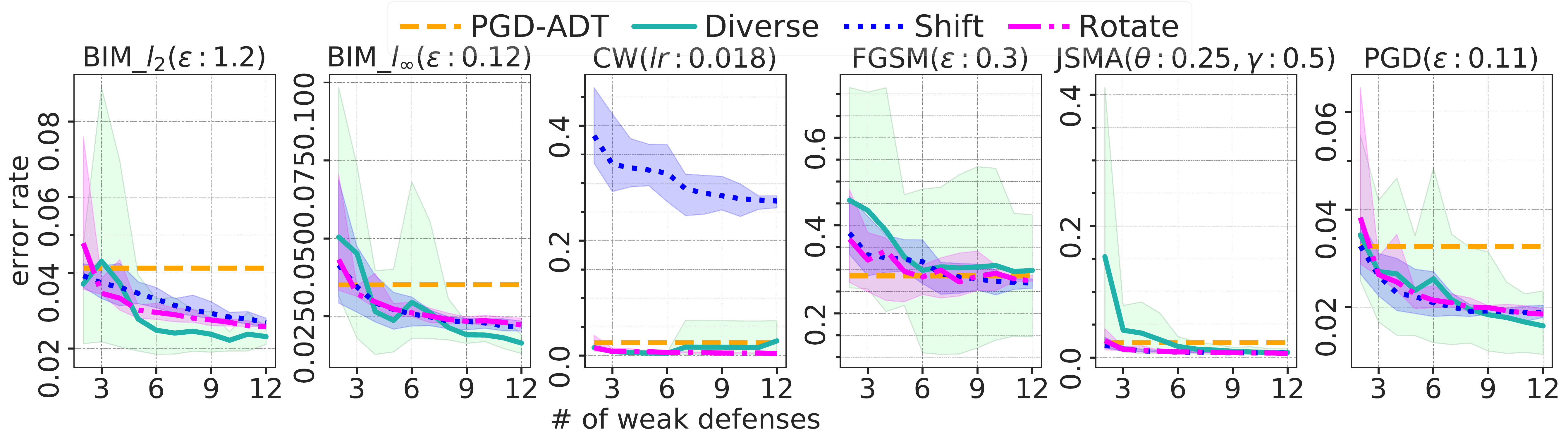}
        \label{subfig:motivation_mnist}
    }
    \\
    \subfloat[CIFAR-100]{
        \includegraphics[width=\linewidth]{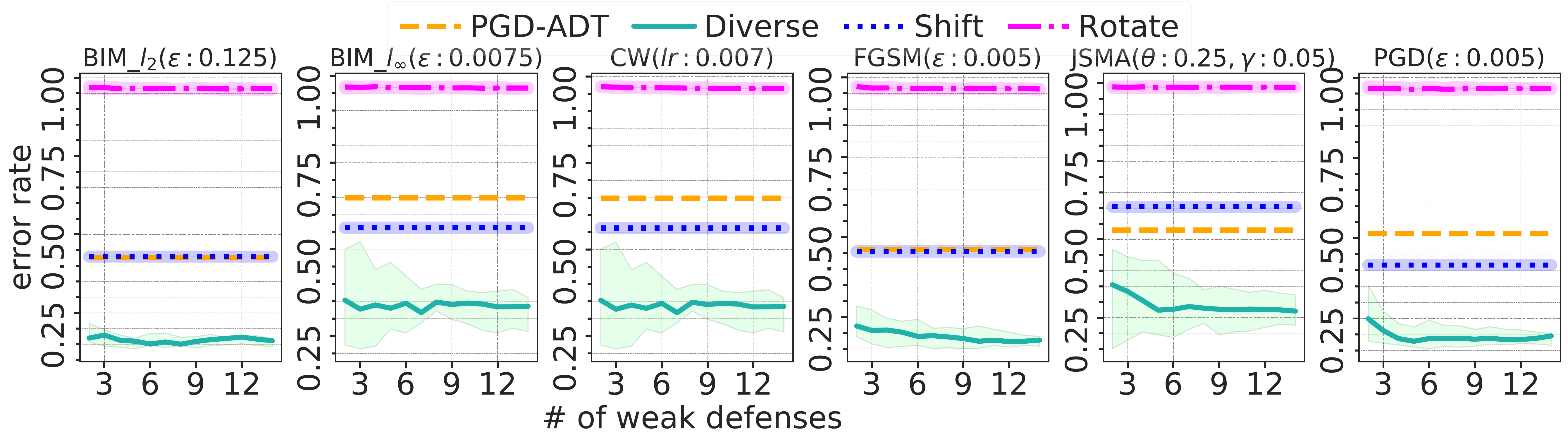}
        \label{subfig:motivation_cifar}
    }
    \caption{A diverse ensemble defense can be robust against adversarial attacks: The error rate decrease as the diversity and the number of weak defenses increases in the ensemble defense.}
    \label{fig:athena_motivation}
\end{figure}

To address this question, we propose \ourframework\footnote{Goddess of defense in Greek mythology.}---an \emph{extensible framework} for building \emph{generic} (and thus, broadly applicable) yet effective defenses against adversarial attacks.
The design philosophy behind \ourframework~is based on ensemble of many \emph{diverse weak defenses} (WDs), where each WD, the building blocks of the framework, is a machine learning classifier (e.g., DNN, SVM) that first applies a transformation\footnote{We, therefore, use \emph{weak defense} and \emph{transformation} interchangeably.} on the original input and then produces an output for the transformed input. Given an input, an ensemble first collects predicted outputs from all of the WDs and then determines the final output, using some \emph{ensemble strategy} such as majority voting or averaging the predicted outputs from the WDs. 
Note that input transformations have been extensively used for defending machine learning systems from adversarial attacks~\cite{grosse2017statistical,metzen2017detecting,das2017keeping,lu2017no,bhagoji2018enhancing,dziugaite2016study,guo2017transformation, Agarwal2020_ImgTrans}, where a singular transformation such as denoise or rotation was used. In this work, rather than looking for a peculiar transformation that effectively mitigates adversarial perturbations, we propose a framework for building an ensemble defense on top of a large population of diverse transformations of various types. 
We show that ensembling diverse transformations can result in a robust defense against a variety of attacks and provide a \emph{tradeoff space}, where one can add more transformations to the ensemble to achieve a \emph{higher robustness} or decrease the weak defenses to achieve \emph{lower overhead and cost}.

To motivate the idea, we first evaluate how the number of WDs affects the ensemble's robustness. We used a variety of random WDs and the output averaging strategy as the ensemble strategy, which determines the label by averaging the outputs from WDs, then tested the ensembles against various attacks. To show that an ensemble is more robust by utilizing a diverse set of transformations, we compared three versions of ensemble-based defenses: (i) ensembles that were created from a diverse set of transformations (referred to as ``Diverse''), and two non-diverse homogeneous ensembles that were created based on a variations of two particular transformations, i.e., (ii) translations (referred to as ``Shift''), and (iii) rotations (referred to as ``Rotate''). Average error rates and variations of 5 trials of the three versions of the ensembles are presented in Figure~\ref{fig:athena_motivation}. 

By disturbing its input, a transformation changes the adversarial optimized perturbations and therefore making the perturbations less effective. By changing the input in different ways, a group of transformations complement each other in blocking adversarial perturbations introduced by different attacks. Thus, with more WDs, an ensemble is likely to be more robust against adversarial attacks. As presented in Figure~\ref{fig:athena_motivation}, for most cases, the error rate drops as an ensemble utilizes more and more WDs. When constructed using 4 or more WDs, all ensembles achieved competitive performance or even outperformed PGD adversarial training (PGD-ADT) \cite{madry2018pgd}, a state-of-the-art defense approach.


The effectiveness of homogeneous ensembles with a single transformation type varies on attacks and datasets. Shifts and rotations perform well in filtering all adversarial perturbations except CW in MNIST. However, the ensembles based on rotations are not effective on CIFAR-100. Compared to rotations, ensembles based on shift are more effective in filtering CW perturbations on CIFAR-100, but less effective on MNIST. A collection of diverse transformations, however, mitigates the perturbations in different ways, adjusting angles or positions, denoising, compressing, and so on. When some transformations may fail to disentangle perturbations by a particular adversary, others may still mitigate the perturbations. We can observe that ``Diverse'' achieves the lowest error rate in most cases, especially for tasks on CIFAR-100, where ``Diverse'' outperforms all, achieving an error rate that is at least 20\% lower than the second-best performers. Providing more freedom in the solution space by varying the transformation types and parameters, the diverse ensembles give broader variations than the homogeneous ensembles.  This suggests the insight that ``diversity matters''. In addition, the results indicate that we could have found a much more robust ensemble from the diverse library if we have designed and selected the WDs very carefully. Besides, by adding more WDs to the ensemble, the error rate of the worst ensemble decreases, showing potentials for worst case scenarios in dynamic environments where the behavior of the attacker may change.

The observations reflected by our results suggest that it is possible to construct an ensemble with many diverse WDs, whereby the model becomes increasingly more robust to many forms of adversarial attacks. Our objective in this work is to construct an efficient defense without any assumption regarding the adversarial attack. That is, building a defense that, in general, is robust against any adversary. 



Overall, we make the following contributions:
\begin{itemize}
    \itemsep0em
    \item We studied the defensive effectiveness of 72 transformations for MNIST and 22 for CIFAR-100. To the best of our knowledge, this is the first work studying such a large variety of transformations. 
    \item We proposed \ourframework as a framework for building ensemble defenses that is flexible to scale by simply adding/removing WDs and to change the ensemble strategy.
    \item We evaluated \ourframework~via 4 threat models for image classification on MNIST and CIFAR-100:
        \begin{itemize}
            \itemsep0em
            \item \emph{Zero-knowledge}: For MNIST, we realized and evaluated 5 ensembles against 9 attacks (with 5 variations of strengths). We also compared our results with 2 adversarial defenses from previous studies---PGD adversarial training and randomized smoothing. Across all attacks, at least one of the 5 ensembles achieved the lowest error rate. The improvements, compared to the undefended model (UM), ranged from $8.03\%$ to $96.17\%$. We performed the same experiments on CIFAR-100, but due to the computational cost, in a smaller scale, where 4 ensembles were tested against 6 attacks. In most cases, \ourframework~outperformed the previous adversarial defenses. The improvements ranged from $19.61\%$ to $90.29\%$ against other defenses (Section~\ref{subsec:eval_zk}).
            \item \emph{Black-box}: We evaluated with transfer-based and query-based black-box attacks on CIFAR-100, and observed the effectiveness of \ourframework~in protecting the UM under both attacks. Against the transfer-based attack, \ourframework~reduces the transferability rate of generated AEs by $25.82\%$ on average. When facing the query-based attack, \ourframework~significantly restricts the attacker's capability to minimize the distance between generated AEs and benign samples. Given a query budget, the averaged distance between \ourframework-targeted AEs and benign samples is 0.35x to 15.45x larger than that between the UM-targeted AEs and benign samples (Section~\ref{subsec:eval_bb}).
            \item \emph{Gray-box and White-box}: We conducted an evaluation of gray-box and white-box attacks on an ensemble based on majority voting among predicted labels of WDs. We generated AEs using \FGSM~attack. 10 variants were crafted using different constraints on the maximum dissimilarity of AEs. The error rate of \ourframework~is $10\%$ to $20\%$ lower than that of the UM except the weakest set where our approach is equally ineffective as the UM. However, it is computationally costly to craft effective AEs (5x more time to craft an AE on average), and the generated AEs are much more prone to be detected (Section~\ref{subsec:eval_wb_gb}). 
            
        \end{itemize}
    \item We also performed a comprehensive empirical study to understand \emph{why ensembles of many WDs work} and \emph{why the diversity of WDs matters}. 
    \item We released the source code and experimental data: \\ \url{https://github.com/softsys4ai/athena}
\end{itemize}


\section{Background and Definitions}\label{sec:background}

\subsection{Notation}
In this paper, we refer to the legitimate data set as $\mathcal{D}\subset \mathds{R}^d$; a transformation operation as $t_i ~(i = 1, 2, 3, \dots)$; and $t_i(\bx)$ as the output from applying transformation $t_i$ on the input $\bx$. The composition of $n$ transformations is also a transformation: $t_g = t_n \circ t_{n-1} \circ \dots \circ t_1$. Therefore, we use $t_i$ to denote a transformation or a composition of transformations. $\mathcal{D}_{t_i} = \{\bx_{t_i} = t_i(\bx) | \bx \in \mathcal{D}\}$ denotes the set of transformed examples of $\mathcal{D}$ associated to $t_i$. That is, $\mathcal{D}_{t_i} = t_i(\mathcal{D})$.

We focus our work on \emph{supervised machine learning}, where a classifier is trained on labeled data in $\mathcal{D}$. Here, a classifier is 
a function $f(\cdot)$ that takes a data point $\bx \in \mathds{R}^d$ as input and produces a vector $\by$ of probabilities of all the classes in $\mathcal{C}$. 
Given a target model $f(\cdot)$ and an input $\bx$ with ground truth $y_{\text{true}}$, an adversary attempts to produce an AE $\bx'$ (limited in $l_p$-norms), such that $\argmax(f(\bx')) \neq y_{\text{true}}$, by solving an optimization to maximize the loss function:
\iftrue
\begin{equation}\label{attack}
    \begin{split}
        \max_{\delta}(\mathcal{L}(f, \bx + \delta, y_{\text{true}})), \\
        \text{s.t.}~||\delta||_p \leq \epsilon ~ \text{and}~ \bx' = \bx + \delta \in [0, 1]^D,
    \end{split}
\end{equation}
\fi
where $\argmax(f(\bx))$ is the predicted output given an input $\bx$, $\mathcal{L}$ is the loss function, $\delta$ is the perturbation, and $\epsilon$ is the magnitude of the perturbation. To remain undetected, the adversarial example $\bx'$ should be as \emph{similar} to the \emph{benign sample} (BS) as possible; therefore, attacks (e.g., \FGSM~\cite{goodfellow2014explaining}, \PGD~\cite{madry2018pgd}, and \CW~\cite{carlini2017evaluaterobustness}) use different norms (such as $l_0$, $l_2$, or $l_{\infty}$) to constrain the distance between $\bx$ and $\bx'$. 

When crafting an adversarial example for an input, some attackers force the target model to produce a specific output label, $t\in \mathcal{C}$, for a given input, while other attackers only seek to produce an output label that does not equal the ground truth. The former is referred to as a \textit{targeted}, while the latter is referred to as an \textit{untargeted} or \textit{non-targeted attack}.

Given a set of $N$ input data $\{\bx_1,\ldots,\bx_N\}$ and a target classifier $f(\cdot)$, an \emph{adversarial attack} aims to generate $\{\bx_1',\ldots,\bx_N'\}$, such that each $\bx_n'$ is an adversarial example for $\bx_n$. The \emph{success rate} of an attack is measured by the proportion of predictions that were successfully altered by an attack: $\frac{1}{N} \sum_{n=1}^N \mathds{1}[f(\bx_n) \neq f(\bx_n')]$. The magnitude of the perturbations performed by the attack, in this work, is measured by \emph{normalized $l_2$-dissimilarity}, $\frac{1}{N} \sum_{n=1}^N \frac{\| \bx_n - \bx_n' \|_2}{\| \bx_n \|_2}$. A strong adversarial attack has a high success rate while its normalized $l_2$ dissimilarity is low.


A \emph{defense} is a method that aims to make the prediction on an adversarial example equal to the prediction on the corresponding benign sample, i.e., $\argmax(f(\bx')) = y_{\text{true}}$. In this work, our defense mechanism is based on the idea of \emph{many diverse weak defenses}, which are essentially \emph{transformation-based defenses}, i.e., they produce output label via $f_{t_i}(t_i(\bx'))$. Typically, $t(\cdot)$ is a complex and non-differentiable function, 
which makes it difficult for an adversary to attack the machine learning model $f(t(\bx))$, even when the attacker knows both $f(\cdot)$ and $t(\cdot)$.

A model that was trained using data set $\mathcal{D}_{t_i} (i = 1, 2, \dots)$ is referred to as a \emph{weak defense} and denoted as $f_{t_i}$. Each of these trained models are collectively used to construct defense ensembles. On the other side, the original model $f$ that is trained on $\mathcal{D}$ is referred to as the \emph{undefended model}.

\subsection{Adversarial Attack}

\textbf{Gradient-based Attacks} perturb their input with the gradient of the loss with respect to the input.
Some attacks in this family perturb the input only in one iteration. For example, \FGSM~processes an adversarial example as follows:
\begin{equation}\label{equ_fgsm}
    \begin{split}
        \bx' = \bx + \epsilon \cdot \sign(\bigtriangledown_{\bx}J(\bx, \by)),
    \end{split}
\end{equation}
where $J$ is the cost function of the target model $f$, $\bigtriangledown_{\bx}$ is the gradient with respect to the input $\bx$ with corresponding true output $\by$, and $\epsilon$ is the magnitude of the perturbation.

Other variations, like \BIM~\cite{kurakin2017physical}, \MIM~\cite{dong2017MIM}, and \PGD, are iterative and gradually increase the magnitude until the input is misclassified. For example, \BIM, an extension of \FGSM, rather than taking one big jump $\epsilon$, takes multiple smaller steps $\alpha < \epsilon$ with the result clipped by $\epsilon$. Specifically, \BIM~begins with $\bx'_{0} = \bx$, and at each iteration it performs: 
\begin{equation}\label{equ_bim}
    \begin{split}
        \bx'_{i} = \clip_{\bx, \epsilon}\{\bx'_{i-1} - (\alpha \cdot \sign(\bigtriangledown_{\bx}J(\bx'_{i-1}, \by))\},
    \end{split}
\end{equation}
where $\text{clip}_{\bx, \epsilon}(\textbf{A})$ denotes the element-wise clipping of $\bx$; the range of $\textbf{A}$ after clipping will be $[\bx-\epsilon, \bx+\epsilon]$. 

\JSMA~\cite{papernot2016limitations}, another gradient-based approach, greedily finds the most sensitive direction, such that changing its values will significantly increase the likelihood of a target model labeling the input as the target class:
\begin{equation}\label{equ_jsma}
    \begin{split}
        s_t = \frac{\delta t}{\delta x_i}; s_o = \Sigma_{j \neq t}\frac{\delta j}{\delta x_i}; \\
        s(x_i) = s_t |s_o| \cdot (s_t < 0) \cdot (s_o > 0),
    \end{split}
\end{equation}
where $s_t$ represents the Jacobian of target class $t\in \mathcal{C}$ with respect to the input image, and where $s_o$ is the sum of Jacobian values of all non-target classes.

\CW~constraints adversarial samples to stay within a certain distance from the benign sample. \DF~\cite{dezfooli2015deepfool} takes iterative steps to the direction of the gradient of a linear approximation of the target model. Both attacks generate adversarial examples by solving an optimization problem of the form:
\begin{equation}\label{equ_opt}
    \begin{split}
        \argmin(d(s, x + \delta) + c \cdot \mathcal{L}(x + \delta)),
    \end{split}
\end{equation}
where $\mathcal{L}$ is the loss function for solving $f(x + \delta) = t$ and $t\in \mathcal{C}$ is the target label.

\textbf{Black-box Attacks} do not require any knowledge of target models, yet are shown to be effective in fooling machine learning models. \OP~\cite{su2017onepixel}, one of the extreme adversarial attack methods, generates adversarial examples using an evolutionary algorithm called Differential Evolution~\cite{Storn1997DE}.

\section{Athena: A Framework of Many WDs}\label{sec:approach_overview}
In this section, we present our defense framework, called \ourframework, which is based on \emph{ensembles} of \emph{many diverse weak defenses}. With \ourframework~one can construct effective defenses that are resilient to adversarial attacks. Further, such defenses are not tied to a particular ML classifier and can be deployed in different domains beyond image classification. 

\begin{figure}[t]
    \captionsetup[subfigure]{font=scriptsize,labelfont=scriptsize}
    \footnotesize
    \centering
    \subfloat[][Training a weak defense]{
        \includegraphics[width=0.85\linewidth]{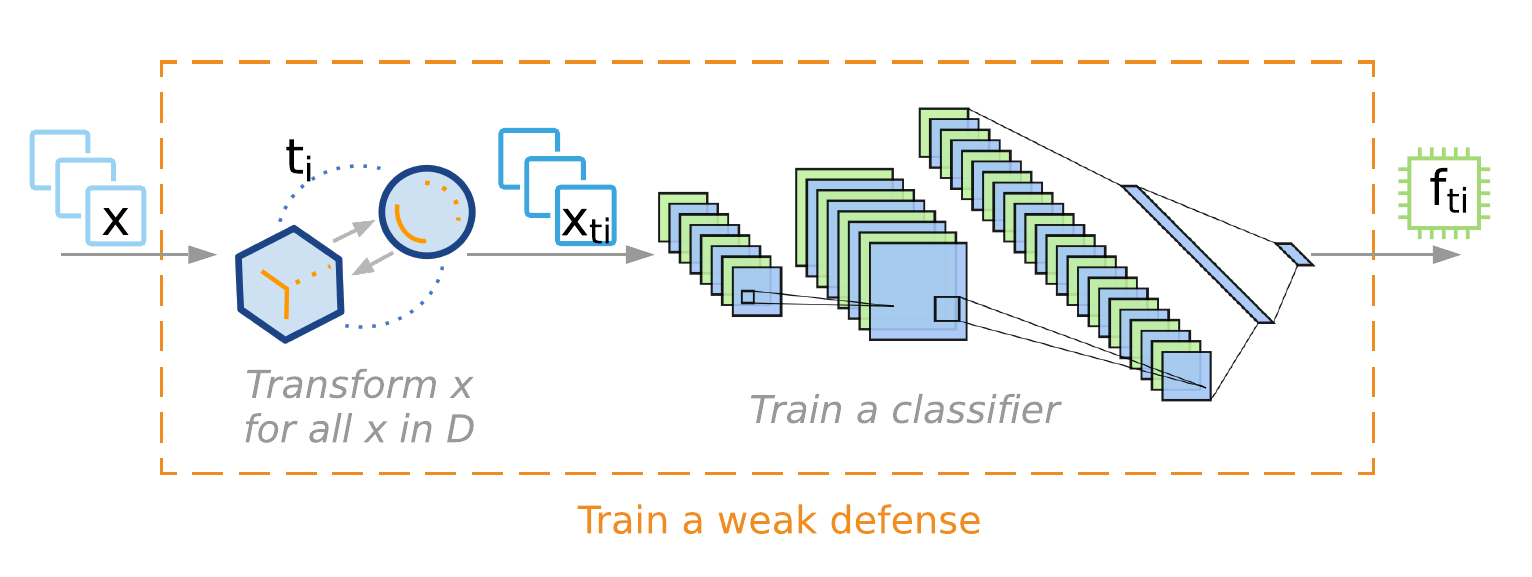}\label{subfig:dia_training_phase}
    }
    \\
    \subfloat[][Producing output for a given input]{
        \includegraphics[width=0.8\linewidth]{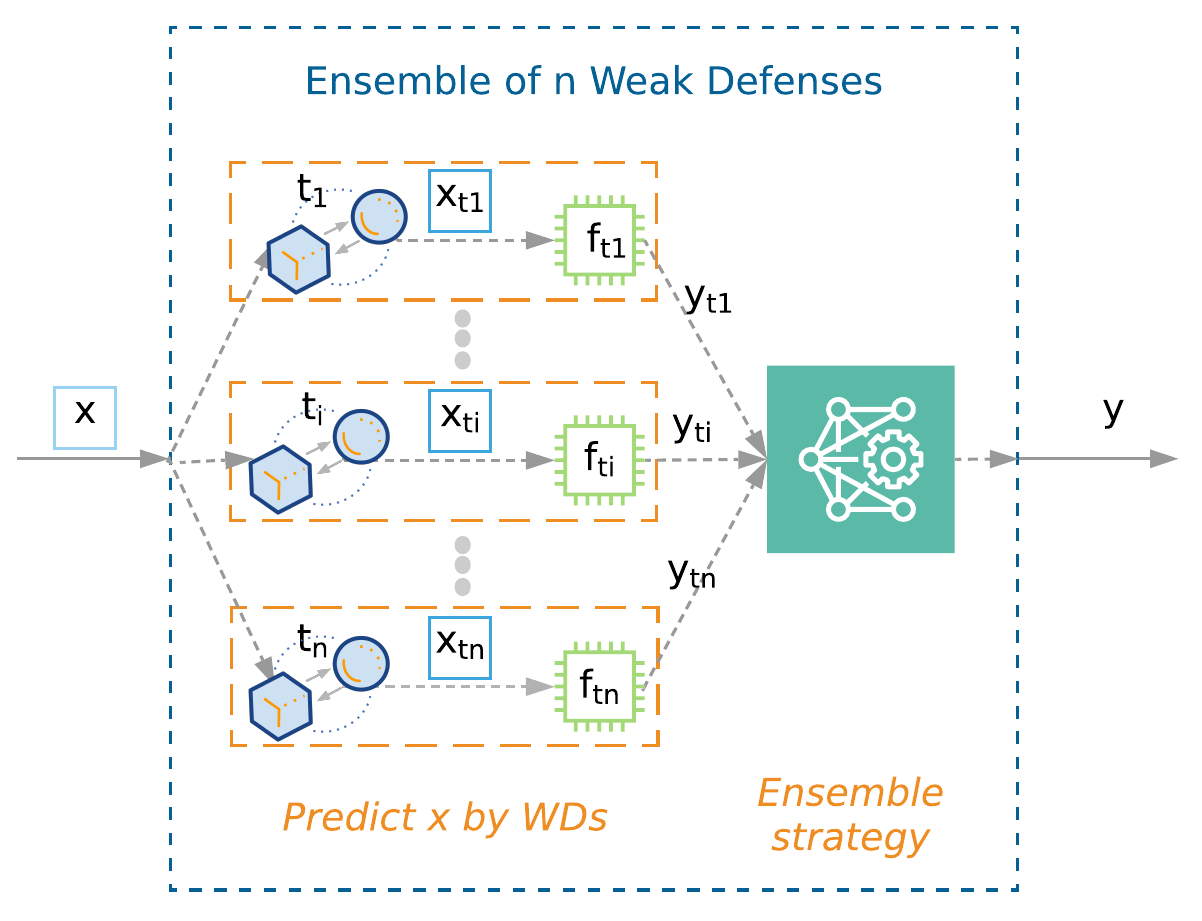}\label{subfig:dia_testing_phase}
    }
    \caption{\ourframework, a framework for building adversarial defense: (1) Training a weak defense on transformed inputs. (2) Predicting the output label using predictions of all weak defenses.}
    \label{fig:dia_ensemble_of_wds} 
\end{figure}

\ourframework~works as follows: At test time, for a given input data $\bx$, (i) it first collects outputs from all WDs, and then (ii) it uses an ensemble strategy (e.g., majority voting) to compute the final output (Figure~\ref{fig:dia_ensemble_of_wds}~\subref{subfig:dia_testing_phase}). Each WD can be any ML model (e.g., a CNN) trained separately (Figure~\ref{fig:dia_ensemble_of_wds}~\subref{subfig:dia_training_phase}). 

Our proposed defense framework is:
\begin{itemize}
    \item \textbf{Extensible}: One can \emph{add} new WDs and, therefore, improve its effectiveness or can \emph{remove} WDs, which will sacrifice effectiveness but will also decrease the overhead. 
    \item \textbf{Flexible}: One can \emph{update} the ensemble, making it robust by replacing any (i) WDs and/or (ii) ensemble strategy. 
    \item \textbf{General}: One can use different \emph{models} to train WDs.
\end{itemize}


\subsection{Transformation as a Weak Defense}\label{subsec:trans_as_def}

\begin{figure}[t]
    \centering
    \includegraphics[scale=0.5]{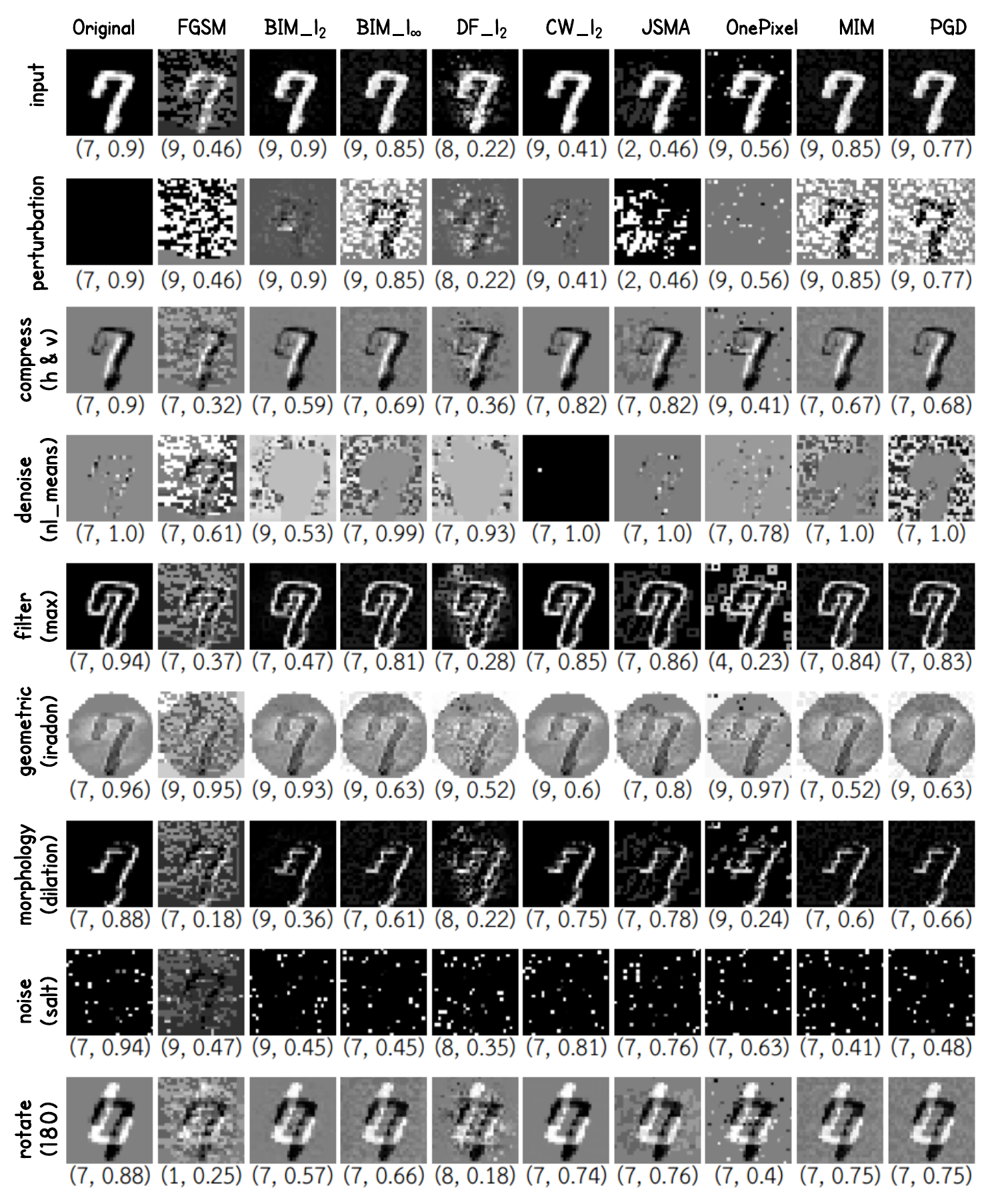} 
    \caption{Sample adversarial images generated by various adversaries and their corresponding distortions created by transformations in each row. (i) In the 1st row, we present inputs---an original input followed by the adversarial examples with predicted label and probability produced by the UM $f(\cdot)$. (ii) In the 2nd row, we present the perturbation generated by the adversary corresponding to the column. (iii) In each cell in rows 3--9, we present the distortion generated by applying the transformation in a current row to the input in the column with the predicted label and probability produced by the weak defense associated to the corresponding transformation.} 
    \label{fig:overview} 
\end{figure}

Transformations have shown to be effective in deterring some adversarial perturbations~\cite{guo2017transformation, tian2018transformation,bhagoji2018enhancing}.
To provide a more comprehensive understanding of the effectiveness of transformations, we examined a large variety of transformations (see Table~\ref{tab:transformations}) against AEs generated by various attacks. 
Figure~\ref{fig:overview} presents sample inputs from MNIST in the 1st row---an benign sample followed by AEs generated by the delineated adversaries that correspond to each column; the corresponding adversarial perturbations, $\|\bx-\bx'\|$, are shown in the 2nd row; and the distortions, $\|\bx' - t_i(\bx')\|$, generated by some sample transformations are shown in the 3rd--9th rows. 
Overall, the following observations have been made: 
\begin{itemize}
    \itemsep0em
    \item \emph{Each WD is able to correctly classify some of the AEs}. For example, the WD associated to morphology(dilation) can correctly classify samples generated by \FGSM, $\text{\BIM}\_l_{\infty}$, $\text{\CW}\_l_2$, \JSMA, \MIM, and \PGD.
    \item \emph{WDs complement each other}. For example, the AE generated by \FGSM~is misclassified by rotate($180^{\circ}$); however, it is correctly classified by morphology(dilation).
\end{itemize}

\subsection{Ensemble of Many Diverse Weak Defenses}\label{subsec:ensemble_mwd}

As observed in Figure~\ref{fig:overview}, the effectiveness of transformations varies according to the type of adversary. It is difficult to find a small set of transformations (i.e., 2--3) that are effective against all adversaries. Nevertheless, the observations from Figure~\ref{fig:athena_motivation} (and Figure ~\ref{fig:perf_surrogates}) suggest that WDs can complement each other by increasing the quantity and diversity of WDs. 


We train a WD $f_{t_i}$ by first applying a transformation, $t_i$, on the original data $\mathcal{D}$, and then training a classifier using the transformed data set $\mathcal{D}_{t_i}$. At test time, given an input $\bx$, we first apply a transformation $t_i$ on the input $\bx$ and then feed the transformed input $\bx_{t_i}$ to the corresponding WD $f_{t_i}$. \ourframework uses an ensemble strategy (e.g., majority voting) to compute the final predicted label $y$ by utilizing the predictions (probabilities and/or logits) of WDs. We elaborate on specific ensemble strategies in Section~\ref{subsec:ens_strategy}.

\section{Experimental Evaluation}\label{sec:empirical study}
\subsection{Experimental Setup}\label{sec:exp_setup}

We evaluated \ourframework~on MNIST~\cite{lecun2010mnist} and CIFAR-100~\cite{Krizhevsky09learningmultiple} against 9 different attack methods: \FGSM, \BIM($l_2\text{- and}~l_{\infty}$-norms), \CW($l_2$-norm), \JSMA, \PGD($l_{\infty}$-norm), \DF, \OP\, and \MIM.\footnote{\DF, \OP, and \MIM~were implemented on MNIST only.} For each attack, 5 sets of AEs were generated using various perturbation magnitudes. 
The attacks were implemented with Google CleverHans~\cite{Papernot2018cleverhans} for MNIST, and with \textsc{IBM ART}~\cite{art2018} for CIFAR-100.

\begin{table}[!thb]
    \centering
    \footnotesize
    \caption{The architecture and training parameters of the undefended model and weak defenses.}
    \scalebox{0.9}{\scriptsize{
\begin{tabular}{l|llll}
\toprule
Model & Architecture       &        &   Parameters          &           \\
\midrule
LeNet   & Conv.ReLU   & $3\times3\times32$      &   Optimizer &   Adam    \\
        & Max pooling & $2\times2$              &   Learning rate       &   0.001   \\
        & Conv.ReLU   & $3\times3\times64$      &   Batch size          &   128     \\
        & Max pooling & $2\times2$              &   Epoch              &   50      \\
        & Dense       & 4096                    &   &\\
        & Dropout     & 0.4                     &   &\\
        & Dense       & 10                      &   &\\
        & Softmax     & 10                      &   &\\
\midrule
SVM     & Kernel        & Linear                & Regularization parameter & 0.1    \\
\midrule
WResNet & Network   & Wide ResNet           &  Optimizer            &  SGD    \\
            & Depth     & $28$                  &   Learning rate       &   0.1   \\
            & $k$       & $10$                  &   Batch size          &   128     \\
            &           &                       &   Epoch               &   200      \\
            &           &                       &   LR scheduler        & cosine    \\
\midrule
ResNet-Shake & Network   & ResNet                & Optimizer             & SGD   \\
            & Depth     & $26$                  & Learning rate         & 0.1   \\
            & Residual branches &  $2$                  & Batch size            & 256   \\
            & 1\textsuperscript{st}-residual width &  $32$   & Epoch                 & 200   \\
            & Regularization &   Shake-Shake   & LR scheduler         & cosine    \\
\bottomrule
\end{tabular}
}
}
    \label{tab:nn_arch}
\end{table}

We built ensemble defenses upon $72$ WDs for MNIST and $22$ WDs for CIFAR-100 due to the computational cost. Each of the WDs is associated to one transformation (Table~\ref{tab:transformation_list}). When training a classifier, we used $80\%$ of the training data as for training and $20\%$ for validation. To show that \ourframework~works independent of the type of models, we built and evaluated two versions of \ourframework~for each data set---$28\times2$ Wide ResNet (WRN)~\cite{Zagoruyko2016WRN} and $26~2\times32d$ ResNet with Shake-Shake regularization (ResNet-Shake)~\cite{Gastaldi17ShakeShake} for CIFAR-100 and LeNet and linear SVMs for MNIST (Table~\ref{tab:nn_arch}). We will mainly illustrate the results using realizations with $28\times2$ WRN for CIFAR-100 and provide results for LeNet MNIST in the appendix.

In this study, we realized 4 ensemble strategies (Section~\ref{subsec:ens_strategy}) and evaluated their effectiveness against 4 forms of threat models (Section~\ref{subsec:threat_models}).
We also compared \ourframework~with two state-of-the-art adversarial defenses: (i)~\textit{PGD Adversarial Training} (PGD-ADT)~\cite{madry2018pgdadvtraining}, which enhances model's robustness by solving the robust optimization problem $\min\limits_{\theta}\max\limits_{\delta \in \Delta}\mathcal{L}(\bx + \delta, y; \theta))$ that performs gradient ascent on the input $\bx$ and gradient descent on the weights $\theta$ using \PGD~attack as the optimization algorithm. We implemented \PGD-ADT on top of IBM ART. (ii) \textit{Randomized Smoothing} (RS)~\cite{cohen2019randsmoothing}, which first generates $n$ noisy samples by adding random Gaussian noises to the input, $\bx$, collects predicted probabilities by feeding generated noisy samples to the model, and then filters out the predictions where the highest probability is not statistically significantly larger than the rest. Finally, RS computes corresponding labels for all valid predictions and determines the most frequent label as the final class of $\bx$. We executed RS experiments using scripts from the original work.\footnote{\url{https://github.com/locuslab/smoothing}}
The experiments (e.g., training WDs, crafting AEs) were conducted on multiple machines (Table~\ref{tab:hardware_config}).

\subsection{Ensemble Strategies}\label{subsec:ens_strategy}
We implemented and evaluated 4 ensemble strategies:
\begin{itemize}
    \itemsep0em
    \item \textbf{Random Defense (RD)} uses a random WD to predict the input. Its expected effectiveness equals to the average of all WDs, therefore, we use it as a baseline against which to compare \ourframework. 
    \item \textbf{Majority Voting (MV)} determines the label agreed upon by most WDs. 
    \item \textbf{Top 2 Majority Voting (T2MV)} 
    collects labels associated to the top two probabilities and, thereafter, performs majority voting among them. As we observed from Figure~\ref{fig:overview}, the predicted probability distributions of some WDs on certain AEs are soft, such that those WDs are less confident with their predictions, whereby the correct answer may lie in the classes for which they are less confident.
    \item \textbf{Average Output (AVEO)} takes an average of the outputs from WDs, and then returns the label with the highest averaged value (i.e., applies $\argmax$ on the averaged outputs). For WDs built from neural networks, the output can be selected from any layer. Two specific variants of AVEO are AVEP and AVEL, which use probability and logit as the output respectively. 
\end{itemize}

\subsection{Threat Models}\label{subsec:threat_models}
We make two assumptions with respect to threat models: that the attackers can attack only at the inference phase and the attackers can modify only the input data (i.e., the attackers cannot modify the models nor the training data).
We evaluate \ourframework~with the following threat models~\cite{carlini2019evaluating,biggio2018wild}:
\begin{itemize}
    \itemsep0em
    \item \textbf{Zero-Knowledge} is the weakest model, where the attacker is not even aware that a defense is in place. However, here we assume that the attacker knows the exact architecture and parameters of the UM. It is important to verify this weak model, because failing this test implies that the stronger tests under adaptive attackers will also fail~\cite{carlini2019evaluating}.
    \item \textbf{Black-box} model assumes that the adversary has no knowledge of the ensemble model, but it is aware of the fact that a defense mechanism is protecting the classifier. Here, we assume that the attacker is not aware of the whole training data, it does, however, have access to a portion of training data albeit without the correct labels.
    \item \textbf{White-box} is the strongest threat model, in which the adversary has full knowledge of the ensemble defense model. That is, the attacker knows the architecture and hyper-parameters of UM and all WDs, the list of transformations (or, in the Grey-box threat model, a partial list), and the ensemble strategy being used. 
\end{itemize}

\subsection{Zero-Knowledge Attack} \label{subsec:eval_zk}

\begin{figure*}[!htb]
    \captionsetup[subfigure]{font=scriptsize,labelfont=scriptsize}
    \footnotesize
    \centering
    \subfloat[][\FGSM]{
        \includegraphics[width=0.33\linewidth, trim={0.5cm 0.7cm 0.5cm 1.5cm}]{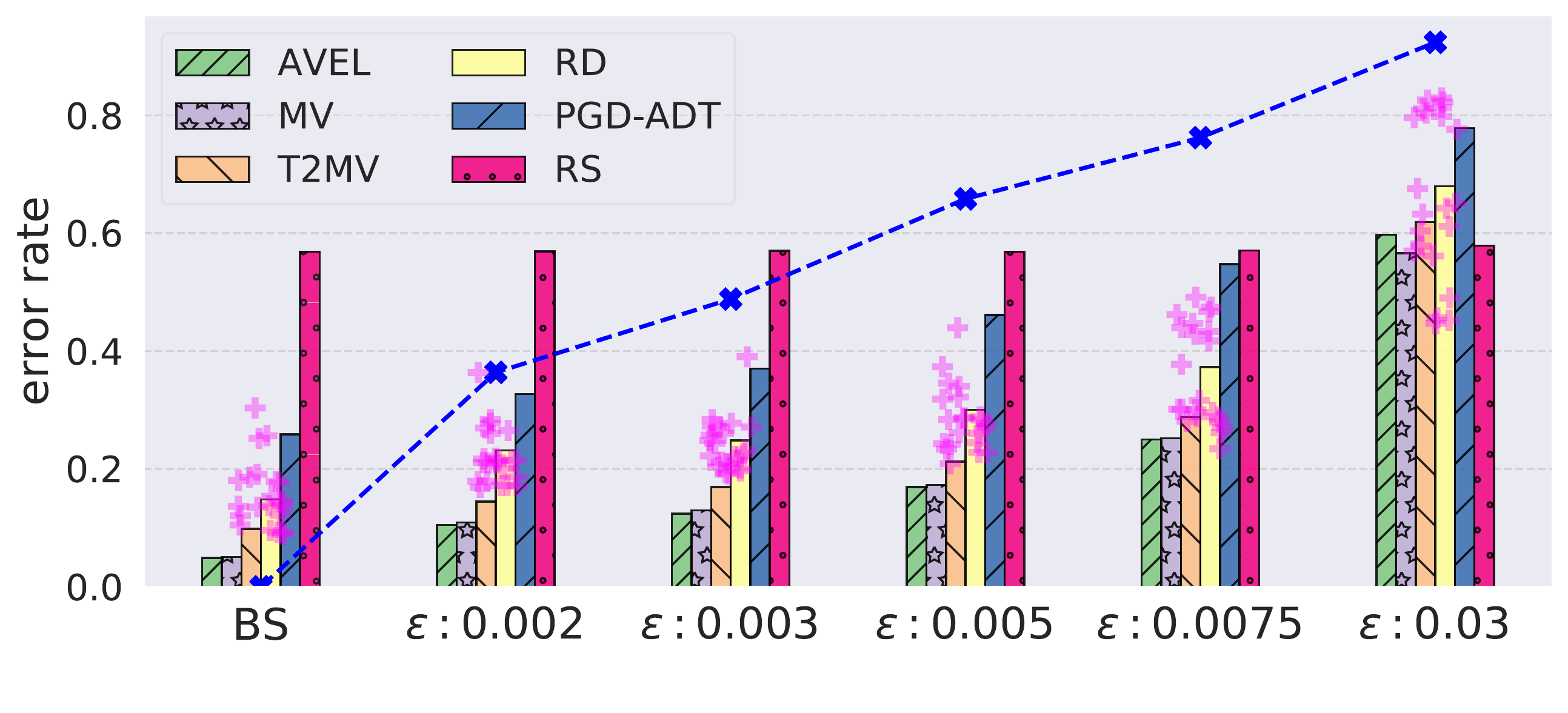}
    }
    \subfloat[][\BIM\_$l_2$]{
        \includegraphics[width=0.33\linewidth, trim={0.5cm 0.7cm 0.5cm 1.5cm}]{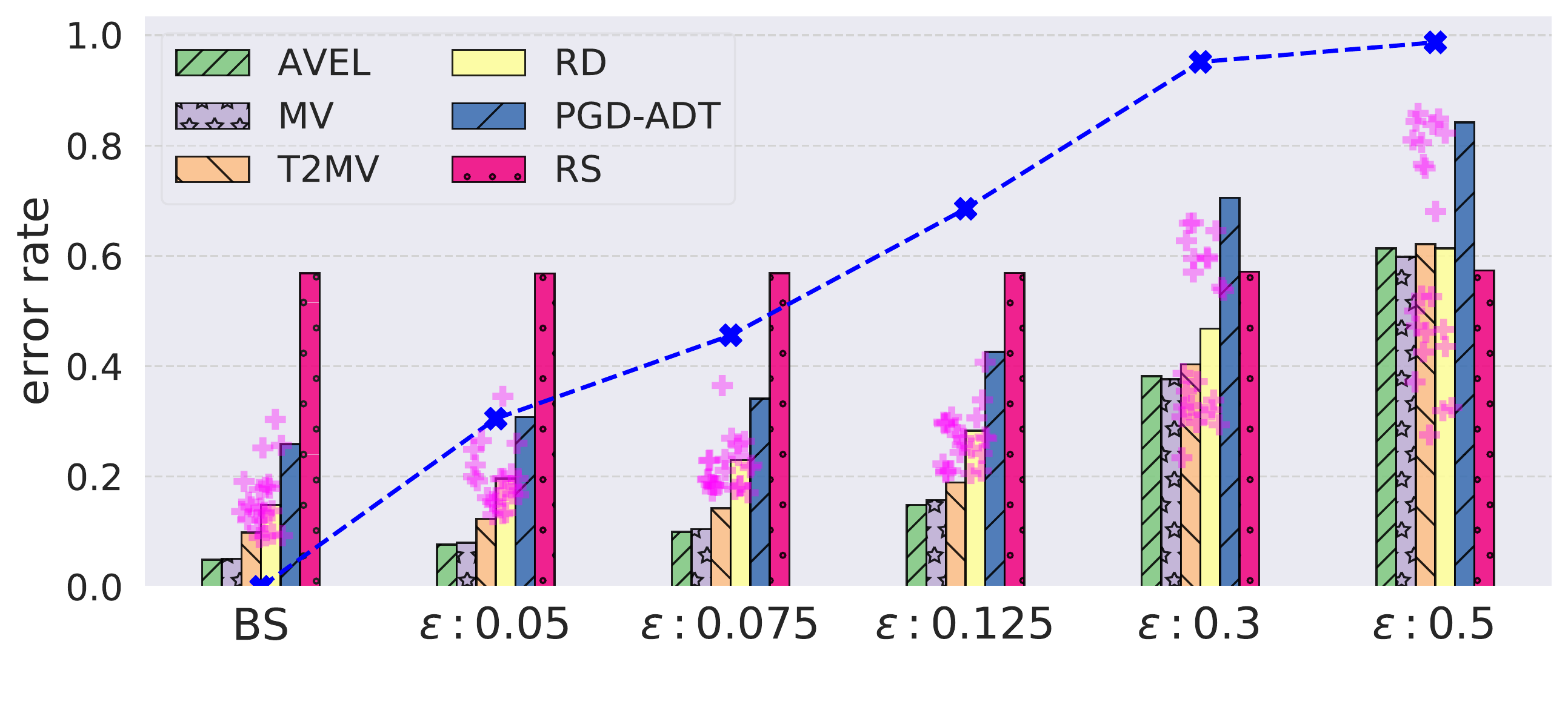}
    }
    \subfloat[][\BIM\_$l_{\infty}$]{
        \includegraphics[width=0.33\linewidth, trim={0.5cm 0.7cm 0.5cm 1.5cm}]{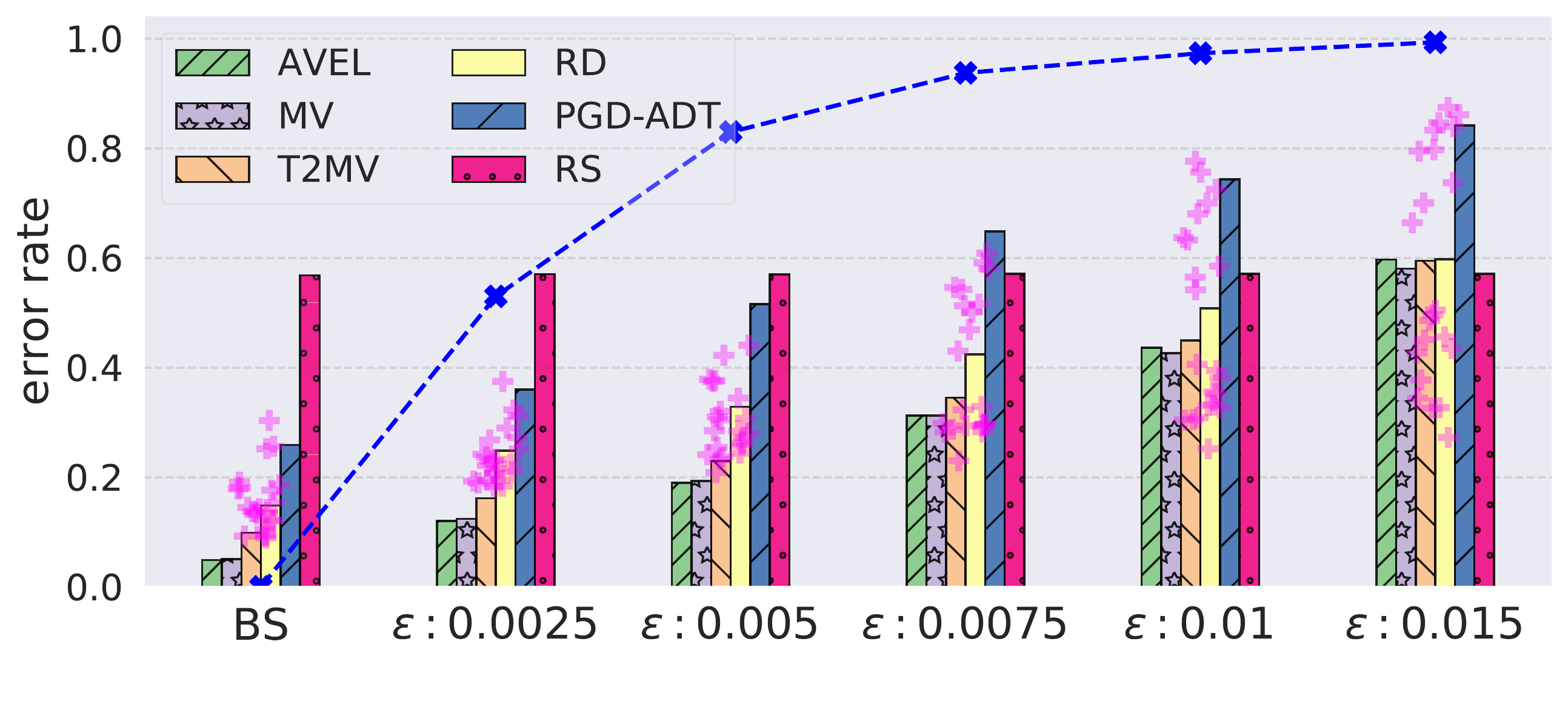}
    } \\
    \subfloat[][\CW\_$l_2$]{
        \includegraphics[width=0.33\linewidth, trim={0.5cm 0.7cm 0.5cm 1.5cm}]{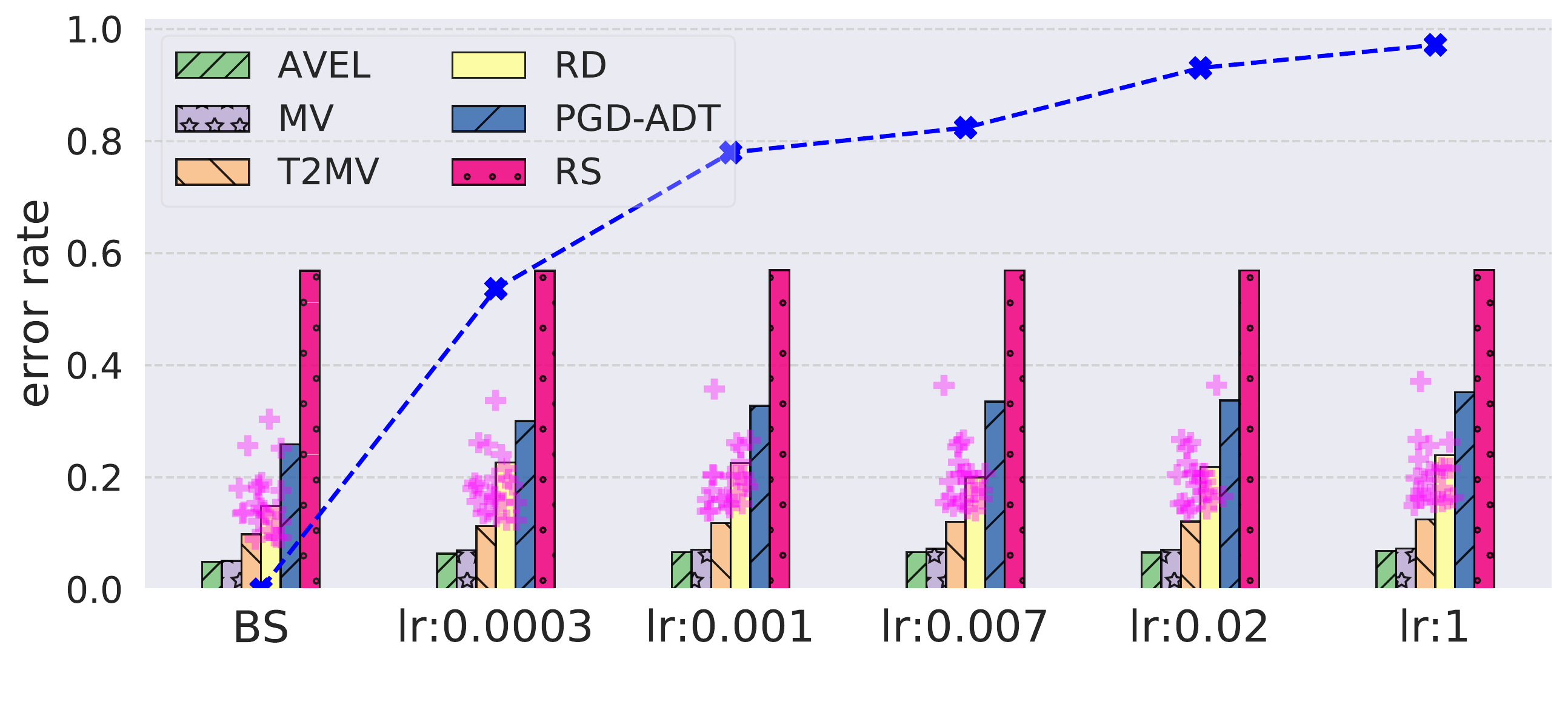}
    }
    \subfloat[][\JSMA]{
        \includegraphics[width=0.33\linewidth, trim={0.5cm 0.7cm 0.5cm 1.5cm}]{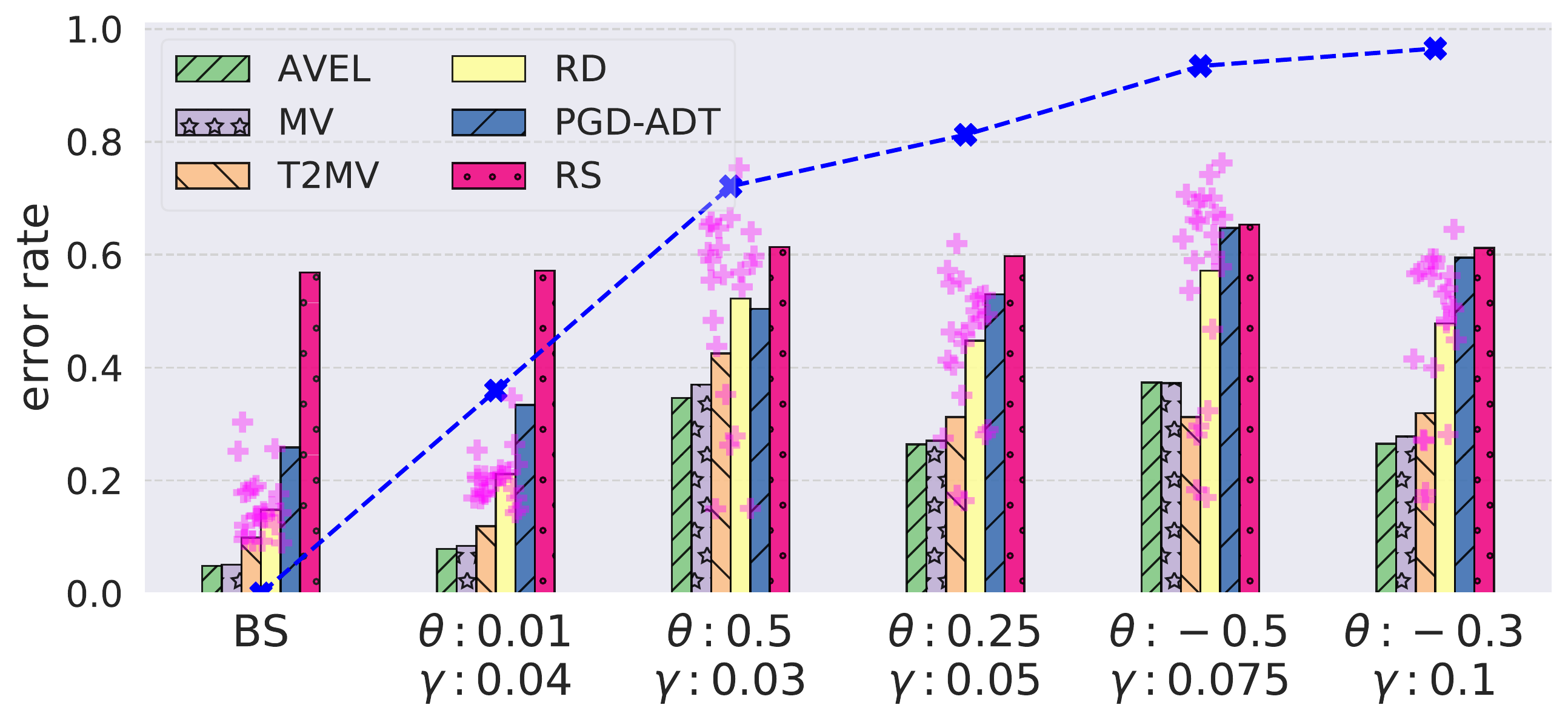}
    }
    \subfloat[][\PGD]{
        \includegraphics[width=0.33\linewidth, trim={0.5cm 0.7cm 0.5cm 1.5cm}]{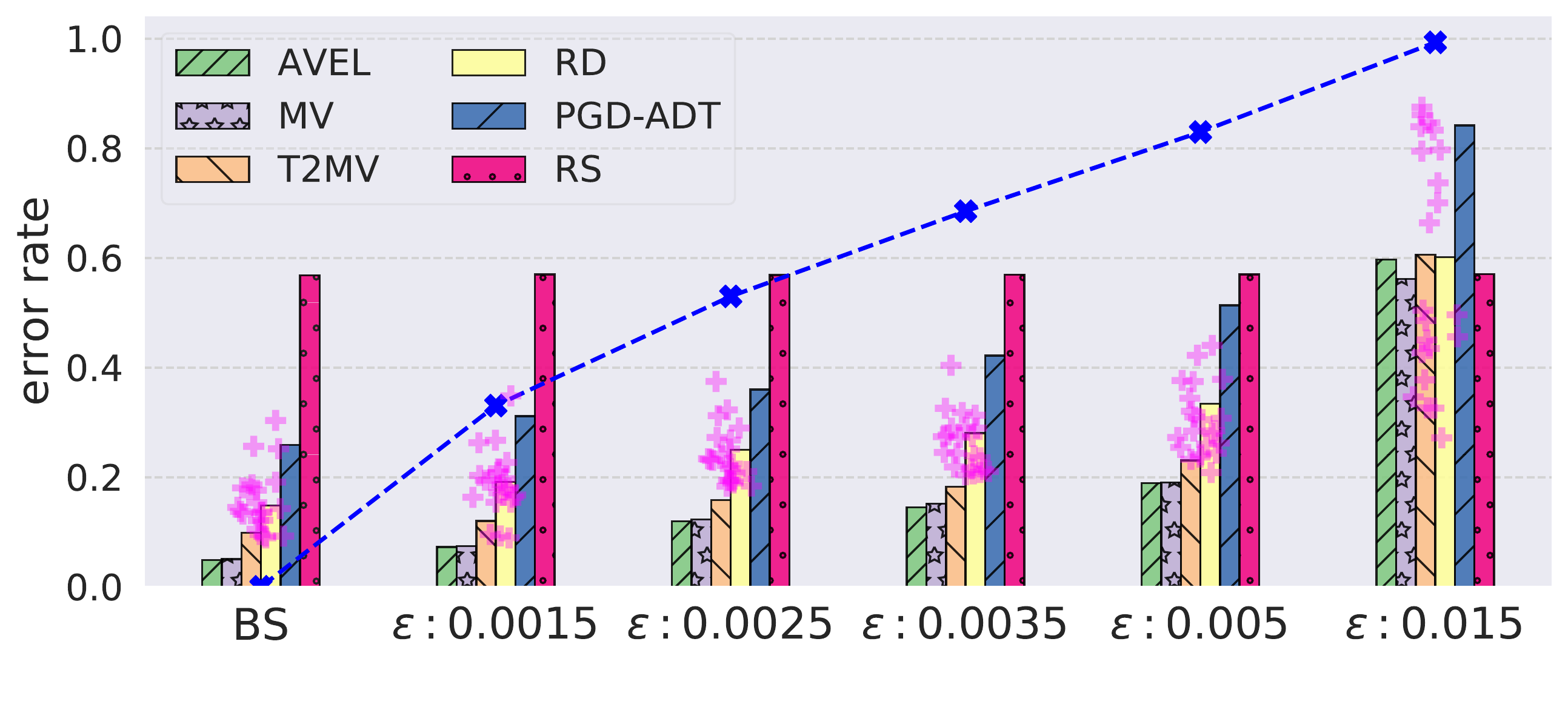}
    }
    \caption{Evaluation of $28\times10$ WRN models on CIFAR-100. Models are evaluated with respect to error rate on AEs.  For each type of attack, the perturbations in the data set becomes stronger from left to right. The results of UM are plotted as blue dotted lines and each WD as pink ``+''.}
    \label{fig:eval_zk_cifar100-wres}
\end{figure*}

In the context of zero-knowledge threat model, we assume that the attacker has full knowledge about UM and nothing else. Therefore, all adversarial examples were generated based on UM. We evaluated 4 realizations (i.e., RD, MV, T2MV, and AVEL) of \ourframework, consisting of 22 WDs, against 6 attacks on CIFAR-100. For each attack, we varied their configurations to generate 5 different strength levels (Table~\ref{tab:zk_attack_config}). 
The error rate of different realization of \ourframework~were then compared with that of PGD-ADT and RS. PGD-ADT was run with $\epsilon=0.015$. Multiple trials were launched to evaluate RS, using various $\sigma$ values: $0.05, 0.075, 0.1, 0.3,$ and $0.5$. Only the best results, where $\sigma=0.075$, were reported. In each attempt, 100 noisy samples were generated for each input.

\begin{figure}[t]
    \footnotesize
    \centering
    \includegraphics[width=\linewidth]{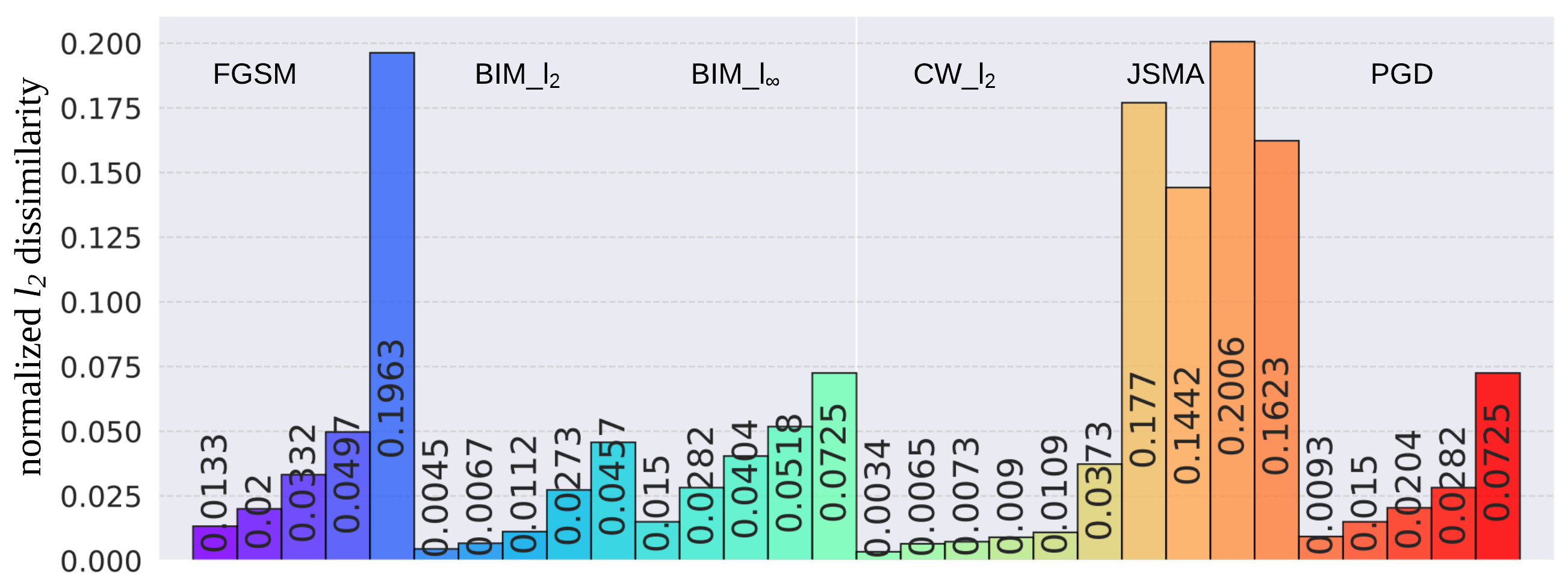}
    \caption{Normalized $l_2$ dissimilarity between BS and each AE generated for CIFAR-100.}
    \label{fig:dist_zk_cifar100} 
\end{figure}

In this study, evaluation results are reported with respect to model's error rate on adversarial examples whose corresponding benign samples were correctly classified by UM. 
Therefore, the error rate of UM on BS in the report is $0$. Our UM achieves a top-1 test accuracy of $75.54\%$ on CIFAR-100.
Figure~\ref{fig:eval_zk_cifar100-wres} presents the evaluation results for \ourframework, being built upon $28\times10$ WRNs. For each type of attack, UM (the blue dotted line) made more errors on stronger AEs (from left to right, cf. $l_2$ dissimilarities in Figure~\ref{fig:dist_zk_cifar100}). 

The effectiveness of RS is stable across attacks and perturbation magnitudes. The variation in the error rate of RS is very small ($8.5\%$ from $56.82\%$ to $65.36\%$), a byproduct of a certified defense. 
However, the effectiveness of RS, on BS and AEs with small perturbations, is even worse than UM (an expected tradeoff for certified defenses). In general, PGD-ADT makes fewer errors than RS on AEs with weak perturbations in each group of attacks. While as the perturbations become stronger, the error rates of PGD-ADT increase and eventually go beyond that of RS. 

The effectiveness of individual WDs varies on each test set (BS or AE). The variation of the error rates achieved by WDs spans wider as the perturbation magnitude becomes stronger in each group. For example, evaluating on \BIM\_$l_2$ AEs, error rates of WDs vary from $13.07\%$ to $34.56\%$ when $\epsilon=0.05$ and from $27.54\%$ to $85.82\%$ when $\epsilon=0.5$. The RD ensemble identifies the baseline of \ourframework. As by labeling the input using a random WD each time, RS ensemble is expected to achieve the average error rates of all WDs. Therefore, as long as there are enough number of WDs are efficient, we have a good chance of building a robust RD ensemble. Furthermore, by applying a more intelligent strategy, like MV, T2MV, or AVEL, \ourframework~becomes more resilient against such attacks. For example, on \JSMA($\theta=0.5,\gamma=0.03$), the baseline made an error rate of $52.25\%$ vs. $50.4\%$ on PGD-ADT, where more than $69\%$ of WDs make more errors than PGD-ADT does. However, by applying $\argmax$ on the averaged predictions of all WDs (AVEL), \ourframework~achieves an error rate of $34.64\%$, which is much more efficient than PGD-ADT. 

Overall, all of the 4 ensembles are more efficient than PGD-ADT and RS on both BS and AEs. They are the 4 top performers (i.e., make the fewest errors on AEs)  across attacks and perturbation magnitudes, except the extreme cases in 4 groups---\FGSM~($\epsilon=0.03$), \BIM\_$l_2$ ($\epsilon=0.5$), \BIM\_$l_{\infty}$ ($\epsilon=0.015$), and \PGD~($\epsilon=0.015$). In those extreme cases, the ensemble defenses achieve an effectiveness close to the top performer (RS or PGD-ADT), the gaps between the best ensemble defense and the top performer is $[2\%, 4\%]$.

A very similar evaluation was performed on MNIST, but in a larger scale: (i) realized 5 ensembles---RD, MV, T2MV, AVEL, and AVEP, (ii) each ensemble consists of 72 WDs, and (iii) 9 attacks have been launched. Our UM achieves a test accuracy of $99.01\%$ on MNIST. As presented in Figure~\ref{fig:eval_zk_mnist-cnn} and Figure~\ref{fig:dissimilarity_MNIST}, the MNIST results are similar to that in CIFAR-100. PGD-ADT and RS achieve an error rate lower than $10\%$ across all attacks except on \FGSM~AEs (where $\epsilon=0.2, 0.25,~\text{and}~0.3$) and \DF~AEs (where $\text{os}=8, 16, 50$). \ourframework~with AVEL or T2MV ensemble strategy achieves the lowest error rate, across all attack and perturbation magnitudes, however, with small margin. 

\subsection{Black-box Attack}\label{subsec:eval_bb}

We evaluated \ourframework~with two black-box attacks: (i)~\emph{transfer-based}~\cite{papernot2017blackbox, guo2019simple}, (ii) \emph{gradient-direction estimation} approach based on HopSkipJump~\cite{chen2019hopskipjumpattack}. We evaluated one realization of \ourframework---AVEP ensemble with 12 WDs. 

\textbf{The transfer-based attack} assumes AEs are transferable between models \cite{szegedy2013intriguing, goodfellow2014explaining}, and consist of substitute model training and adversarial sample crafting. The attackers construct a substitute model, $f_{\text{sub}}$, mimicking the target model, $f_{\text{target}}$, using only limited training data.\footnote{It may be flagged suspicious if the attacker queries the system many times.} The attackers then use the substitute model to create AEs and to launch attacks on the system. This strong assumption challenges the attacker in many ways when crafting an AE, such as selecting a particular model architecture for building the substitute model and preparing a sample set for label collection given a limited query budget~\cite{papernot2017blackbox,guo2019simple}. Although we could consider different assumptions for the knowledge of the attacker in black-box attacks, we instead considered the strongest possible black-box attack, where the adversary knows the exact architecture, optimization strategy, and hyper-parameters of the original target model. The attacker can then use the substitute model's parameters and perform a white-box attack on the substitute model. Finally, we test how well these AEs can transfer from the substitute model to the target model by calculating the \textit{transferability rate}, the proportion of these AEs misclassified by the target model. We evaluate the black-box threat model with transferability-based attack as follows:
\begin{enumerate}[1)]
    \itemsep0em
    \item Collect a data set of $N$ samples, $\mathcal{D}_{\text{bb}} = \{(\bx, \by) | \by = f_{\text{target}}(\bx), \bx \in \mathcal{D}\}$, by querying the target model $N$ times.
    \item Build a substitute model $f_{\text{sub}}$, mimicking the target model $f_{\text{target}}$, which is trained on the collected data set $\mathcal{D}_{\text{bb}}$.
    \item Generate adversarial examples to attack $f_{\text{sub}}$.
    \item Attack $f_{\text{target}}$ using the AEs crafted based on $f_{\text{sub}}$.    
\end{enumerate}

For training substitute models and evaluating the defense, we split the CIFAR-100 test set ($10k$ samples) into two parts: (i) training ($5k$ samples) and testing ($5k$ samples). For a given a budget, $B$, the set, $\mathcal{D}_{\text{bb}}$, for training the substitute model is composed of $B$ samples selected from the training set, which can be correctly classified by the target model. The class labels of these selected $B$ samples evenly distribute across all 100 classes of CIFAR-100. The adopted budget list is $[1k, 2k, 3k, 4k, 5k]$. That is, in total, we trained 5 substitute models for each target model. The performance of each substitute model was evaluated with the test set of $5k$ samples. The top-1 and top-5 accuracy of the trained substitute models built against UM and defended model with \ourframework~are comparable (see Figure~\ref{fig:perf_surrogates}), so the success of the defense can be attributed to the drop in the transferability rate. 

Once we trained the substitute model associated to each budget, we generate AEs against the substitute model. For each substitute model, we crafted AEs using the 3 adversarial attacks, FGSM, BIM\_$l_\infty$ and PGD. For each attack, a list of 5 perturbations was used. 
For each substitute model, we selected $500$ (evenly distributed across all classes) out of the remaining $5k$ test samples for generating AEs. 

\begin{figure}
    \footnotesize
    \centering
    \includegraphics[width=\linewidth]{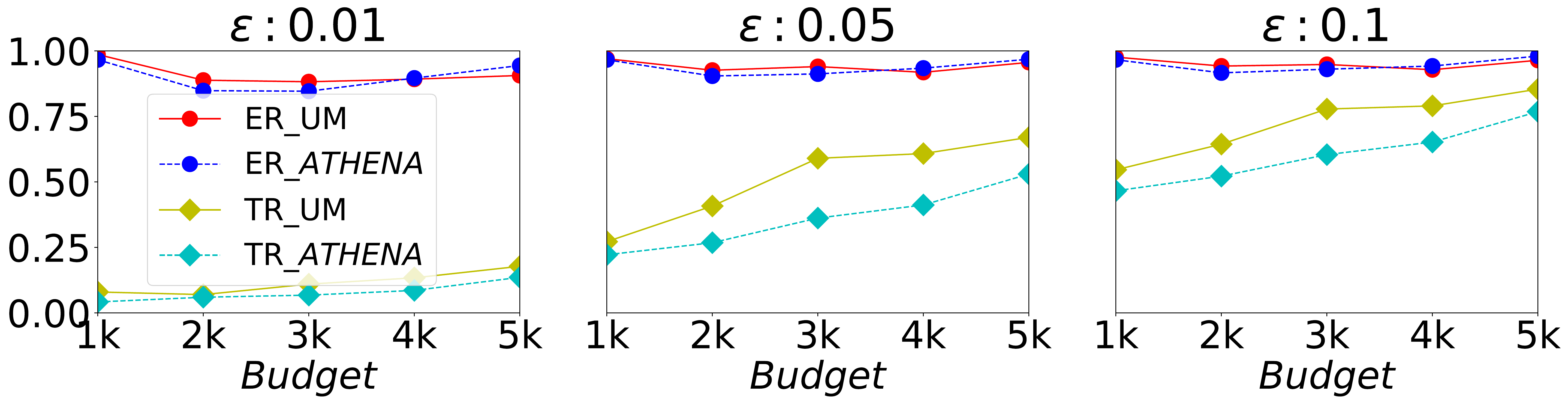}
    \caption{Performance of transferability-based black-box attack with different budgets and using BIM\_$l_{\infty}$ with different perturbations to attack each trained substitute model. Note: ER, TR and UM stand for error rate, transferability rate and undefended model respectively.}
    \label{fig:perf_transfer_bb_attack_BIMLinf}
\end{figure}

Figure~\ref{fig:perf_transfer_bb_attack_BIMLinf} shows the result of this transfer-based attack using BIM\_$l_\infty$ with three perturbations: $[0.01, 0.05, 0.1]$. In all scenarios (all combinations of perturbation and budget here), \ourframework~lowers the transferability rate achieved by the attack when compared to UM. As expected, the transferability rate increases when the budget increases. 
The drop in transferability rate from UM to \ourframework~becomes more evident when the perturbation increases, which indicates that \ourframework~is less sensitive to the perturbation. In the low perturbation ($\epsilon=0.01$) scenario, the drop of transferability rate achieved by \ourframework~is not significant with an average value of $3.6\%$, while the averaged drop in the perturbation ($\epsilon=0.1$) increases to $12\%$. The reason is that AEs generated with the perturbation $\epsilon=0.01$ are mostly ineffective with very low transferability rates of averaged values of $11.44\%$ and $7.84\%$ for UM and \ourframework~respectively, while AEs generated with the perturbation $\epsilon=0.05$ are much effective with average transferability rage of $72.24\%$ and $60.24\%$ for UM and \ourframework~respectively. The ratio of decreased transferability rate is $25.82\%$ on average. To limit the computation, we only realized \ourframework~with 12 WDs, but we expect to observe larger drop in transferability rate with higher number of WDs.

As seen in Figure~\ref{fig:perf_transfer_bb_attack}, a similar trend is also observed when using the other two attacks. It is important to emphasize that the results could have been even better for our defense mechanism under under a weaker black-box threat model, where the knowledge of DNN architecture and hyper-parameters is not known by the adversary, such that the attacker confronts greater difficulty in crafting effective AEs. 
Lastly, \ourframework~can employ randomness, like the RD ensemble, resulting into non-deterministic output of the ensemble. This will propagate randomness to the training process of substitute models, and hence result in lower accuracy of the substitute models, and finally limit the attacker's strength.

\textbf{Gradient-direction estimation} approaches are based on search using an estimation of the gradient direction, where the attacker has access to only the output probabilities (score-based) or the label (decision-based). We evaluated \ourframework~with the state-of-the-art decision-based attack, HopSkipJump Attack (HSJA)~\cite{chen2019hopskipjumpattack}, which only requires the output labels from the target model and is able to create AEs with much fewer budget (the number of queries to the target model) than existing approaches. It finds an AE that is closest to the corresponding benign sample based on an estimation of the gradient between adversarial group and benign group. 

We launched two HSJA variants ($l_2$ and $l_{\infty}$) against both UM and \ourframework~using the open-sourced code.\footnote{HopSkipJumpAttack: github.com/Jianbo-Lab/HSJA} For each of UM and \ourframework, 500 out of $5k$ CIFAR-100 test samples (the same test set mentioned in the experiment of transfer-based black-box attack) were selected for generating AEs. 

\begin{figure}
    \footnotesize
    \centering
    \includegraphics[width=0.9\linewidth]{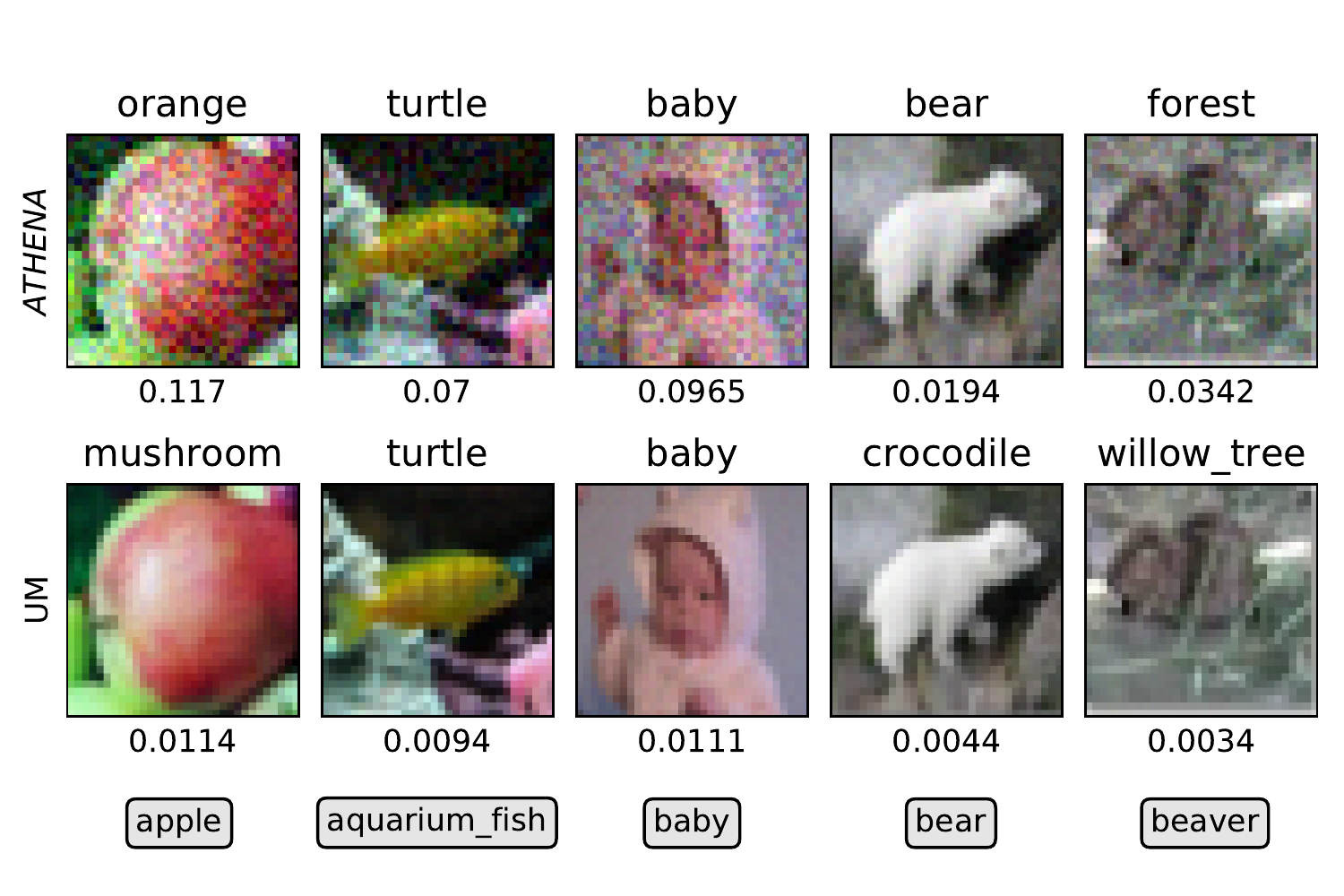}
    \caption{Adversarial examples (CIFAR-100) generated by HSJA with $l_{\infty}$ distance and a budget of 5000 queries against the undefended model (UM) and \ourframework. Note: 1) the text above each subfigure is its predicted label by the corresponding target model; 2) the number under each subfigure indicates its $l_{\infty}$ distance from the original benign sample; 3) true labels are displayed in round rectangles.}
    \label{fig:hsja_linf-um_vs_avep}
\end{figure}

\begin{figure}
    \footnotesize
    \centering
    \includegraphics[width=0.9\linewidth]{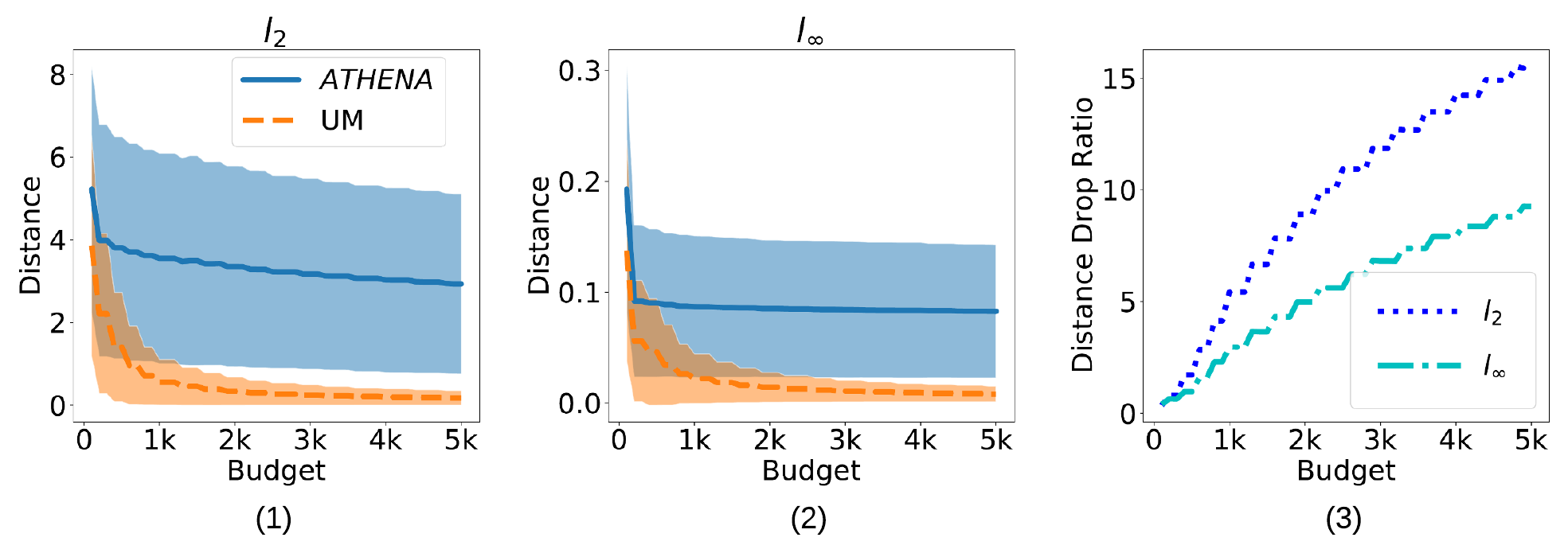}
    \caption{Mean distance versus query budget and distance drop ratio versus query budget on CIFAR-100 with the undefended model (UM) and \ourframework.}
    \label{fig:hsja_result}
\end{figure}

An AE created by HSJA with a given budget is considered as invalid if its distance to the benign sample is larger than the given distance threshold. Therefore, a target model is considered to have a better defense capability if the AE generated from it has a larger distance value. Figure~\ref{fig:hsja_linf-um_vs_avep}, shows some example perturbed images.  We observe that 1) for each of the benign samples of apple, aquarium\_fish and beaver, the crafted AE by HSJA against \ourframework~has a much larger distance (7.5x-10x) to the benign sample when compared to the AE against the UM. 2) HSJA is able to utilize the benign example of bear to launch attack against the UM but fails when facing \ourframework, even though 3.4x more perturbation is added. 3) for the benign sample of baby, HSJA fails to attack both target models. An increase of query budget might help here. With $l_2$ distance, HSJA needs 23.7x more perturbation to attack \ourframework~when compared to the UM (see Figure~\ref{fig:hsja_l2-um_vs_avep}).

To further demonstrate the effectiveness of \ourframework, we explored the averaged distances of AEs versus budget for both UM and \ourframework. As seen Figure~\ref{fig:hsja_result}~(1)~(2), given a budget, \ourframework~forces HSJA to generate AEs with much larger distances when compared to the UM. This clearly corroborates that \ourframework is effectively protecting the UM against HSJA. To quantify the defense capability, we use $\text{Distance Drop Ratio} = \frac{\text{dist}_{\text{\ourframework}}-\text{dist}_{\text{UM}}}{\text{dist}_{\text{UM}}}$, where $\text{dist}_{\text{\ourframework}}$ and $\text{dist}_{\text{UM}}$ are distances of AEs targeting \ourframework~and UM respectively. For a given query budget from $[100, 5000]$, \textit{Distance Drop Ratio} is from 0.36 to 15.45 and from 0.40 to 9.27 for $l_2$ distance-based attack and $l_{\infty}$ distance-based attack respectively. As seen from Figure~\ref{fig:hsja_result}~(3), \textit{Distance Drop Ratio} increases when the given query budget increases. The reason is that distances of UM-targeted AEs drop much more significantly than that of \ourframework-targeted AEs when budget increases. That is, \ourframework~is robust against HSJA with varying query budget. With a higher number of WDs, we expect a larger distance gap, i.e., a better defense capability, since the robustness of the ensemble will be enhanced in general.

\subsection{Gray-box and White-box Attacks}\label{subsec:eval_wb_gb}
We evaluated \ourframework~with two white-box attacks: (i)~a greedy approach \emph{aggregating the perturbations} generated based on individual WDs; (ii) an \emph{optimization-based approach} based on \cite{athalye2017synthesizing, Liu2018_self-ensemble}. We evaluated one realization of \ourframework---AVEL ensemble with 22 WDs on CIFAR-100 and AVEP ensemble with 72 WDs on MNIST.

\textbf{Greedy approach.}
We use Algorithm~\ref{alg:attack_white-box} to generate AEs, each of which is able to fool at most a certain number, \vars{N}, of WDs within a specific distance, measured by normalized $l_2$ dissimilarity, between the perturbed and the benign samples. The value of \vars{N} depends on the ensemble strategy being used, the attack being launched, and the time budget the attacker is willing to pay. For example, for an ensemble model using MV strategy, the attacker has to fool at least half of the WDs to guarantee an ineffective MV ensemble with a targeted attack. By forcing more than half of the WDs agree on a label $y'$, where $y' \neq y$ and $y$ is the true label for a given input $\bx$, the MV ensemble will always incorrectly label $\bx$ as $y'$. The use of a maximum dissimilarity ensures that the sample is not being perturbed too much, also acts as a factor that indicates the efforts that the attacker can afford. In the algorithm, any attack methods (e.g., \FGSM, \CW\_$l_2$, etc.) can be used to generate the adversarial perturbations. 


\begin{algorithm}[t]
\SetAlgoLined
\SetKwInOut{Input}{input}\SetKwInOut{Output}{output}
\scriptsize 
\caption{Crafting white-box AEs (Greedy)}
\label{alg:attack_white-box}
    \Input{\vars{$\bx$}, \vars{$\by$}, \vars{attacker}, \vars{N}, \vars{max\_dissimilarity}}

    $F_{\textit{fooled}} \leftarrow$ \{\}\;
    $F_{\textit{cand}} \leftarrow$ all weak defenses\;
    $\bx' \leftarrow \bx$\;
    
    \While{\func{$size(F_{\textit{fooled}})$} $<$ \vars{N}}{
        $f_{\textit{target}} \leftarrow$ \func{pickTarget($F_{\textit{cand}}$, \vars{strategy})}\;
        
        \textcolor{blue!20!black!30!green}{
        \tcp{$\textit{getPerturbation}(\bx)$ returns $\| \bx - \bx' \|_2$}}
        
        $\vars{perturbation} \leftarrow $ \vars{attacker}.\func{getPerturbation($f_{\textit{target}}$, $\bx'$)}\;
        $\bx'_{\textit{tmp}} \leftarrow \bx' + \vars{perturbation}$
        
        \textcolor{blue!20!black!30!green}{
        \tcp{$\textit{dissimilairity}(\bx', \bx)$ returns the normalized $l_2$ dissimilarities between $\bx'$ and  $\bx$}}
        
        \If{\func{dissimilarity($\bx'_{\textit{tmp}}, \bx$)} $>$ \vars{max\_dissimilarity}}{
            break\;
        }
        
        \For{$f_{t_i} \text{in}~ F_{\textit{cand}}$}{
            \If {$\by \neq f_{t_i}(\bx'_{\textit{tmp}})$}{
                \func{addModel($F_{\textit{fooled}}$, $f_{t_i}$)}\;
                \func{removeModel($F_{\textit{cand}}$, $f_{t_i}$)}\;
            }
        }
        $\bx' \leftarrow \bx'_{\textit{tmp}}$\;
    }
    \textbf{return} $\bx'$\;
\end{algorithm}

In the context of white-box threat model, we evaluated MV ensemble built with 22 WDs on CIFAR-100. In order to add perturbations related to a variety of transformations into a single adversarial example within a reasonable time, we choose \FGSM~($\epsilon=0.01$). The smaller $\epsilon$ being used the longer time it will take to generate an AE to successfully fool a single WD by Algorithm~\ref{alg:attack_white-box}. We evaluated MV ensemble against 500 AEs, evenly distributed across 100 classes, with different configurations as specified in Table~\ref{tab:wb_setting}. In white-box threat model, beside the error rate, we also measure the attacker's effort in launching a white-box attack against \ourframework~with respect to (i) the time for generating one AE and (ii) the number of iterations to craft an AE that is able to fool a sufficient number of WDs, (iii) normalized $l_2$ dissimilarity, and (iv) the number of WDs it is able to fool. 

 
 
\begin{figure}[t]
    \scriptsize
    \centering
    \includegraphics[width=0.75\linewidth]{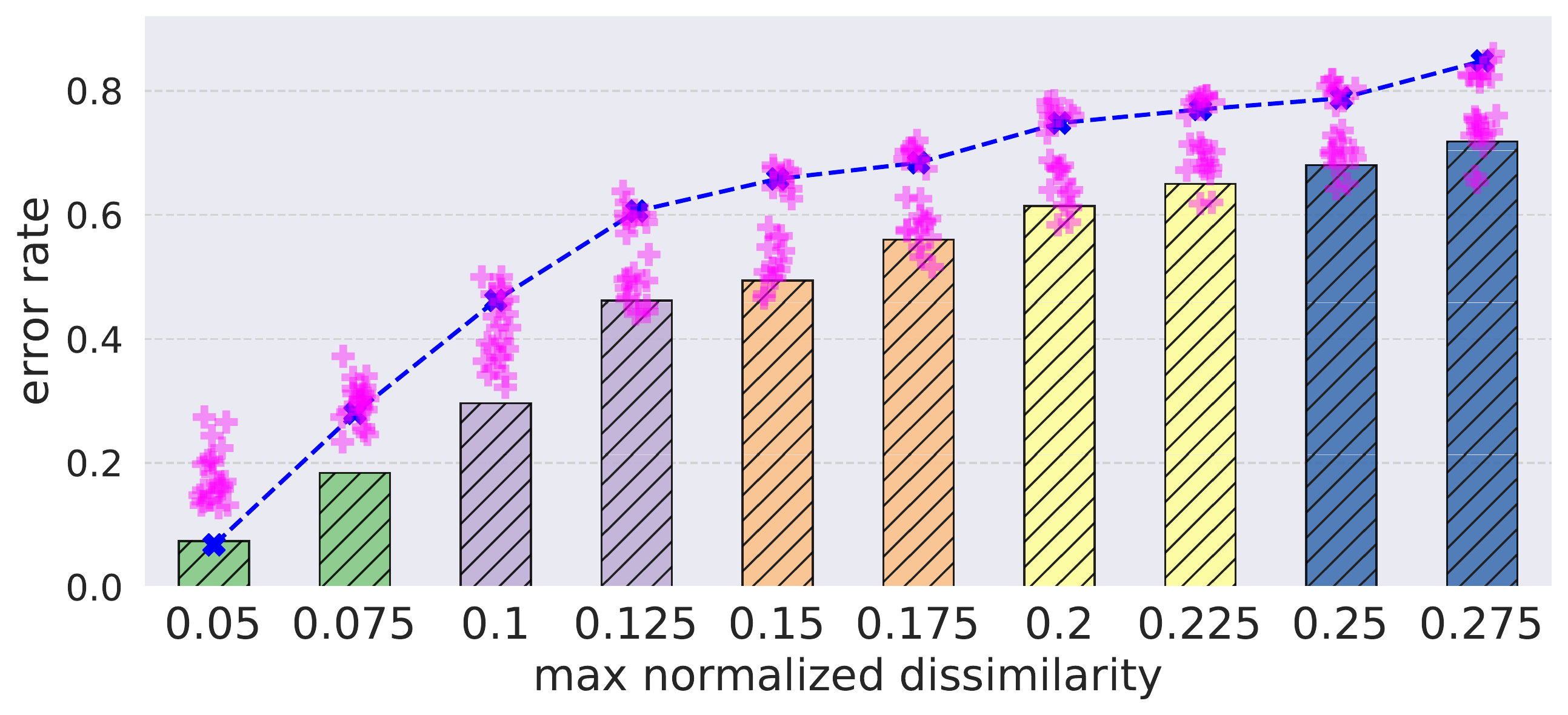}
    \caption{Evaluation of UM and MV ensemble on the greedy attack on CIFAR-100. 10 AE variants were generated using \FGSM. From left to right, the AEs were generated with a larger constraint of max dissimilarity with respect to the corresponding benign samples.}
    \label{fig:errrate_wb_cifar100}
\end{figure}

\begin{figure}[t]
    \tiny
    \centering
        \includegraphics[width=0.65\linewidth]{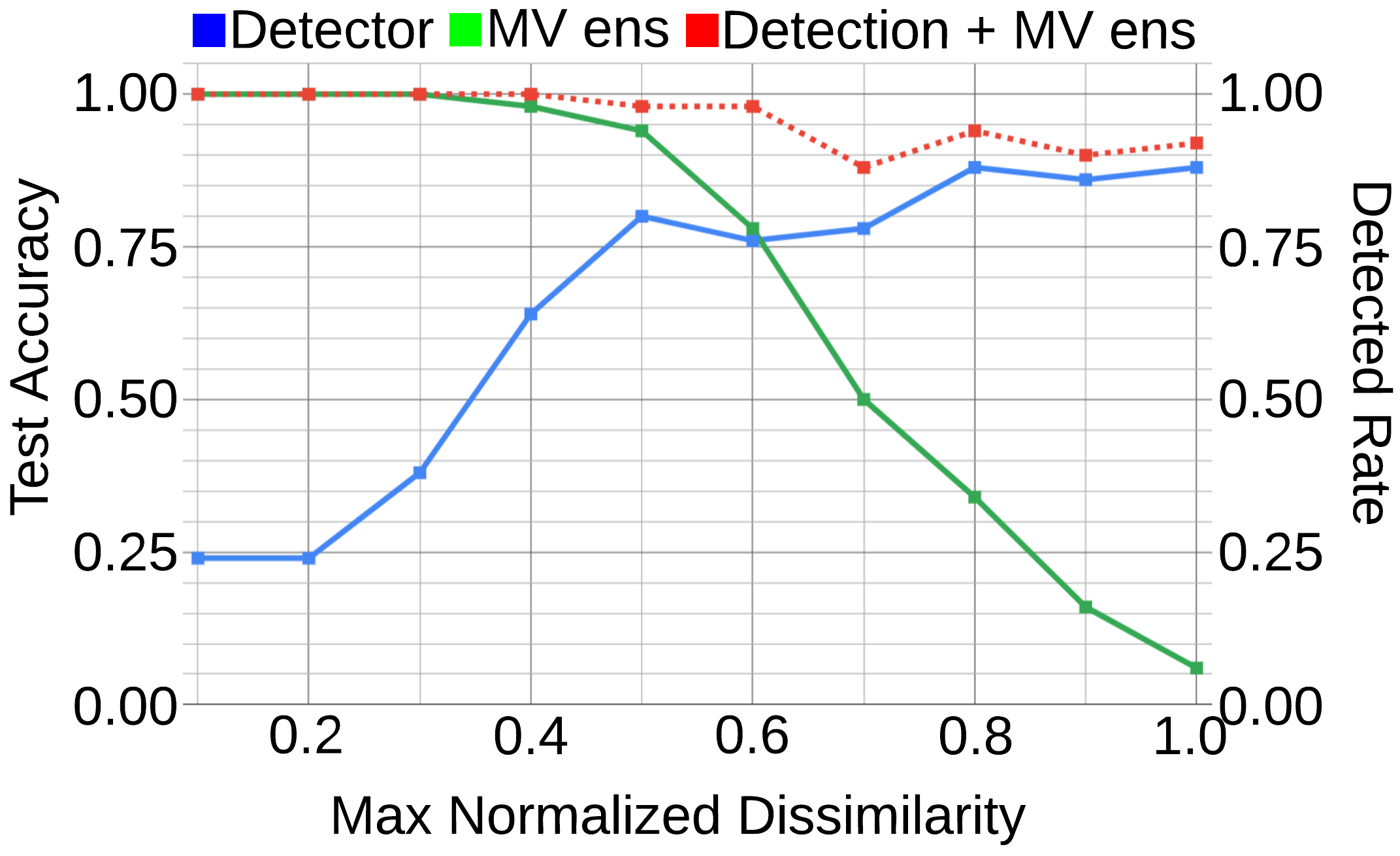}
    \caption{Evaluation results for White-box threat models by detector, MV ensemble, and a simple supreme model combining the detector and the MV ensemble (cf., detection + MV ens): (i) percentage of detected adversarial examples by detector (blue lines), (ii) test accuracy of the MV ensemble (green lines), and (iii) as compared to the maximum normalized dissimilarity of AE sets.}
    \label{fig:detection_wb_mnist}
\end{figure}

As shown in Figure~\ref{fig:errrate_wb_cifar100}, with modest effort, the attacker is able to generate less costly AEs. However, such AEs are not strong enough to fool the MV ensemble. For example, it takes on average less than 2 seconds to generate a single AE with $\text{max\_dissimilarity}=0.05$, which is $10$x of crafting an AE in zero-knowledge threat model. However, only $7.4\%$ of the generated AEs were misclassified by \ourframework. On average, each of the AEs can only fool $5\%$ ($1/22$) of WDs.
As $\text{max\_dissimilarity}$ increases, the attacker is able to generate stronger AEs. For example, with $\text{max\_dissimilarity}=0.275$, $71.8\%$ of the generated AEs successfully fooled the MV ensemble. Each of such AEs, on average, is misclassified by $81.82\%$ ($18/22$) of the WDs, making MV ensembles ineffective. However, this comes at a price:
\begin{itemize}
    \itemsep0em
    \item \textit{High cost}: The time cost for generating a single AE is high, for example 34 seconds for the strongest AE, i.e., $\text{max\_dissimilarity}=0.275$, which is $170$x of the cost for generating an AE in zero-knowledge. For MNIST with 72 WDs, it takes on average $610$ seconds to generate an AE with $\text{max\_dissimilarity}=1.0$, which is $310$x of generating an AE in zero-knowledge model (cf. Figure~\ref{fig:cost_wb_mnist}). This provides a tradeoff space, where realizations of \ourframework~that employ larger ensembles incur more cost to the attackers. 
    \item \textit{Easily detectable}: Furthermore, although such AEs successfully fool the MV ensemble, they were distorted heavily\footnote{The average normalized $l_2$ dissimilarity of AEs ($0.2192$) is larger than that of the strongest \FGSM~in zero-knowledge model ($0.1963$).} and very likely to be detected either by a human (see Figure~\ref{fig:cifar100_wb_greedAE}) or an adversarial detector. We adapted a adversarial detector~\footnote{rfeinman: github.com/rfeinman/detecting-adversarial-samples} and tested on MNIST. As presented in Figure~\ref{fig:detection_wb_mnist}, the detector is able to successfully detect $89.8\%$ of AEs for the white-box. An enhanced version of \ourframework~with the detector (Detection + MV ensemble) can achieve a high accuracy of above $90\%$ on AEs with any strength by either detecting and/or recovering the correct label via WDs. 
\end{itemize}

\begin{figure}[t]
    \captionsetup[subfigure]{font=scriptsize,labelfont=scriptsize}
    \scriptsize
    \centering
    \subfloat[Time cost (seconds)]{
        \includegraphics[width=0.48\linewidth]{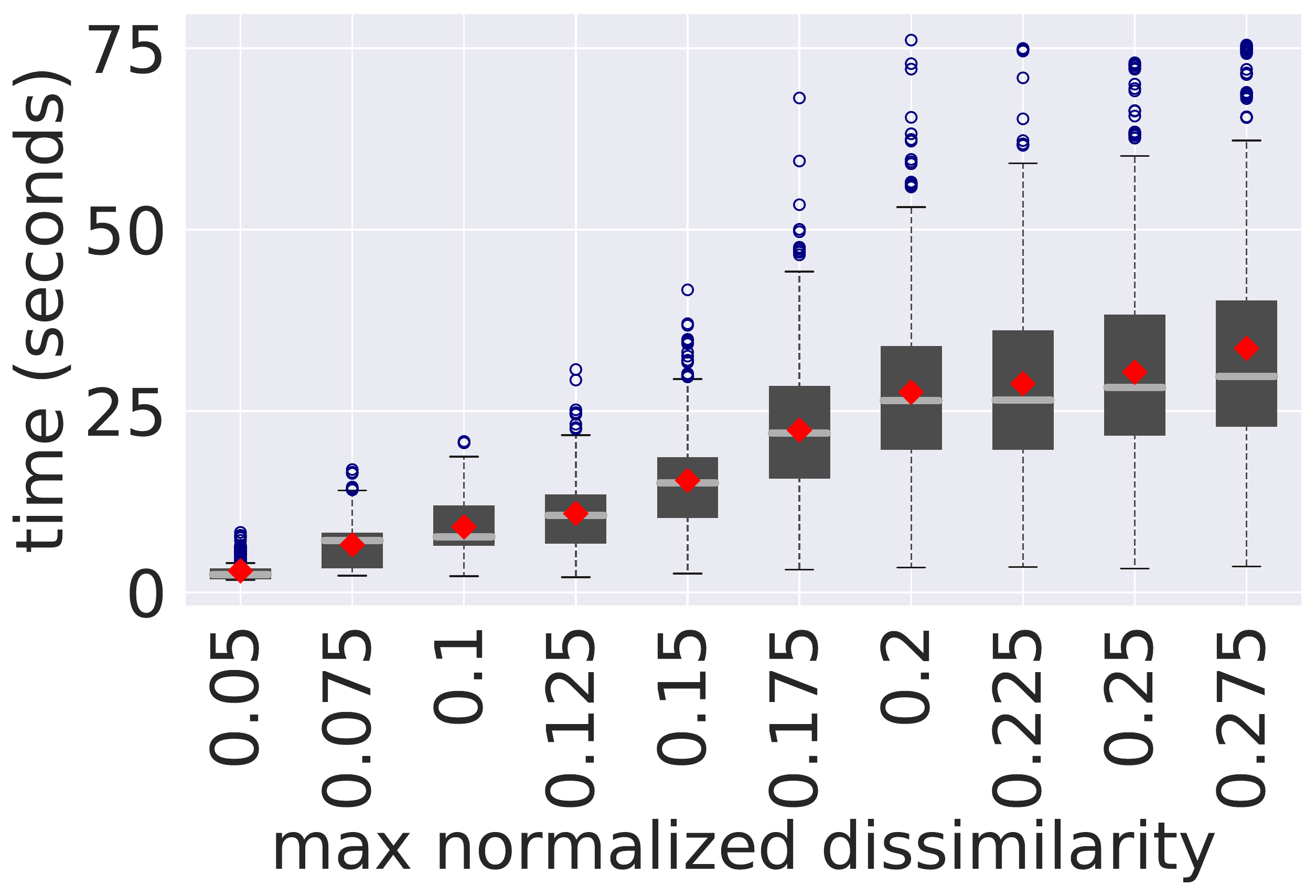}
    } 
    \subfloat[Normalized $l_2$ dissimilarity]{
        \includegraphics[width=0.48\linewidth]{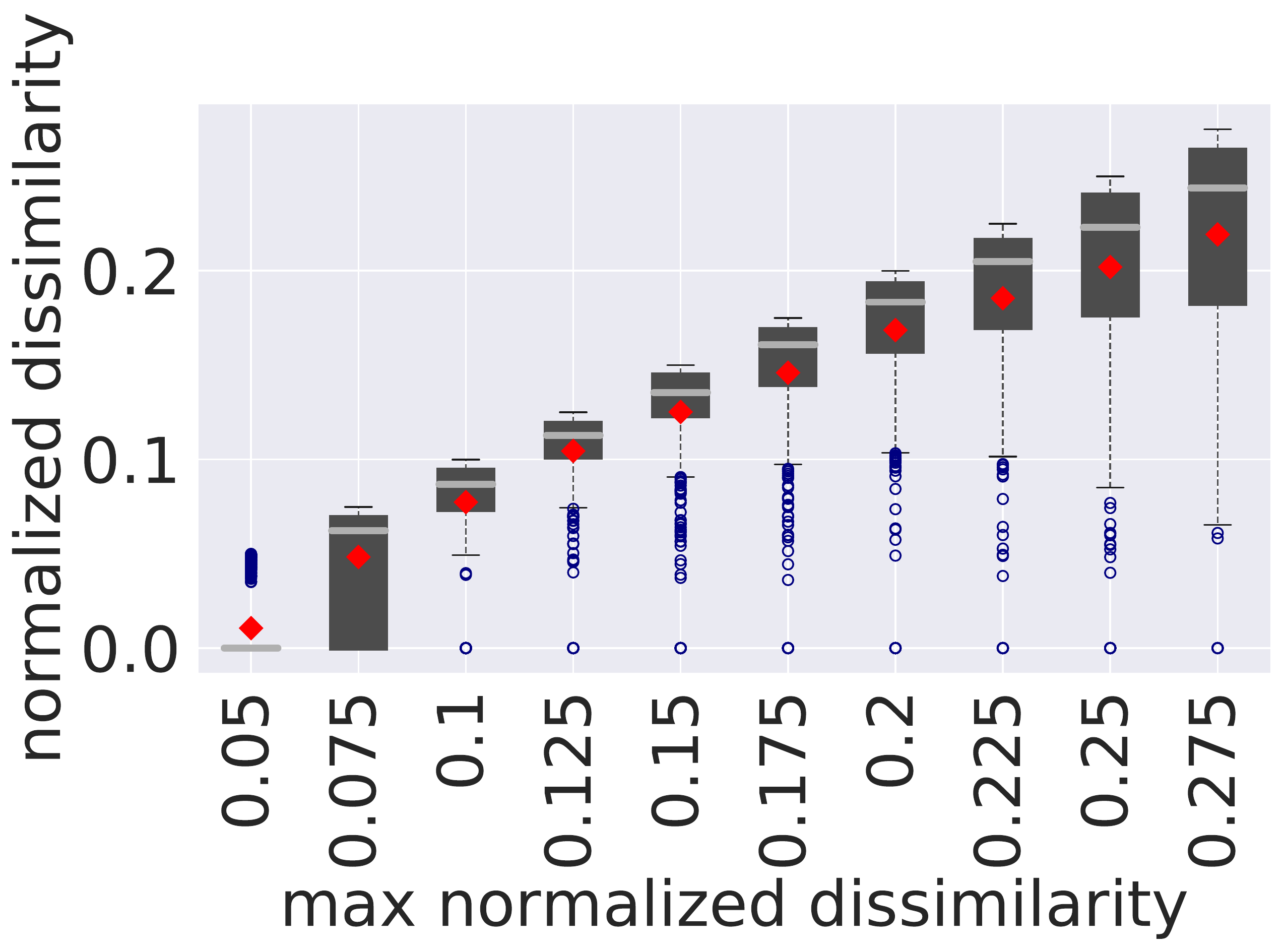}
    } \\
    \subfloat[Number of iterations (Algorithm~\ref{alg:attack_white-box})]{
        \includegraphics[width=0.48\linewidth]{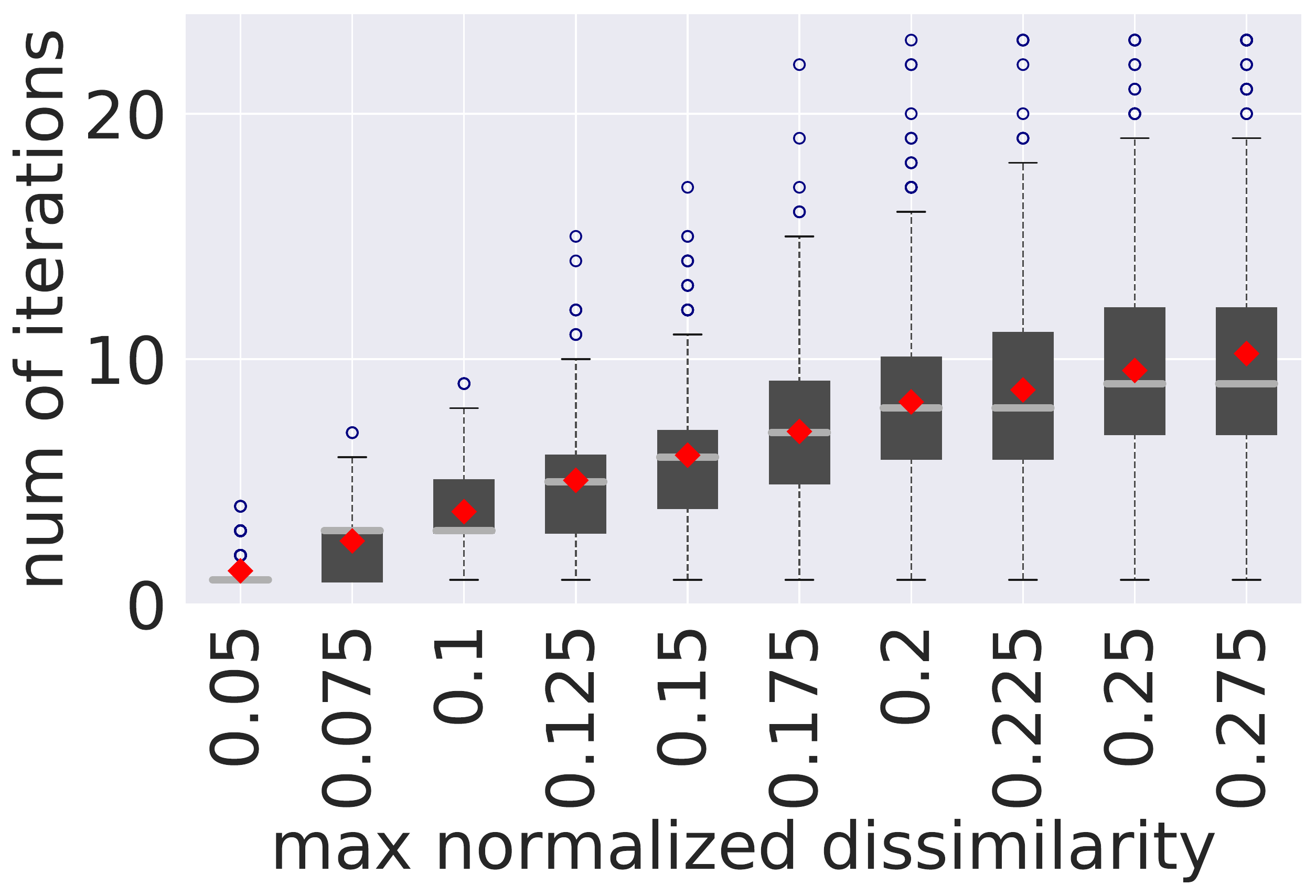}
    }
    \subfloat[Number of WDs being fooled]{
        \includegraphics[width=0.48\linewidth]{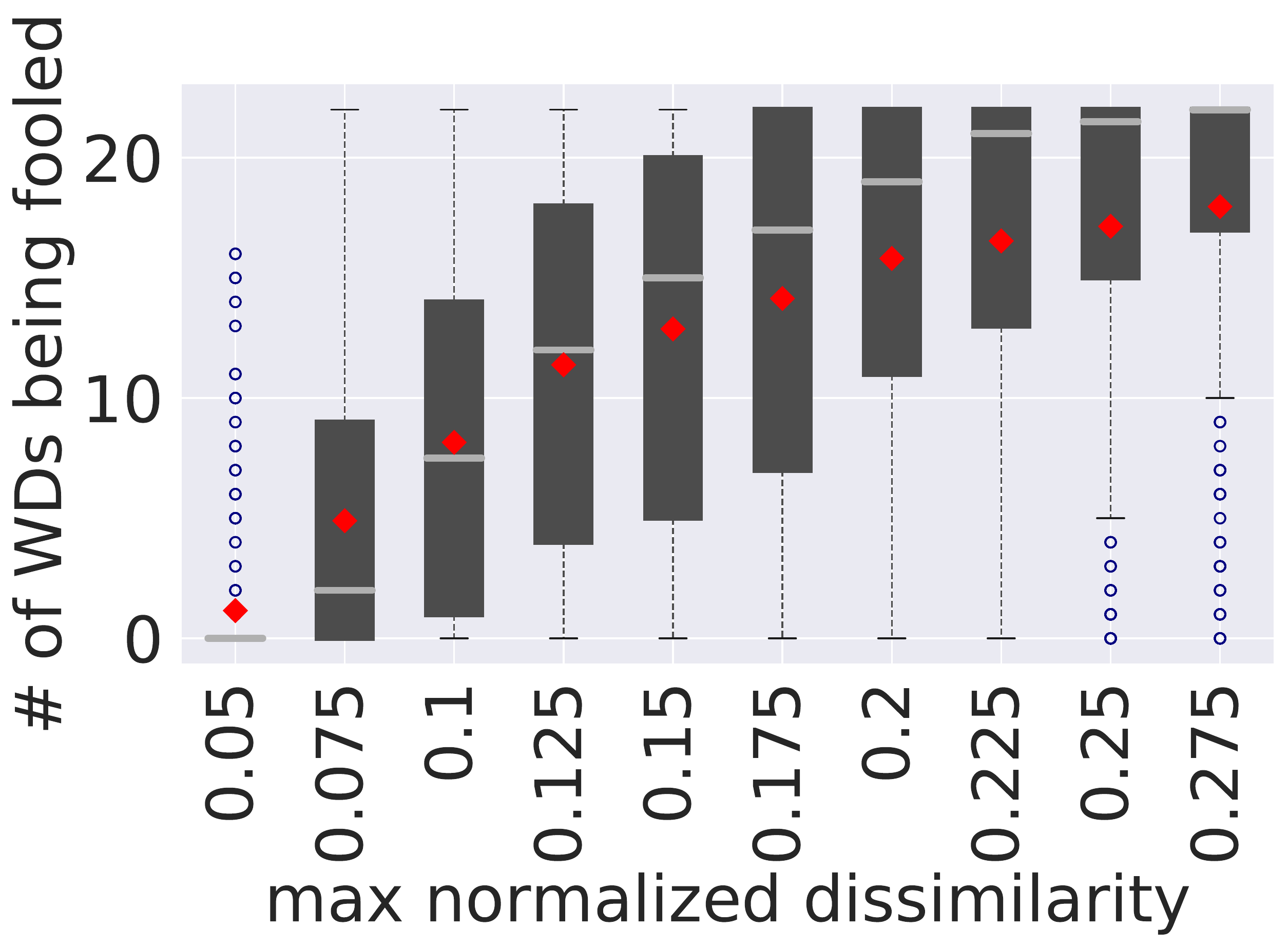}
    }
    \caption{Cost of AEs generation using the greedy attack on CIFAR-100.}
    \label{fig:cost_wb_cifar100}
\end{figure}

\textbf{The optimization-based approach} synthesizes AEs based on an ensemble model by combining \emph{Expectation Over Transformation} (EOT)~\cite{athalye2017synthesizing} with Random Self-Ensemble (RSE)~\cite{Liu2018_self-ensemble}. 
EOT is a general-purpose algorithm for generating AEs that are robust over the chosen distribution of transformations ($T$):
\begin{equation}\label{eq:EOT_attack}
    \begin{split}
        x' = \argmax_{x'} \mathbb{E}_{t\sim T}[\log f(y_t|t(x')] \\
        \text{s.t. } \mathbb{E}[d(t(x'), t(x)] < \epsilon, x \in~[0, 1]^d,
    \end{split}
\end{equation}
which extends the standard attack in~\eqref{attack}. We extend EOT for ensembles,\footnote{For an ensemble of size one, this extended attack will be reduced to EOT.} generating AEs by averaging the EOTs~\cite{Liu2018_self-ensemble} of all the WDs known to the attacker:
\begin{equation}\label{eq:EEOT_attack}
    \begin{split}
     \argmax_{x'} \frac{1}{K}\Sigma_{i=1}^{K}\mathbb{E}_{t\sim T}[\log f_i(y_t|t(x')] \\
     \text{s.t. } \mathbb{E}_{t \sim T}[d(t(x'), t(x))] < \epsilon, x \in [0, 1]^d,
    \end{split}
\end{equation}
where $d(\cdot, \cdot)$ is a distance function, $T$ is a chosen distribution of transformation functions $t(\cdot)$, $f(\cdot)$ is the targeted model
, and $f_i(\cdot)$ is one of the $i\in N$ WDs in the ensemble. Based on this extended approach, a gray-box attack is when the adversary has white-box access to $K$ WDs in the ensemble, where $K < N$, and a white-box attack is when the adversary has white-box access to the full list of WDs, where $K = N$. 



\begin{figure}[t]
    \captionsetup[subfigure]{font=scriptsize,labelfont=scriptsize}
    \scriptsize
    \centering
    \subfloat[Error rate]{
        \includegraphics[width=0.32\linewidth]{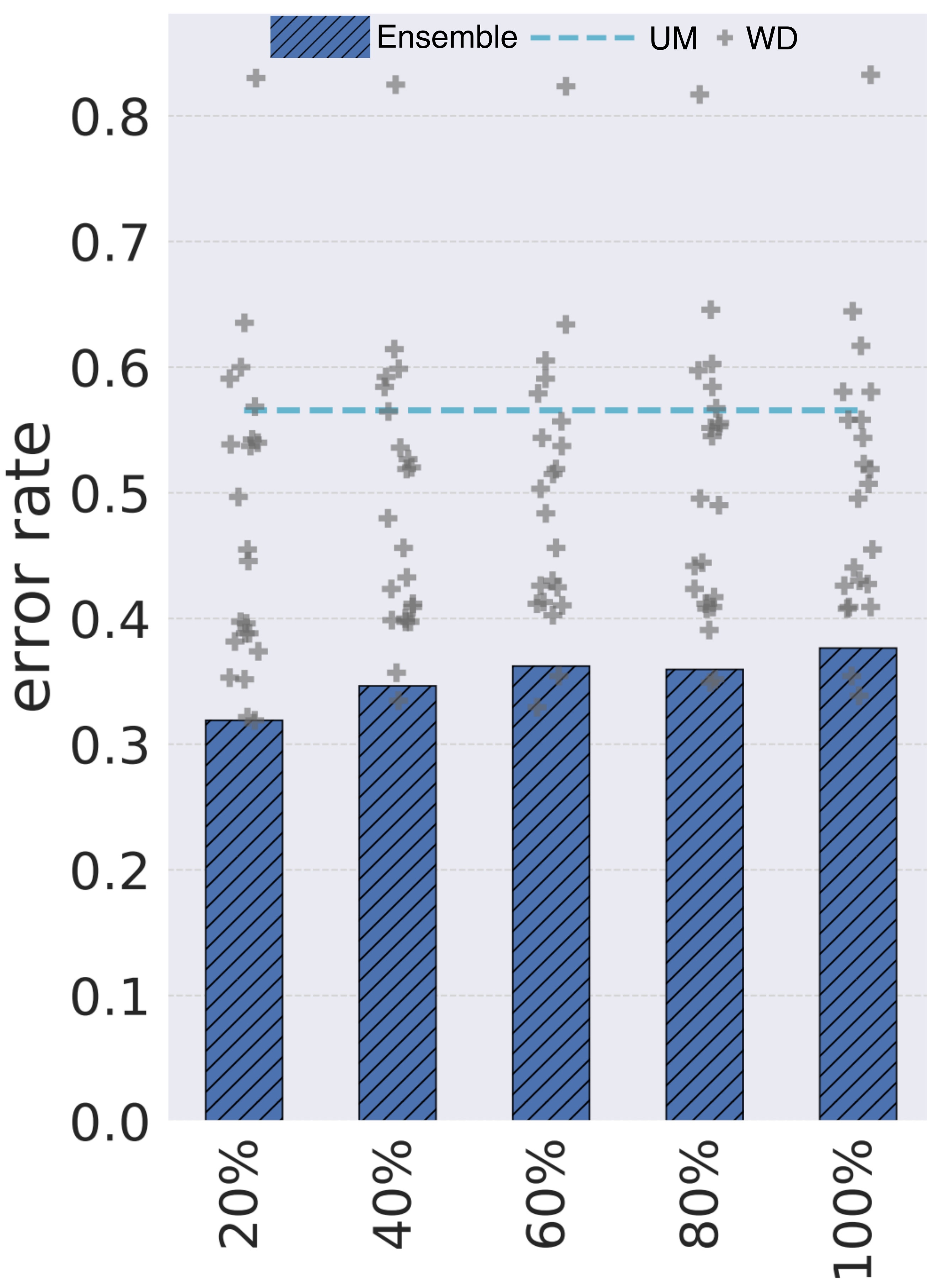}
        \label{subfig:synthesis_cifar100_err}
    }
    \subfloat[Computational cost]{
        \includegraphics[width=0.32\linewidth]{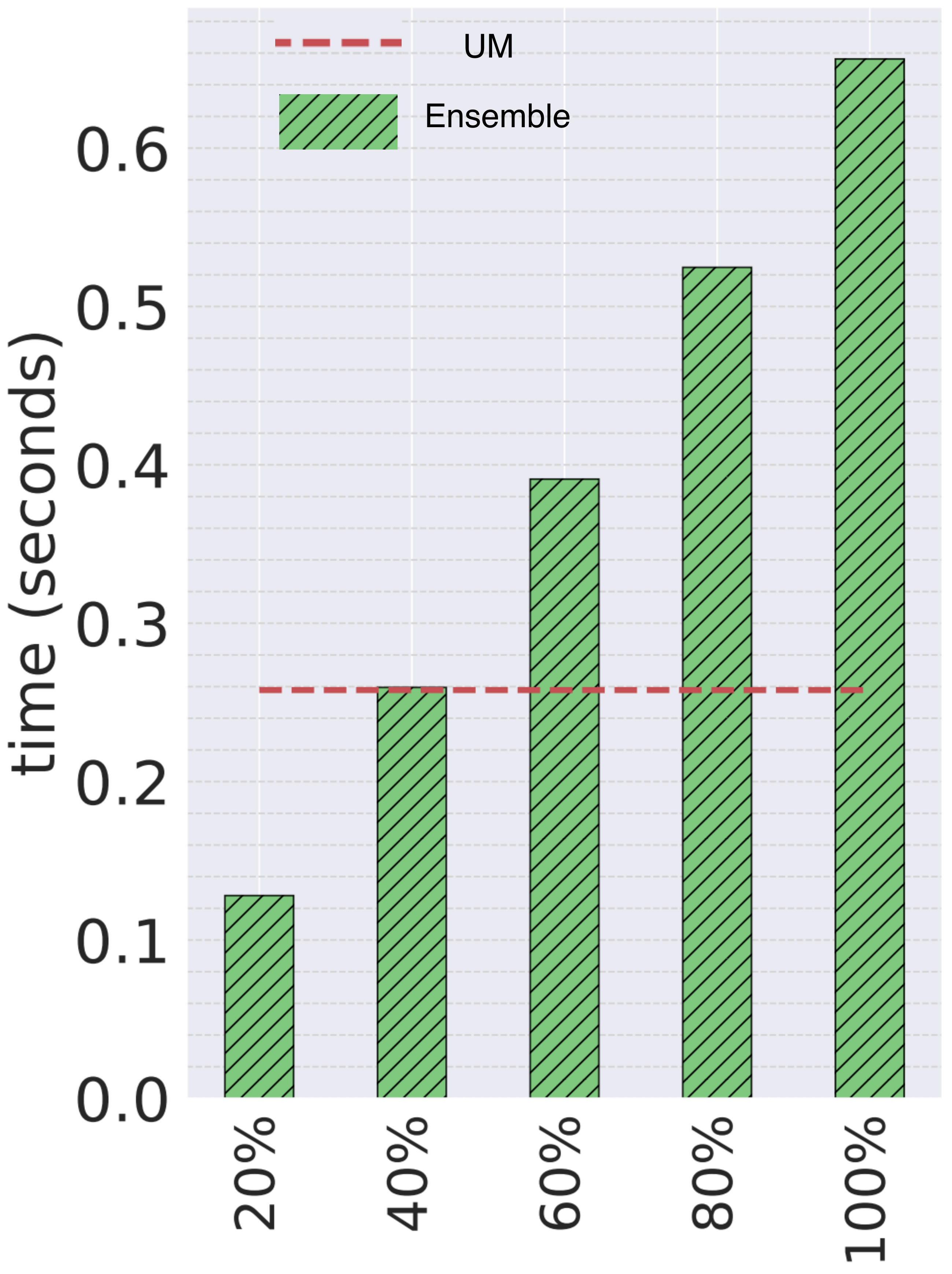}
        \label{subfig:synthesis_cifar100_cost}
    }
    \subfloat[Normalized $l_2$ distance]{
        \includegraphics[width=0.33\linewidth]{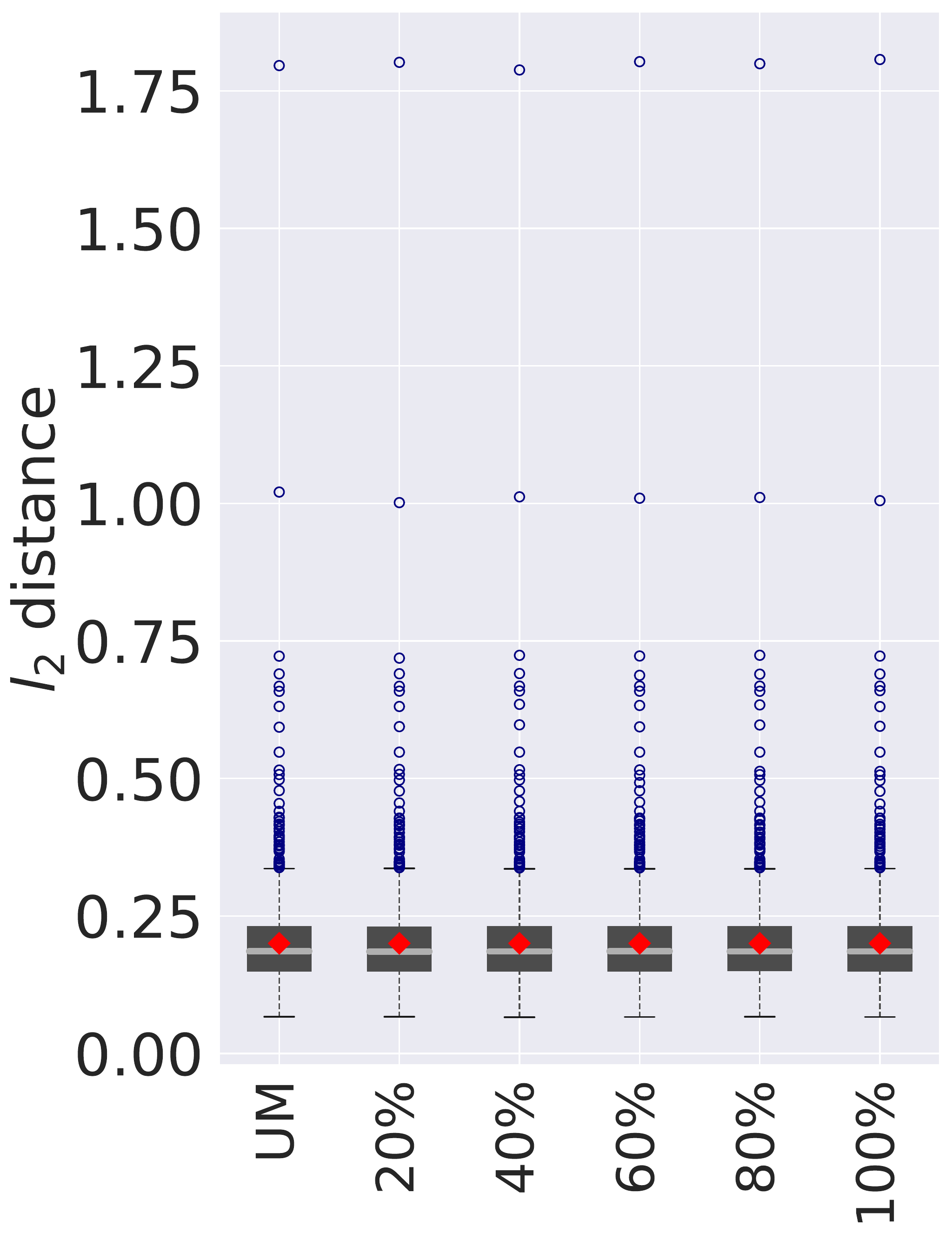}
        \label{subfig:synthesis_cifar100_dist}
    }
    \caption{Evaluation of AVEL ensemble against the optimization-based white-box attack on CIFAR-100.} 
    \label{fig:eval_synthesized_ae_cifar100}
\end{figure}

For simplicity, we consider the distribution of transformations consist of rotations (a random angle between $-20$ to $20$ degrees), adding Gaussian noise, and translation (a random offset between $-25\%$ to $25\%$ on x- and/or y-axis). We evaluated several variants by evaluating an input over $30$, $100$, and $500$ randomly sampled transformations for individual WD. We generated $10$ variants on MNIST: 9 in the gray-box model, with $10\%$ to $90\%$ of the WDs accessed by the attacker and 1 variant in the white-box model with all WDs are known to the attacker. For CIFAR-100, we generated 5 variants: 4 variants in the gray-box model, with $20\%$, $40\%$, $60\%$, and $80\%$ of the WDs are known to the attacker and 1 variant in the white-box scenario. For comparison, we used EOT to generate AEs based on the UM. The evaluation results for CIFAR-100 are shown in Figure~\ref{fig:eval_synthesized_ae_cifar100}, and for MNIST are shown in Figure~\ref{fig:err_synthesis_mnist} and Figure~\ref{fig:eval_synthesized_ae_mnist}.



As expected, as the adversary becomes more white-box (i.e., having access to more WDs), it can launch more successful attacks (Figure~\ref{fig:eval_synthesized_ae_cifar100}~\subref{subfig:synthesis_cifar100_err}) without even increasing the perturbations (Figure~\ref{fig:eval_synthesized_ae_cifar100}~\subref{subfig:synthesis_cifar100_dist}). However, the cost of generating AEs is increased (Figure~\ref{fig:eval_synthesized_ae_cifar100}~\subref{subfig:synthesis_cifar100_cost}). The experimental results show that \ourframework can even blocks adversarial attacks from a strong adversary specifically designed to target \ourframework. The attacker has obviously the choice to sample more random transformations and also chooses a distribution of a large number and diverse transformations to launch stronger attacks. However, this will incur a large computational cost to the attacker.

\section{Research Questions}\label{sec:research questions}
As shown in our extensive evaluations, \ourframework~is effective against a wide variety of adversaries from the weakest threat model (zero-knowledge) to the strongest one (white-box). 

Here we wish to address the following questions:
\begin{enumerate}[1)]
    \itemsep0em
    \item \textbf{RQ1}: Why do the quantity and diversity of WDs matter?
    \item \textbf{RQ2}: Does our defense generalize across different models?
    \item \textbf{RQ3}: What is the overhead of \ourframework?
\end{enumerate}

\section{RQ1. Diversity of Weak Defenses}\label{subsec:diverse_cnn_wds}

\textbf{Study Design.} 
Here we investigate how the number of WDs and the diversity relates to an ensemble's effectiveness for countering adversarial attacks.

Several metrics  for ensemble diversity have been proposed, e.g., diversity among the non-maximal predictions of individual members (cf. WDs in \ourframework) in the ensemble~\cite{Pang2019_EnsembleDiversity}, or negative correlation learning in training individual members to encourage different members to learn different aspects of the training data~\cite{Islam2003_ConstructiveAlg, Dietterich2000_EnsembleMethods, Liu1999_NegativeCorrelation, Liu1999_SimuTrainingNegCorrelation}.  Fiversity on the non-maximal predictions of individual WDs promotes the ensemble's diversity at the training phase by serving as a penalty in adaptive diversity promoting (ADP) regularizer~\cite{Pang2019_EnsembleDiversity}. 
Since we have trained individual WDs without such regularizers, we could not use these metrics in this work, and in particular, we have measured the diversity of several ensembles in \ourframework and the diversity of the non-maximal predictions was low as expected. Incorporating such regulaizers during the training of WDs is a potential future direction of this work. 


In this work, based on our empirical insights that were discussed in Section~\ref{subsec:trans_as_def}, we defined a new diversity metric by considering both the individual WDs performance and how they complement each other in an ensemble:
\begin{equation} \label{eq:diversity}
 \psi = ( \min_{i\in \{1,\dots,K\}}{|S_i|}) - (|\bigcap_i S_i|),
\end{equation}
where $S_i$ is the set of examples correctly predicted by $\text{WD}_i$ on dataset $D$ and $|\cdot|$ is the cardinality of $S_i$. The intuition behind the proposed diversity metric is that it encourages the effectiveness of individual WDs via $\min|S_i|$ and a wide diverse coverage on correctly predicted inputs by requiring small overlaps on correctly predicted samples via $|\bigcap_i S_i|$. Note that $\psi$ is always positive for any ensemble defense. 

\begin{figure}[t]
    \small
    \centering
    \includegraphics[width=1.\linewidth]{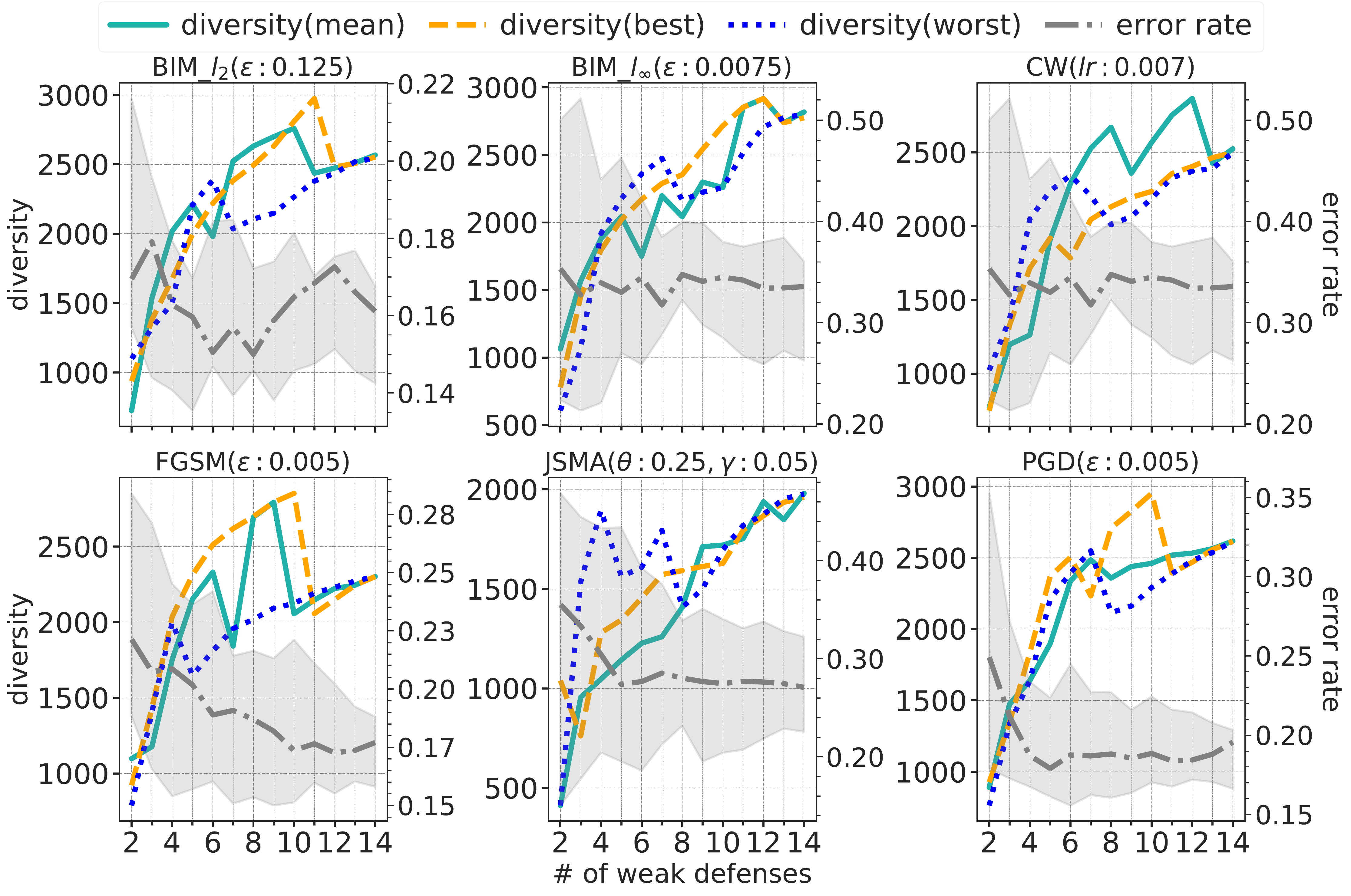}
    \caption{Ensemble diversities and error rates of best, worst, and median performers on CIFAR-100.}
    \label{fig:diversity_cifar100}
\end{figure}

To investigate the impacts of an ensemble's size and diversity on the effectiveness, we first created AVEL ensembles with 2 to 14 random WDs from a library of 22 WDs and then evaluated the ensemble's effectiveness (in terms of the error rate) against various attacks and computed the corresponding ensemble diversities. We repeated the experiment 5 times on CIFAR-100 and reported the results for the best, worst, and the mean performers over the 5 trials in Figure~\ref{fig:diversity_cifar100}. Experiments were also performed on MNIST with the size varying from 2 to 11 random WDs from a library of 72 WDs, and the experiment results are presented in Figure~\ref{fig:ensemble_diversity_mnist}.

\textbf{Results.}
The results indicate that as more WDs are incorporated, the ensembles tend to become more diverse (according to the metric defined in \eqref{eq:diversity}), and the ensembles in the worst case scenario\footnote{The focus on worst case scenario is by intention as there could be large number of potential ensembles. Given a transformation library of size $K$, we have $\binom{n}{K}$ choices for constructing an ensemble defense of size $n$.} (i.e., see the upper bound of error rates) become more effective to adverse adversarial attacks.  For example, in \CW$(\text{lr}: 0.007)$, the worst performer's error rate declines from 50\% to 26\%. Note that this experiment is by no means comprehensive, as it only incorporates one diversity metric and a specific library of transformations. We leave such comprehensive study of the impact of diversity to ensemble-based defenses as future work. However, the results indicate a promising direction for future research to develop search mechanisms to dynamically synthesize such ensemble-based defenses incorporating heuristics such as diversity metrics to construct effective ensembles to counter adversarial attacks. 



\begin{center}
\fbox{\parbox{0.97\columnwidth}{
\textbf{Summary.} A diverse set of weak defenses is necessary to build a robust ensemble defense against adversarial attacks.}}
\end{center}

\section{RQ2. Generality of \ourframework}\label{subsec:generality}


\textbf{Study Design.}
To evaluate how general \ourframework~is, we realized ensembles using different type of models and evaluated the ensembles in the context of zero-knowledge threat model. For the CIFAR-100 dataset, we built 4 ensembles with 22 WDs, associated with same list of transformations when using WRN classifiers, now using $26~2\times 32d$ ResNet with Shake-Shake regularization (ResNet-Shake). The undefended ResNet-Shake model achieved a test accuracy of $75.55\%$ on BS. The ensembles using ResNet-Shake models (ResNet-Shake \ourframework) were tested against 6 attacks in 2 configurations. For MNIST, we built 5 ensembles with $71$ Linear SVM classifiers, which achieved a test accuracy of $94.97\%$ on BS. The ensembles built with SVM models (SVM \ourframework) were against 7 different attacks with 3 configurations. 

\begin{figure}[t]
    \tiny
    \centering
    \subfloat[\FGSM]{
        \includegraphics[width=0.49\linewidth, trim={0.8cm 0.7cm 0.8cm 1.5cm}]{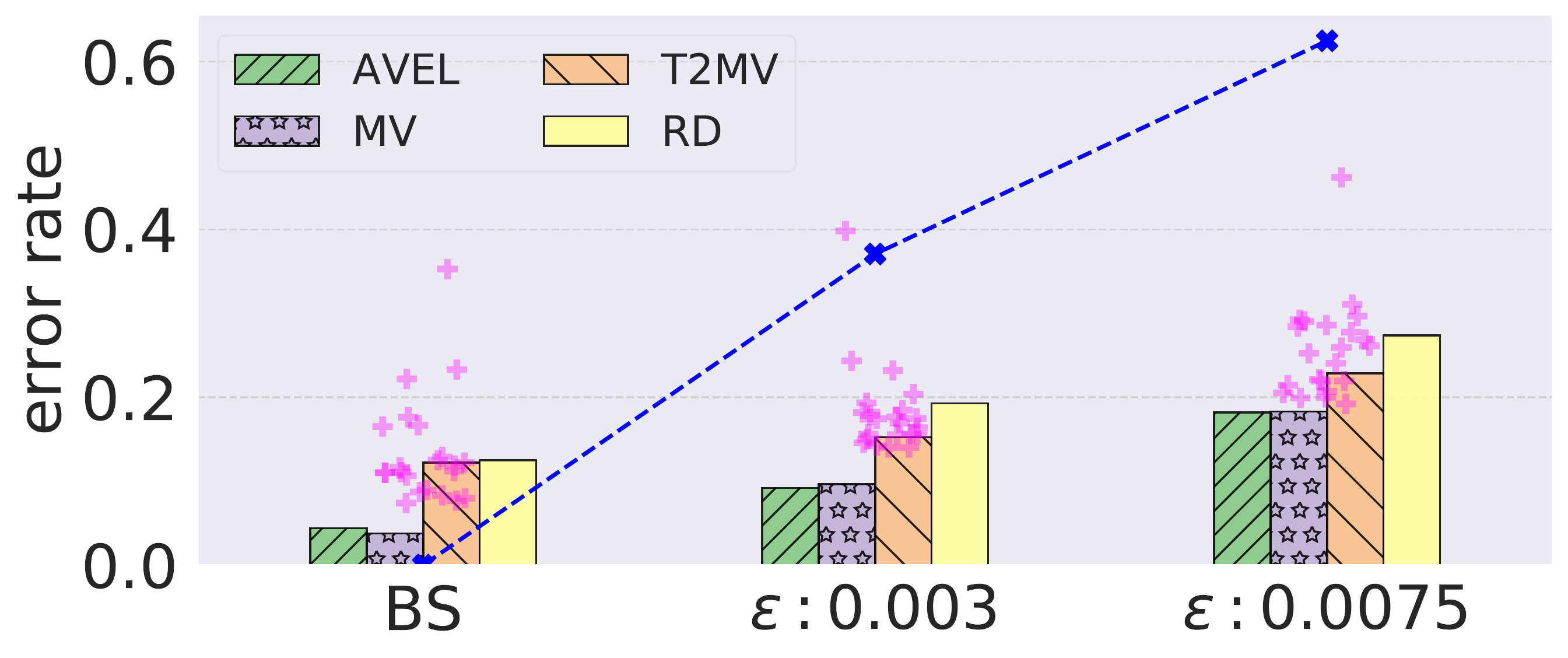}
    }
    \subfloat[\BIM\_$l_2$]{
        \includegraphics[width=0.49\linewidth, trim={0.8cm 0.7cm 0.8cm 1.5cm}]{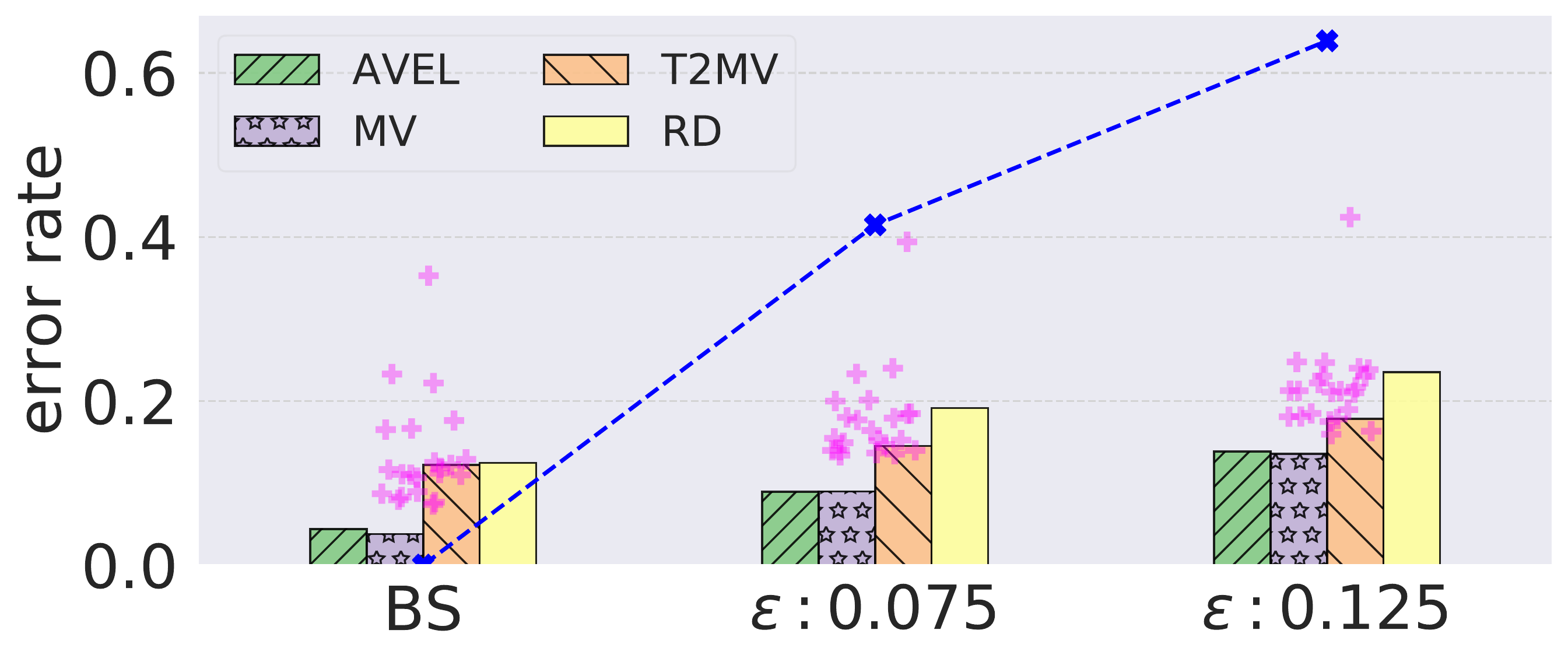}
    } \\
    \subfloat[\BIM\_$l_{\infty}$]{
        \includegraphics[width=0.49\linewidth, trim={0.8cm 0.7cm 0.8cm 1.5cm}]{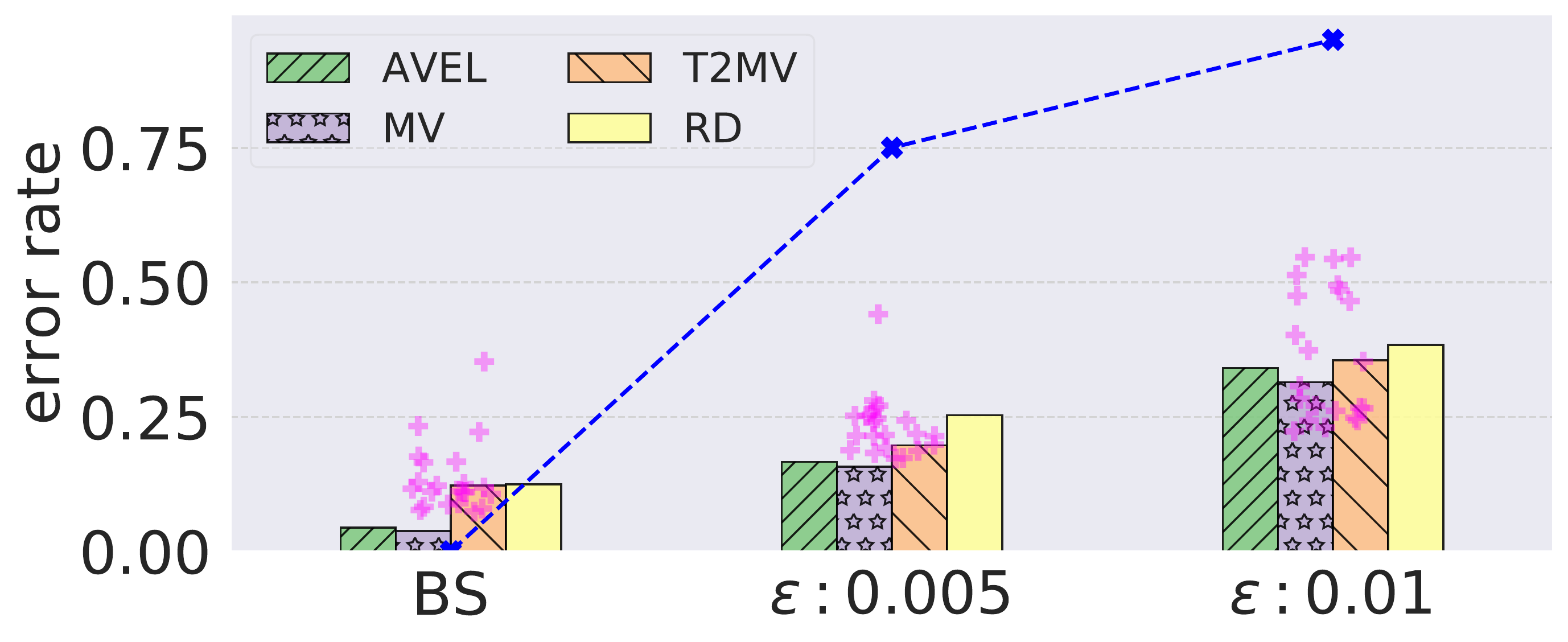}
    } 
    \subfloat[\CW\_$l_2$]{
        \includegraphics[width=0.49\linewidth, trim={0.8cm 0.7cm 0.8cm 1.5cm}]{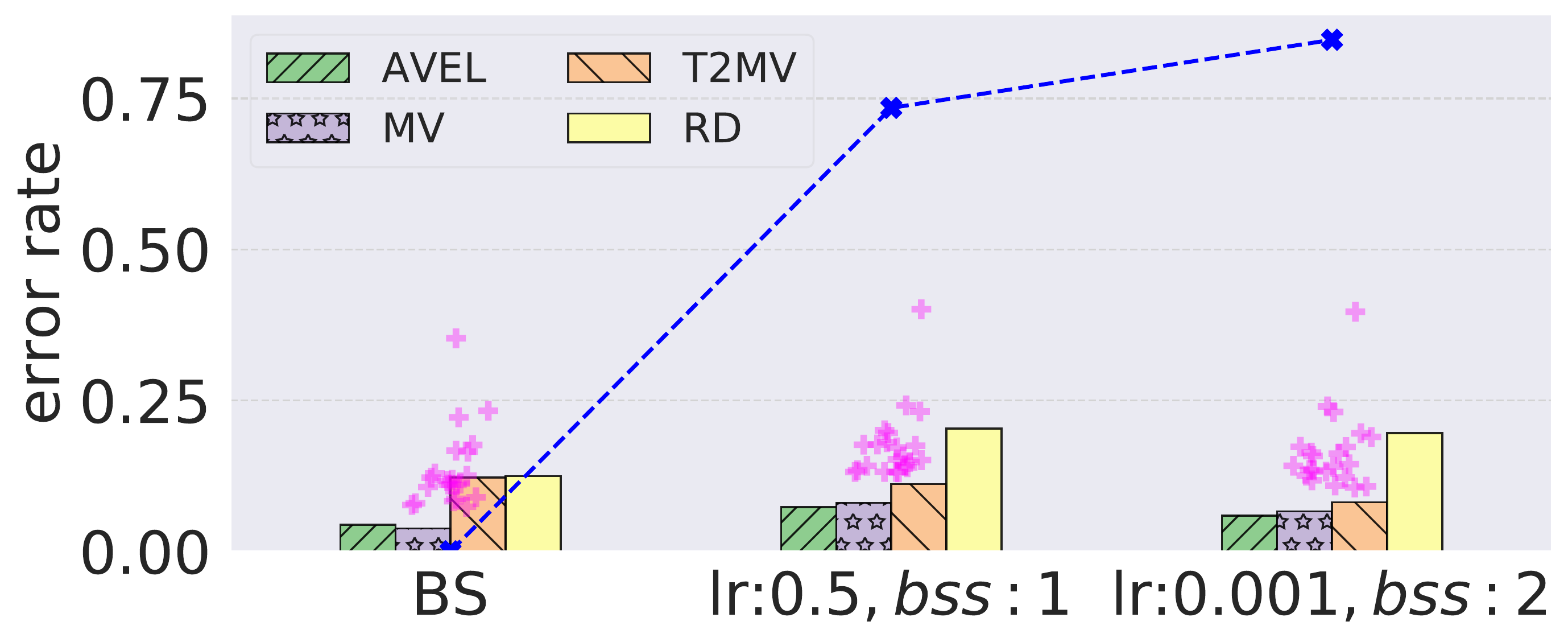}
    } \\
    \subfloat[\JSMA]{
        \includegraphics[width=0.49\linewidth, trim={0.8cm 0.7cm 0.8cm 1.5cm}]{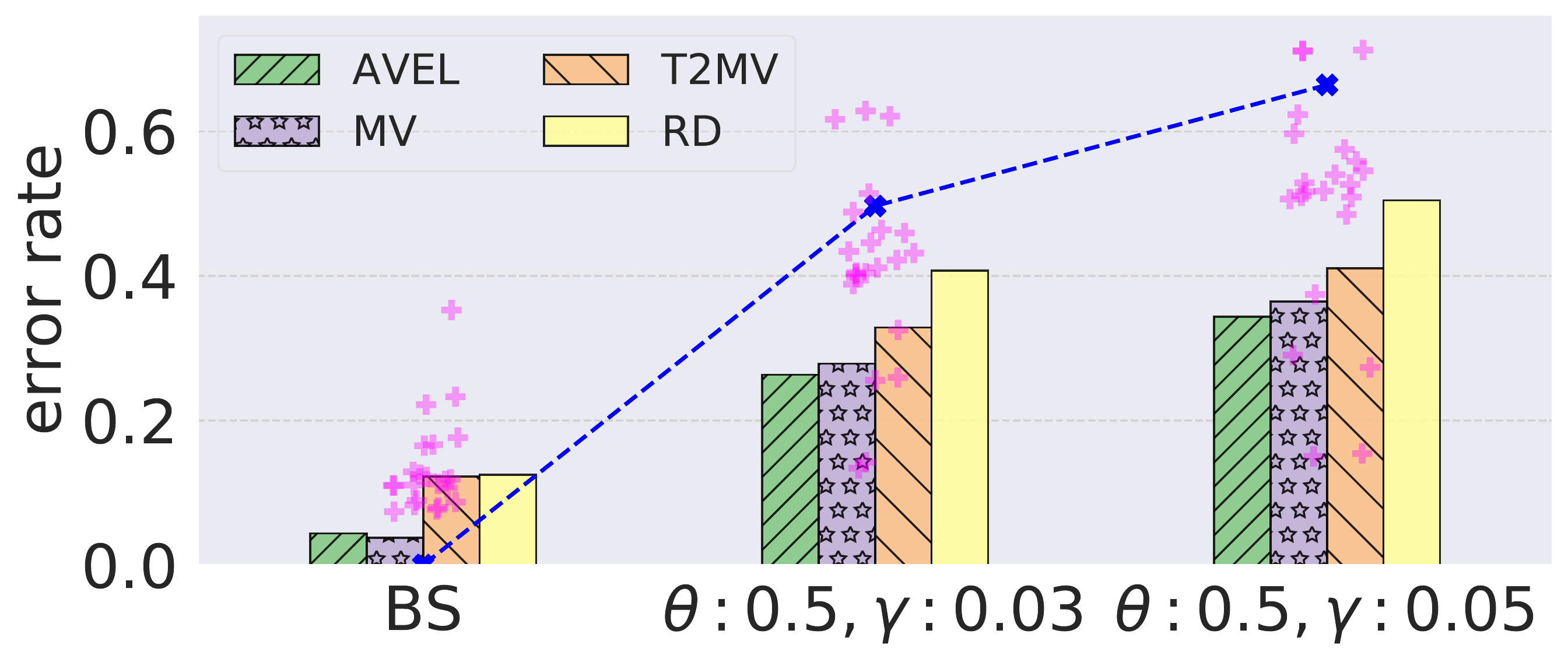}
    } 
    \subfloat[\PGD]{
        \includegraphics[width=0.49\linewidth, trim={0.8cm 0.7cm 0.8cm 1.5cm}]{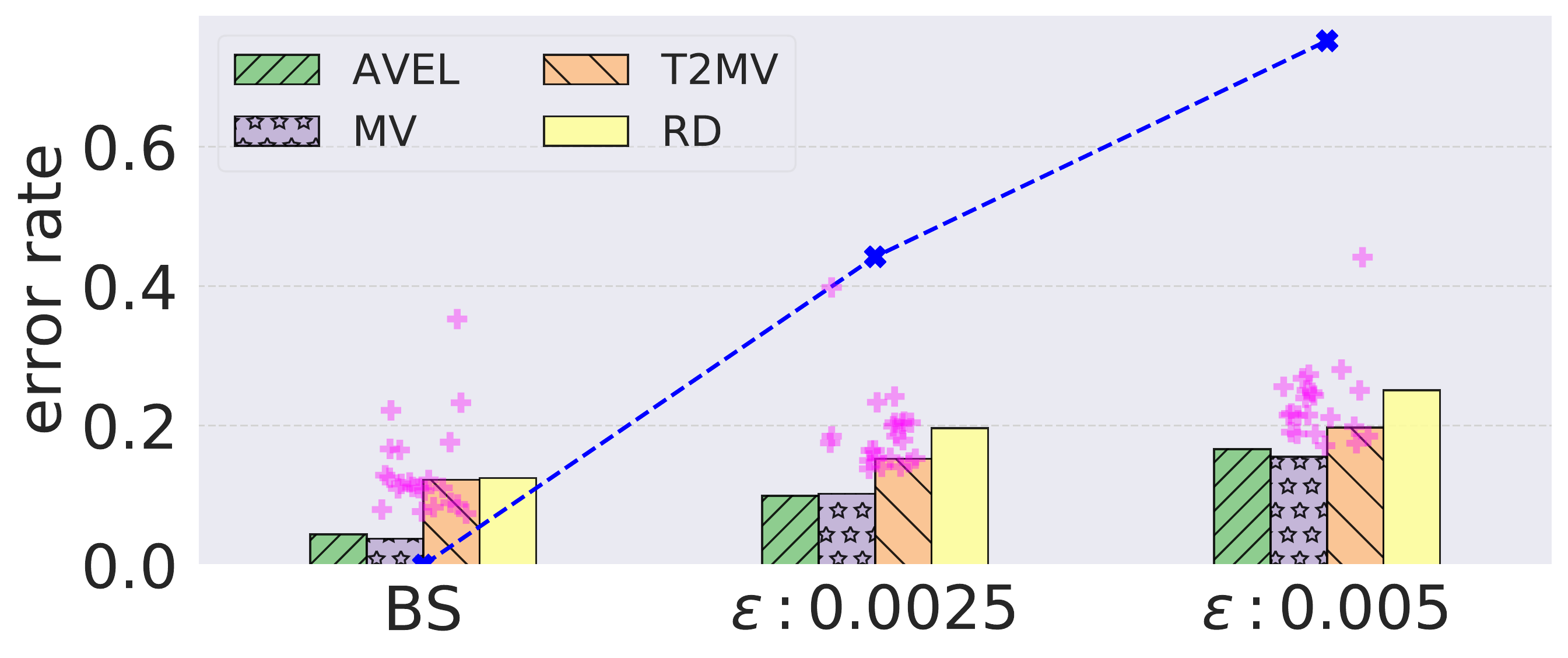}
    }
    \caption{\ourframework~using ResNet-Shake as weak defenses on CIFAR-100.} 
    \label{fig:eval_zk_cifar100-shake}
\end{figure}

\begin{figure}[t]
    \captionsetup[subfigure]{font=scriptsize,labelfont=scriptsize}
    \scriptsize
    \centering
    \scalebox{0.75}{
        \subfloat[Rank correlation across adversaries]{
            \includegraphics[width=0.9\linewidth]{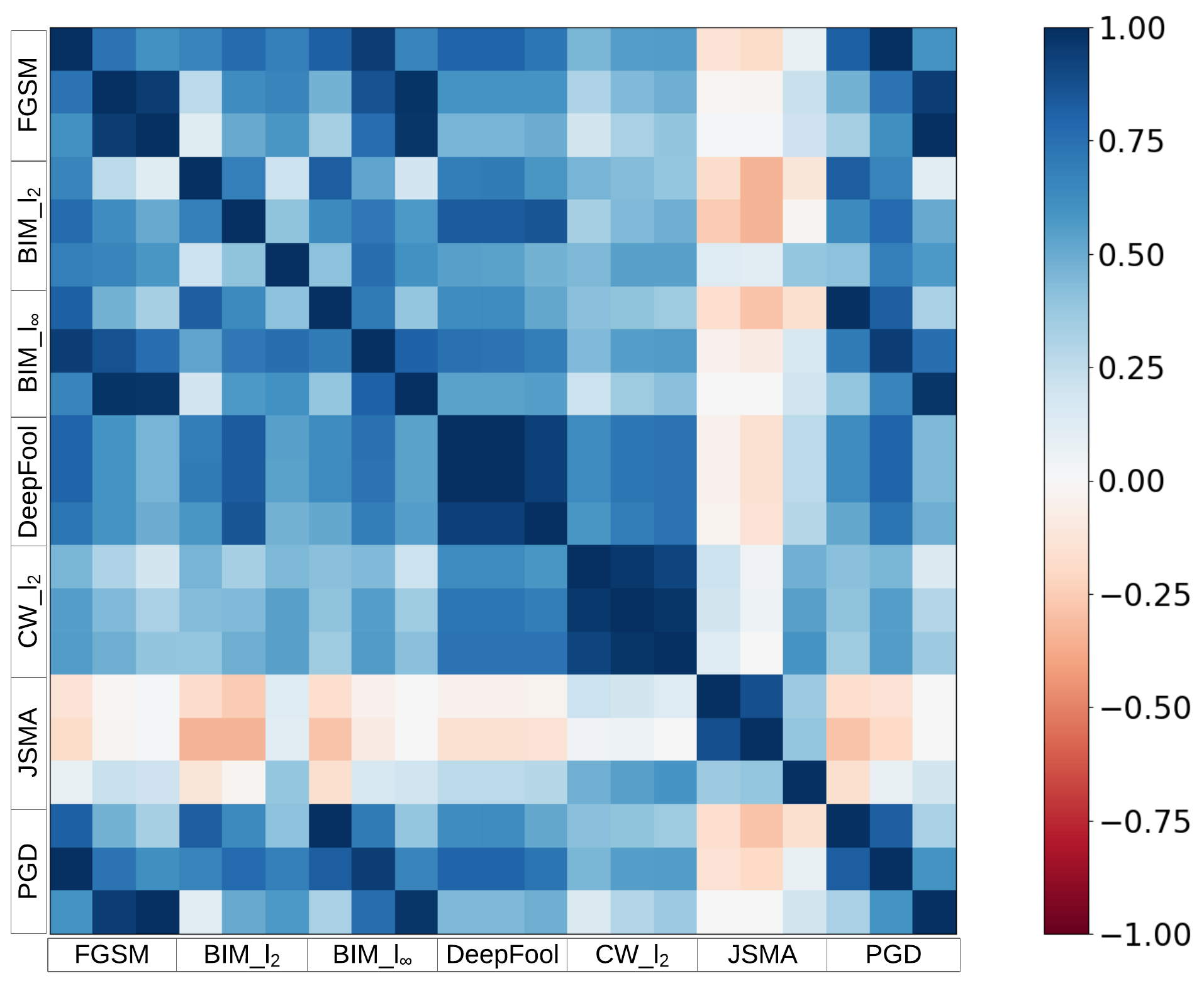}
            \label{subfig:rank_cross_svm_aes}
        }
    } 
    \\
    \scalebox{0.8}{
        \subfloat[Rank correlation across CNN and SVM models]{
            \includegraphics[width=0.75\linewidth]{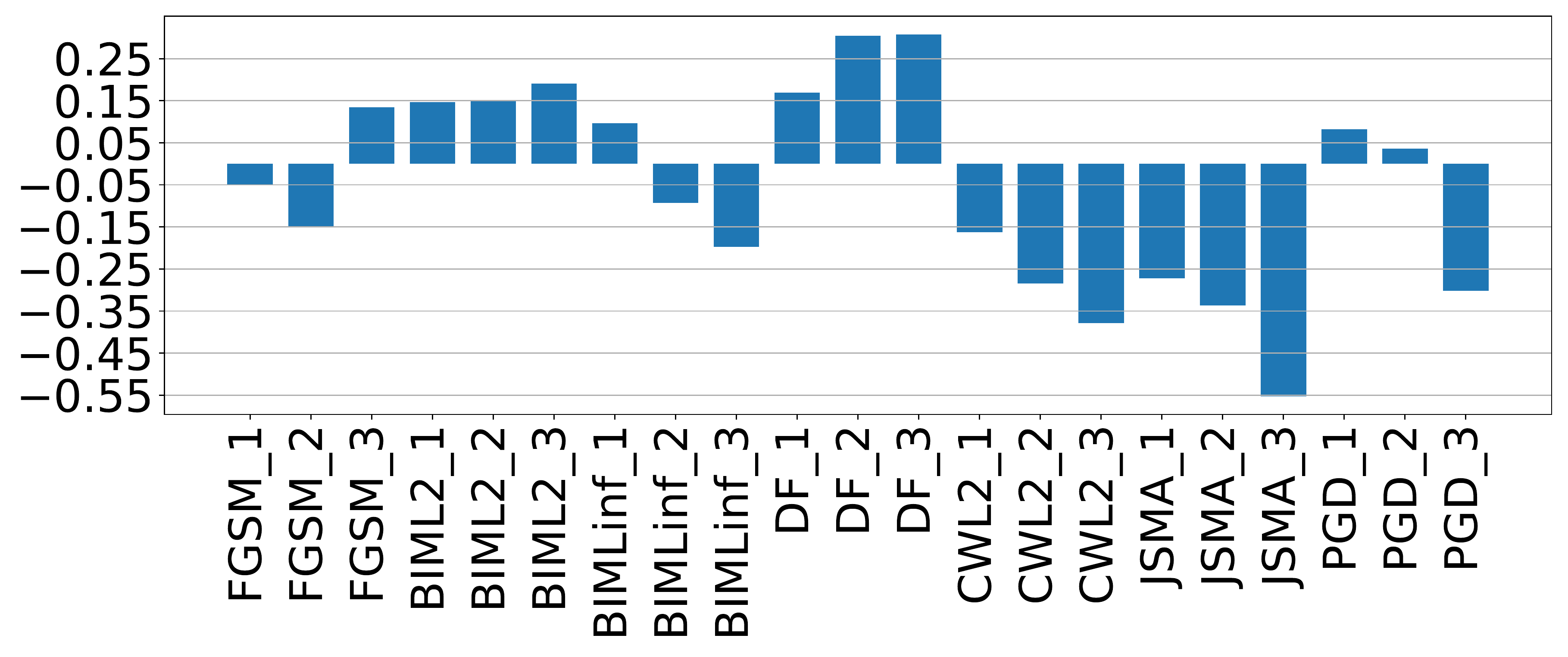}
            \includegraphics[width=0.16\linewidth]{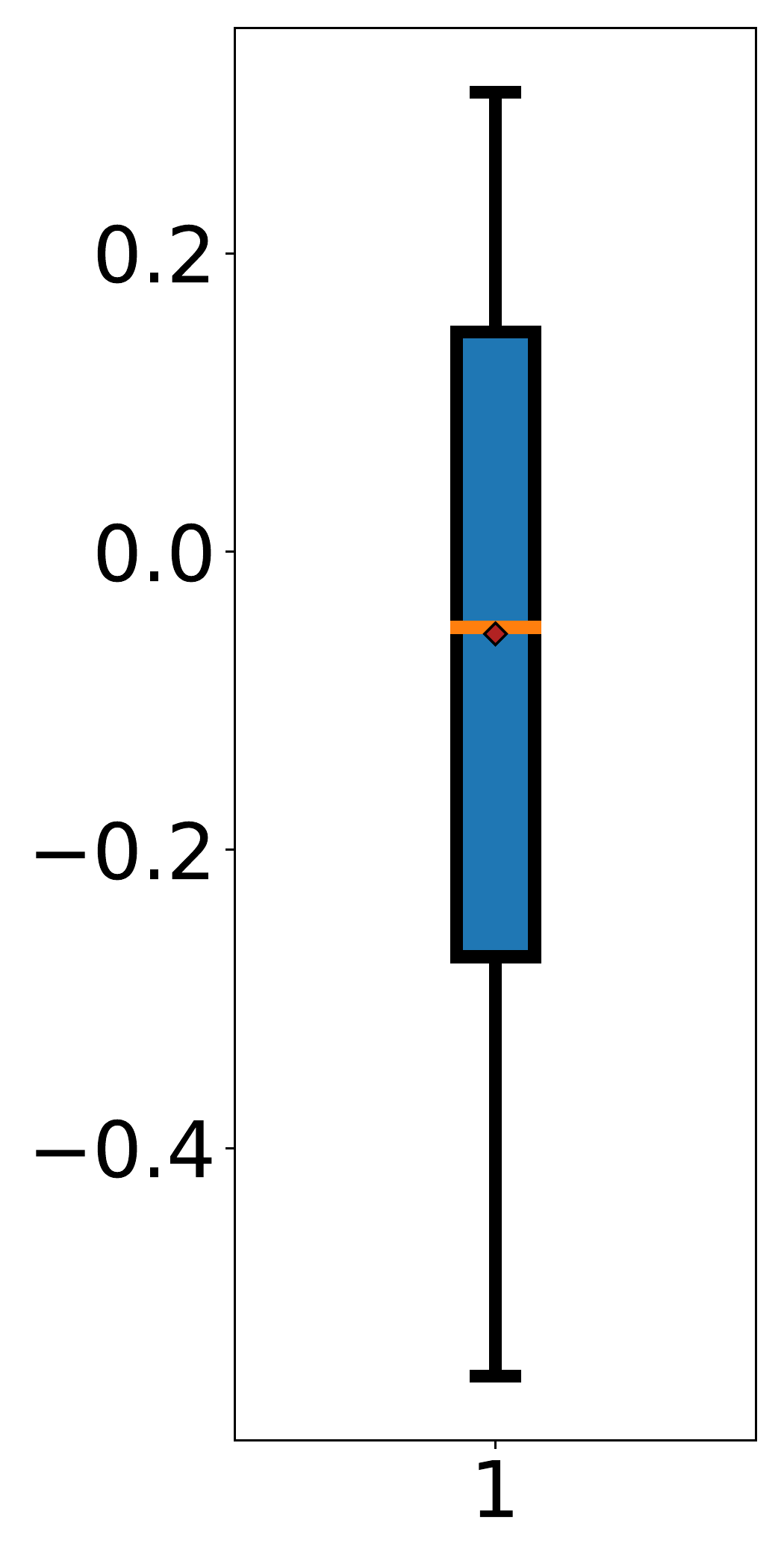}
            \label{subfig:rank_cross_models}
        }
    }
    \caption{Rank correlations of effectiveness of WDs (a) across AEs that were generated based on SVM UM (Settings of AEs follow the same order shown in Figure~\ref{fig:eval_zk_svm}), (b) across CNN models and SVM models.} 
    \label{fig:rank_corr_svm}
\end{figure}

\textbf{Results.}
The results are presented in Figure~\ref{fig:eval_zk_cifar100-shake} for CIFAR-100 (cf. Figure~\ref{fig:eval_zk_cifar100-wres}) and in Figure~\ref{fig:eval_zk_svm} for MNIST (cf. Figure~\ref{fig:eval_zk_mnist-cnn}). We made the following observations:
\begin{itemize}
    \itemsep0em
    \item Transformations are able to improve model robustness regardless of the type of model. Almost every individual WD makes fewer mistakes than UM. 
    \item The model architecture impacts the model's robustness, as also shown in previous research where it was shown that model capacity plays an important role~\cite{madry2018pgd}. For example, on \BIM\_$l_\infty(\epsilon:0.01)$ AEs, the error rates of WRN WDs lie in a wider range than that of ResNet-Shake WDs ($(21\%, 80\%)$ versus $(20\%, 58\%)$). Further, the error rate of \ourframework~with WRN models is higher than that of ResNet-Shake using the same ensemble strategy. For example, by using AVEL strategy, WRN \ourframework~has an error rate of $42\%$ and ResNet-Shake \ourframework~achieves a smaller error rate of $38\%$ on \BIM\_$l_\infty(\epsilon:0.01)$ AEs. The same observation found in MNIST, where we compared \ourframework~with CNN WDs in contrast of SVM \ourframework. However, fewer transformations in SVM \ourframework~are effective against an adversary, therefore making ensembles based on majority voting ineffective. For example, for the strongest set of \CW~AEs, all CNN WDs achieve a test accuracy above $90\%$, while only $8.5\%$ ($6$ out of $71$) SVM WDs achieve a test accuracy of $90\%$ or above, again indicating the model capacity is an important factor for countering adversarial attacks. When taking a closer look at the SVM WDs, we observed that SVM WDs associated to transformations like rotation, flipping, and distortion are not effective, indicating an interaction with the decision boundary formed by certain models. Previous research shown that deterministic SVMs are not robust against adversarial attacks~\cite{chen2018randomizing}, there is potential for investigating the connection between robust decision boundary formed by randomized SVMs~\cite{chen2018randomizing} in connection with ensemble defenses such as \ourframework.
    \item The selection of model type impacts the consistency of transformations on different attacks. For example, the performances of WDs are more consistent across various adversaries in SVM-\ourframework~than in CNN-\ourframework. Filters and segmentation achieve top $30\%$ performance across all types of adversaries, while rotation and distortion tend to work inefficiently. To further investigate this, we performed Spearman's rank correlation on classification accuracy of the $71$ SVM WDs for each pair of AE sets as shown in Figure~\ref{fig:rank_corr_svm}~\subref{subfig:rank_cross_svm_aes}. The high correlation coefficients indicate that, compared with CNN \ourframework~in Figure~\ref{fig:errrate_wb_mnist}, SVM-based WDs tend to perform more similarly across attacks. 
    \item The performances of transformations vary on different types of ML models. For example, segmentation performs the worst against most adversaries in CNN models, while it is one of the top 10 performers against in SVM-\ourframework. Specifically, we computed Spearman's rank correlation between classification accuracy of the $71$ pair of CNN and SVM classifiers per adversary. The results in Figure~\ref{fig:rank_corr_svm}~\subref{subfig:rank_cross_models} indicate that the stronger the AEs the more different transformations perform (except \BIM\_$l_2$ and \DF). 
\end{itemize}



These observations point toward a some promising further work: (i) Optimizing transformation parameters (e.g., rotation angles) and/or types based on automated search over the infinite space of possible transformations rather than manual design as we did in this paper; (ii) A search problem over the space of model architectures~\cite{guo2020meets}, model type, in addition to the ensemble parameters such as the number of individual WDs and ensemble strategy. (iii) Extending \ourframework~using hybrid ensembles including WDs that were trained on different models such as CNN and SVM.

\begin{center}
\fbox{\parbox{0.97\columnwidth}{
\textbf{Summary.} Irrespective of the type of models used for building WDs, \ourframework~can enhance robustness, but with different degrees of enhancements.}}
\end{center}

\section{RQ 3. Overhead of \ourframework}\label{subsec:overhead}

\textbf{Study Design.} 
Since we need to run WDs in parallel in \ourframework, it is clear that \emph{memory consumption} will be roughly $N$ times of that used by the UM if $N$ WDs are deployed. However, the \emph{inference time} is only limited by the slowest WD in the ensemble. Constructing \ourframework~also involves training $N$ WDs and the \emph{training time} for individual WDs is determined by the model type and training data.
To empirically evaluate the runtime overhead, we deployed \ourframework~in a Google Cloud machine (see Table~\ref{tab:hardware_config}). 




\begin{figure}[t]
    \centering
    \includegraphics[width=\linewidth]{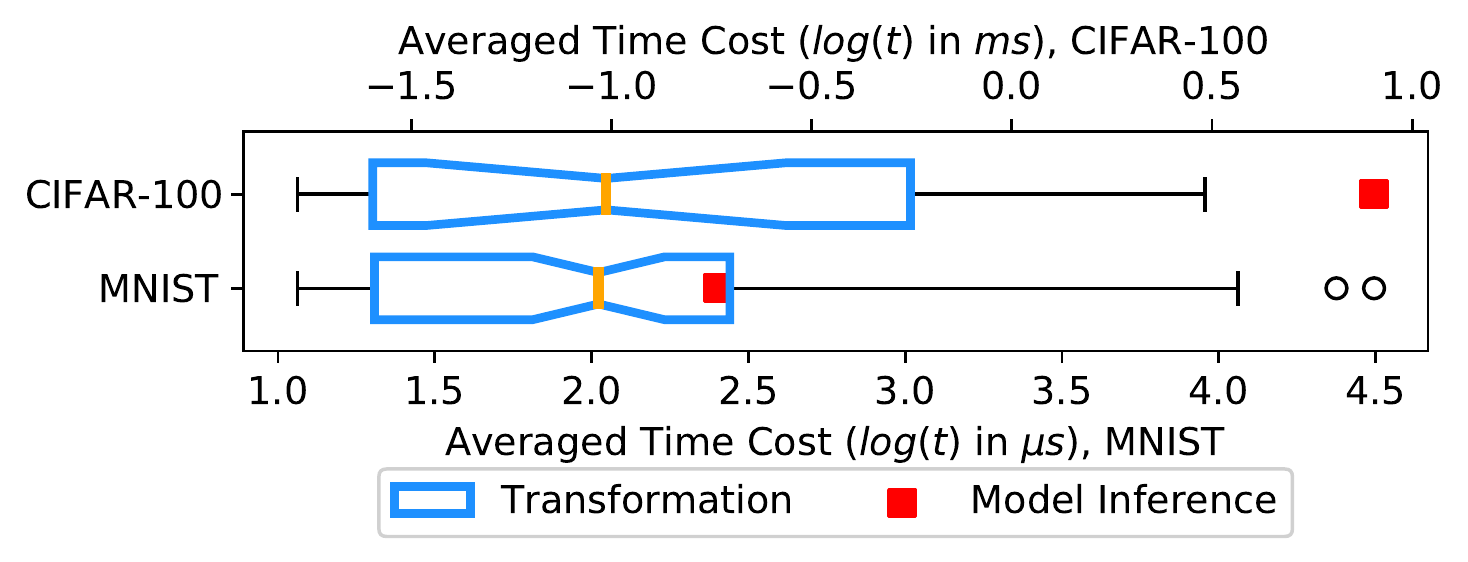}
    \caption{Overhead of \ourframework~against \FGSM.}
    \label{fig:overhead_mnist_cifar100}
\end{figure}


\textbf{Results.} 
As seen in Figure~\ref{fig:overhead_mnist_cifar100}, the most expensive transformation still consumes less time than model inference in CIFAR-100; however, for MNIST, the most expensive transformation is more costly than the undefended model inference\footnote{The cost for calculating the ensemble strategy was 10x lower than model inference, so we only reported time for performing transformations}. This points to a tradeoff space, where the size of ensemble can be adjusted to tradeoff the ensemble performance (cf. results in Section~\ref{subsec:diverse_cnn_wds}) with inference overhead, which may be necessary for time-critical applications (especially when deployed in resource constrained devices). 



\begin{center}
\fbox{\parbox{0.97\columnwidth}{
\textbf{Summary.} The memory overhead \ourframework~is proportional to the number of WDs; however, \ourframework~can be designed to have runtime overhead on par with the model inference by trading the number of transformations.}}
\end{center}

\section{Discussions: Limitations \& Future Directions}\label{sec:discussion}

Even though \ourframework~increases model robustness, the following several areas are known as areas for improvement:
\begin{itemize}
\item \emph{Integration with other defense mechanisms}: While our defense causes a significant drop in error rate in a variety of settings, there remains room for improvement. For example, in white-box scenario, we have shown that combining \ourframework with an adversarial detection make the defense significantly more effective. \ourframework~can be integrated with other defense mechanisms given its well-defined interface. 

\item \emph{Automated construction of \ourframework}: Here, we  ``manually" constructed an instance of \ourframework~with only 72 WDs; however, there is a tradeoff space, where the framework can be expanded with more (or fewer) types of transformations depending on the strength of the defense one may desire and the overhead that one can handle. 
Based on the experimental results in Sections~\ref{subsec:diverse_cnn_wds}, \ref{subsec:generality}, and \ref{subsec:overhead}, the process of constructing the ensemble can be formulated as a multi-objective optimization problem, where the aim is to automatically search for an optimal and diverse set of WDs with low overhead using a search method. Although a very difficult problem, this can be alleviated using heuristics to guide the search and incorporating composition of transformations ($t_g=t_1\circ t_2$) to facilitate the search over transformations. 

\item \emph{Building a hybrid \ourframework}:  \ourframework~is a generic framework that can be used in conjunction with a variety of ML classifiers with minimal or no modification of the original classifiers, while still being effective at recovering adversarial examples. 
The results in Section~\ref{subsec:generality} indicated that building a \emph{hybrid framework} using optimizations on the space of (i) individual WDs, (ii) model types and architecture, and (ii) ensemble strategies is a fruitful direction for a future work. 

\item \emph{Integration with cloud-based and edge-based environments}: \ourframework~is flexible so it can be adopted and integrated with machine learning models within different contexts. For cloud-based services, it has the capability to be expanded by incorporating more WDs. For resource-constrained environments such as edge, \ourframework~can be adjusted to incorporate a lower number of WDs to decrease the overhead. In both scenarios, the WDs and the ensemble strategy can be adjusted at deployment time given the environmental conditions such as availability of resources, change in the behavior of adversaries, the toleration-level of attacks by the service providers, and the cost that the provider wants to impose upon to the attacker.\footnote{We have seen that the more diverse the ensemble becomes, the more difficult it is to attack it.} Within this tradeoff space, there is potential for \emph{dynamically adapting the ensemble defense}, as well as for \emph{hybrid deployments} over multiple deployment platforms (e.g., multi-cloud~\cite{PJZ:TOIT}).


\end{itemize}

\section{Related Work}\label{sec:related_work}


Existing defense techniques against adversarial attacks can classified in two broad categories:  
(i) \emph{Model-agnostic (reactive)} approaches either detect adversarial examples before feeding them to classifiers~\cite{grosse2017statistical,metzen2017detecting}; attempt to remove adversarial perturbations from the input~\cite{das2017keeping,lu2017no,bhagoji2018enhancing,dziugaite2016study,guo2017transformation,luo2015foveation,xie2017mitigating,mustafa2019image} or features~\cite{wang2016random,zhang2015adversarial}; or combining both detection and removal~\cite{meng2017magnet}; (ii) \emph{Model-specific (proactive)} approaches, on the other hand, make changes to the model itself or the supervised learning procedure.
For example, several forms of \emph{adversarial training} based on combining normal and adversarial examples in the training~\cite{szegedy2013intriguing,tramer2017ensemble,kurakin2017physical,papernot2016distillation} or using an adversarial objective as a regularizer~\cite{goodfellow2014explaining,cisse2017parseval,shaham2018understanding,kurakin2016adversarial,cisse2017parseval,liao2018defense,kannan2018adversarial} have been proposed. 


Model-agnostic defenses have also been shown insufficient for removing adversarial perturbations from input data~\cite{carlini2017magnet}. Even small ensembles of model-agnostic defenses have been shown to be brittle even against non-adaptive adversaries~\cite{he2017adversarial}.
Model-specific defenses make strong assumptions (e.g., the norm they use for creating adversarial examples) about the type and nature of the adversary; therefore, the adversary can alter its attack to circumvent such model-specific defenses.

The concept of ``diversity" has been previously explored: (i) regularization that promotes diversity in ensembles during training~\cite{Pang2019_EnsembleDiversity}, (ii) by applying a carefully designed ensemble strategy that each individual member is randomly assigned to different tasks by special code design~\cite{verma2019error}, and (iii) mixed-precision neural networks~\cite{sen2020empir}. These approaches rely on simultaneous training of individual members and, therefore, incurs a computational cost of $\mathcal{O}(m^3)$, where $m$ is the size of the ensemble~\cite{islam2003constructive}. As a result, these methods cannot scale to large ensembles. Building defenses based on small sets of weak defenses can be easily circumvented by even weak adversaries~\cite{he2017adversarial,tramer2020adaptive}.


\section{Conclusions}\label{sec:conclusion}
We proposed \ourframework, a framework with which one can customize a specific realization of an adversarial defense based on ensembles of many and diverse sets of weak defenses given robustness properties and constraints.
We evaluated the proposed defense against state-of-the-art adversarial attacks under zero-knowledge, black-box, gray-box, and white-box threat models and found out that our defense makes the target model more robust against evasion attacks. To the best of our knowledge, even though there has been previous work on using transformation as a defense, our work is the first work that proposes a framework for building a customized defense and our comprehensive study provides evidence that an ensemble of ``many diverse weak defenses" provide such tradeoff space and has some viable potential properties:
(1) applicability across multiple models, (2) applicability in different domains (image, voice, video), and (3) agnosticism to particular attacks. 


\section{Acknowledgement}
This research is partially supported by Google (via GCP cloud research grant) and NASA (via EPSCoR 521340-SC001). We thank the Research Computing staff at the University of South Carolina (in particular Paul Sagona and Nathan Elger) for providing compute support for the experiments have been conducted in this project.
We would like to thank Forest Agostinelli and Biplav Srivastava for their feedback. 

For more information about \ourframework please refer to the project website: \url{https://softsys4ai.github.io/athena/}.

{\footnotesize \bibliographystyle{plain}
\bibliography{pooyan,adversarial}}

\appendix

\begin{table}[hp]
    \scriptsize
    \centering
    \caption{Zero-knowledge attack configurations. ``lr'' stands for the learning rate, and ``bss'' stands for the binary-search-steps.}
    \scalebox{0.9}{
        \begin{tabular}{l|ll}
        \toprule
        \textbf{Dataset}     &   \textbf{Attack}               &  \textbf{Attack configuration} \\
        \midrule
                    &   \FGSM           &  $\epsilon: 0.002, 0.003, 0.005, 0.0075, 0.03$   \\
                    &   \BIM\_$l_2$          & $\epsilon: 0.05, 0.075, 0.125, 0.3, 0.5$  \\
                    &                               & iterations: $100$\\
                       &   \BIM\_$l_{\infty}$   & $\epsilon: 0.0025, 0.005, 0.0075, 0.01, 0.015$    \\
                    &                        & iterations: $100$          \\
                    &   \CW\_$l_2$          & lr: $0.0003, 0.001, 0.007, 0.02, 1.0$ \\
        CIFAR-100            &                       & bss: $6$ \\
                    &                        & iterations: $10$ \\
                    &   \JSMA                & $\theta: 0.01, 0.5, 0.25, -0.5, -0.3$ \\
                    &                       & $\gamma: 0.04, 0.03, 0.05, 0.075, 0.1$ \\
                    &   \PGD            & $\epsilon:  0.0015, 0.0025, 0.0035, 0.005, 0.015$  \\
                    &                       & eps\_iter: $\epsilon/10$ \\
                    &                       & iterations: $40$ \\
        \midrule
                    &   \FGSM       & $\epsilon:  0.1, 0.15, 0.2, 0.25, 0.3$ \\
                    &   \BIM\_$l_2$   & $\epsilon:  0.75, 0.85, 1.0, 1.1, 1.2$ \\
                    &                & iterations: $100$ \\
                    &   \BIM\_$l_{\infty}$   & $\epsilon: 0.075, 0.082, 0.09, 0.1, 0.12$ \\
                    &                               & iterations: $100$ \\
                    &   \CW\_$l_2$          & lr:  $0.0098, 0.01, 0.012, 0.015, 0.018$\\
                    &                               & bss: $5$ \\
                    &                               &  iterations: $100$ \\
               &   \DF                      & overshoot: $3/255, 5/255, 8/255, 16/255, 50/255$ \\
                    &                           & iterations: $50$ \\
                    &   \JSMA    & $\theta: 0.15, 0.17, 0.18, 0.21, 0.25$ \\
          MNIST          &                               & $\gamma: 0.5$ \\
                    &   \OP   & pixel-count: $15, 30, 35, 40, 75$ \\
                    &                               & population-size: $100$ \\
                    &                               & iterations: $30$ \\
                    &   \MIM            &  $\epsilon: 0.06, 0.067, 0.075, 0.085, 0.1$  \\
                    &                   & iterations: $1000$ \\
                    &                   & delay\_factor: $1.0$ \\
                    &   \PGD            & $\epsilon:  0.075, 0.082, 0.09, 0.1, 0.11$ \\
                    &                       & eps\_iter: $\epsilon/10$ \\
                    &                               & iterations: $40$ \\
        \bottomrule
        \end{tabular}
    }
    \label{tab:zk_attack_config}
\end{table}

\begin{table}[!htb]
    \scriptsize
    \centering
    \caption{White-box and gray-box attack configurations.}
    \begin{tabular}{l|llll}
    \toprule
    Dataset & Strategy & \# of       & Max  & Attack \\
            &           & WDs     & Dissimilarity & Configuration \\
    \midrule
    CIFAR-100 & MV & $22$ & $0.05, 0.075, 0.1,$ & $\epsilon=0.01$ \\ 
            &   &       &$0.125, 0.15, 0.175, 0.2,$ &   \\ 
            &   &       &$0.225, 0.25, 0.275$ &\\
    \midrule
    MNIST  & MV & $72$ & $0.1, 0.2, 0.3, 0.4, $ & $\epsilon=0.1$ \\
         &    &      &$ 0.5, 0.6, 0.7,$     & \\
            &   &     & $0.8, 0.9, 1.0$ & \\
    \bottomrule
    \end{tabular}
    \label{tab:wb_setting}
\end{table}

\begin{table}[hp]
    \scriptsize
    \caption{Hardware configurations for our experiments.}
    \scalebox{0.78}{
        \begin{tabular}[width=\linewidth]{c|l}
        \toprule
        \textbf{Dataset}    &   \textbf{Experiment}     \\
        \midrule
                            &   \textbf{Training WDs}   \\
                            & $\quad$ PC 12 Intel(R) Core(TM)i7-8700 CPU @ 3.20GHz, 32 GB memory  \\[0.6em]
                            &   \textbf{Crafting AEs}    \\ 
                             &   zero-knowledge \\
                            & $\quad$ Google Cloud VM: \\
                            & $\quad$ 8 Intel(R)Xeon(R) CPU @ 2.20GHz, 30 GB memory, 1 x NVIDIA Tesla P100  \\[0.3em]
                            &   black-box   \\
         MNIST    & $\quad$ Google Cloud VM: \\
                            & $\quad$ 16 Intel(R)Xeon(R) CPU @ 2.30GHz, 14.4 GB memory  \\[0.3em] 
                            &  white-box   \\
                            & $\quad$ Google Cloud VM: \\
                            & $\quad$ PC 12 Intel(R) Core(TM)i7-8700 CPU @ 3.20GHz, 32 GB memory\\[0.6em]
                            &   \textbf{Examining overheads} \\
                            &  $\quad$ Google Cloud VM: \\
                            & $\quad$ 16 Intel(R)Xeon(R) CPU @ 2.30GHz, 14.4 GB memory  \\
        \midrule
                            &   \textbf{Training WDs}   \\
                            & $\quad$ Google Cloud VM: \\
                            & $\quad$ 8 Intel(R)Xeon(R) CPU @ 2.20GHz, 30 GB memory, 1 x NVIDIA Tesla P100  \\[0.6em]
                            &   \textbf{Crafting AEs}  \\
                              &   zero-knowledge          \\
                            & $\quad$ Google Cloud VM: \\
      CIFAR-100             & $\quad$ 8 Intel(R)Xeon(R) CPU @ 2.20GHz, 30 GB memory, 1 x NVIDIA Tesla P100  \\[0.3em]
                            &   black-box  \& white-box \\
                            & $\quad$ Hyperion new 16gb GPU node:    \\
                            & $\quad$  48 cores, 128 GB memory, 1.5T local scratch, 2 x NVIDIA Volta V100 16GB gpgpus \\[0.6em]
                            &   \textbf{Examining overheads} \\
                            & $\quad$ Hyperion new 16gb GPU node:    \\
                            & $\quad$  48 cores, 128 GB memory, 1.5T local scratch, 2 x NVIDIA Volta V100 16GB gpgpus \\
        \bottomrule
        \end{tabular}
    }
    \label{tab:hardware_config}
\end{table}

\begin{table}[h]
    \scriptsize
    \centering
    \caption{List of input transformations}
    \scalebox{0.72}{\footnotesize{\sf
\begin{tabular}[width=\linewidth]{ll}
\toprule
\textbf{Category}    &   \textbf{Description} \\
\midrule
Rotation    & Rotate an input by a certain angle: rotate $\text{90}^{\circ}, \text{180}^{\circ}, \text{and}~ \text{270}^{\circ}$.  \\
\midrule
Flip        & Flip an input horizontally$^*$ and/or vertically. \\
\midrule
Shift       & Shift an input in a direction by some pixels: shift left$^*$, right, up, \\   
            & down$^*$, top-left, top-right, bottom-left, and bottom-right.    \\
\midrule
Cartoonify$^*$      & Cartoonify apples a sequence of image process operations on\\
                & the input, including bilateral filter, gray scaling, median blur,\\
                & create edge mask, and add the mask to original input. \\
\midrule
Affine$^*$      & Transform an input by mapping variables (e.g., pixel intensity                    \\   
            & values at position $(x_1, y_1)$) into new variables (e.g., $(x_2, y_2)$)    \\               
            & by applying a linear combination of shift, rotation,                 \\               
            & scaling and/or shearing operations: compress vertically        \\
            & and/or horizontally, stretch vertically and/or horizontally.   \\
\midrule
Denoise     & Noise is generally considered to be a random variable with zero\\
            & mean. Denoise transformations average out noises from inputs:\\
            & nl\_means\_fast$^*$, nl\_means, tv\_menas, tv\_chambolle, tv\_bregman$^*$, and wavelet$^{*}$. \\
\midrule
Morphology  & Morphological transformations apply operations based on            \\   
            & image shape: erosion$^*$, dilation$^*$, opening$^*$, closing$^*$, and gradient.    \\
\midrule
Noise       &   Add noises to an input: \\
            &   gaussian, localvar, pepper, poison$^*$, salt, and salt\&peper. \\
\midrule
Augmentation& Real-time data augmentation:                     \\
            & feature-wise std normalization$^*$ and sample-wise std normalization$^*$.  \\              
\midrule
Segmentation  & Segmentation divides the image into groups of pixels based on \\
              & specific criteria. We segment an input based on colors. \\
\midrule
Quantization& Quantization reduces the number of colors in an input using   \\
            & k-mean technique: 4 clusters and 8 clusters.               \\
\midrule
Distortion  & Distortion deforms the pixel grid in an input and maps the deformed \\
            & grid to the destination image: distort by x-axis or y-axis.\\
\midrule
Filter      & Filter transformations smooth an input using various filter kernels:            \\
            & entropy, gaussian, maximum, median$^*$, minimum, prewitt, rank$^*$,            \\
            & scharr, roberts, sobel.                                          \\
\midrule
Compress    & Save images in different image formats: jpeg \\
            & (quality: 80\%$^{*}$, 50\%, 30\%, and 10\%), png(compression: 1$^*$, 5$^*$, and 8$^*$). \\
\midrule
Geometric   & Apply different geometric transformations to images: iradon, \\
            & iradon\_sart, and swirl$^*$ \\
\bottomrule
\label{tab:transformations}
\end{tabular}
}}
    \label{tab:transformation_list}
\end{table}

\begin{figure}[hp]
    \tiny
    \centering
    \includegraphics[width=0.9\linewidth]{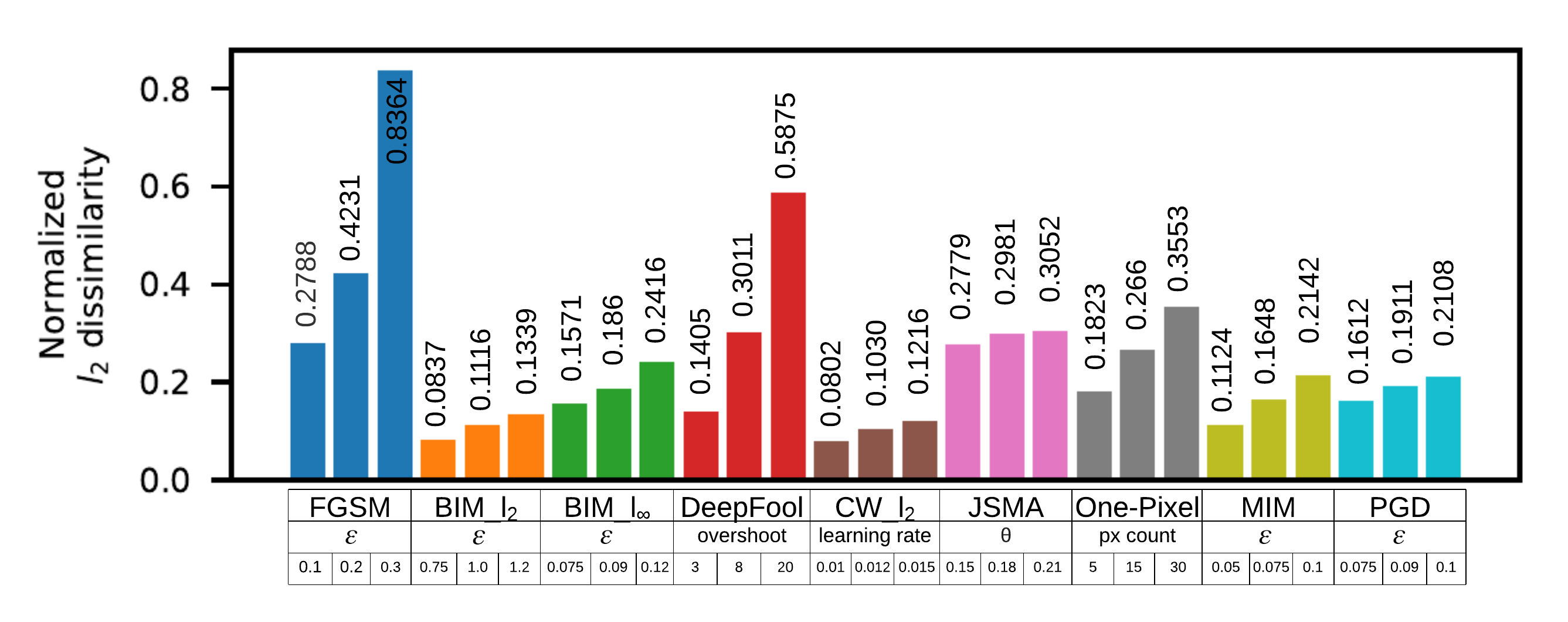}
    \caption{Normalized $l_2$ dissimilarity between BS and each AE generated for MNIST.}
    \label{fig:dissimilarity_MNIST}
\end{figure}

\begin{figure}[hp]
    \tiny
    \centering
    \includegraphics[width=0.9\linewidth]{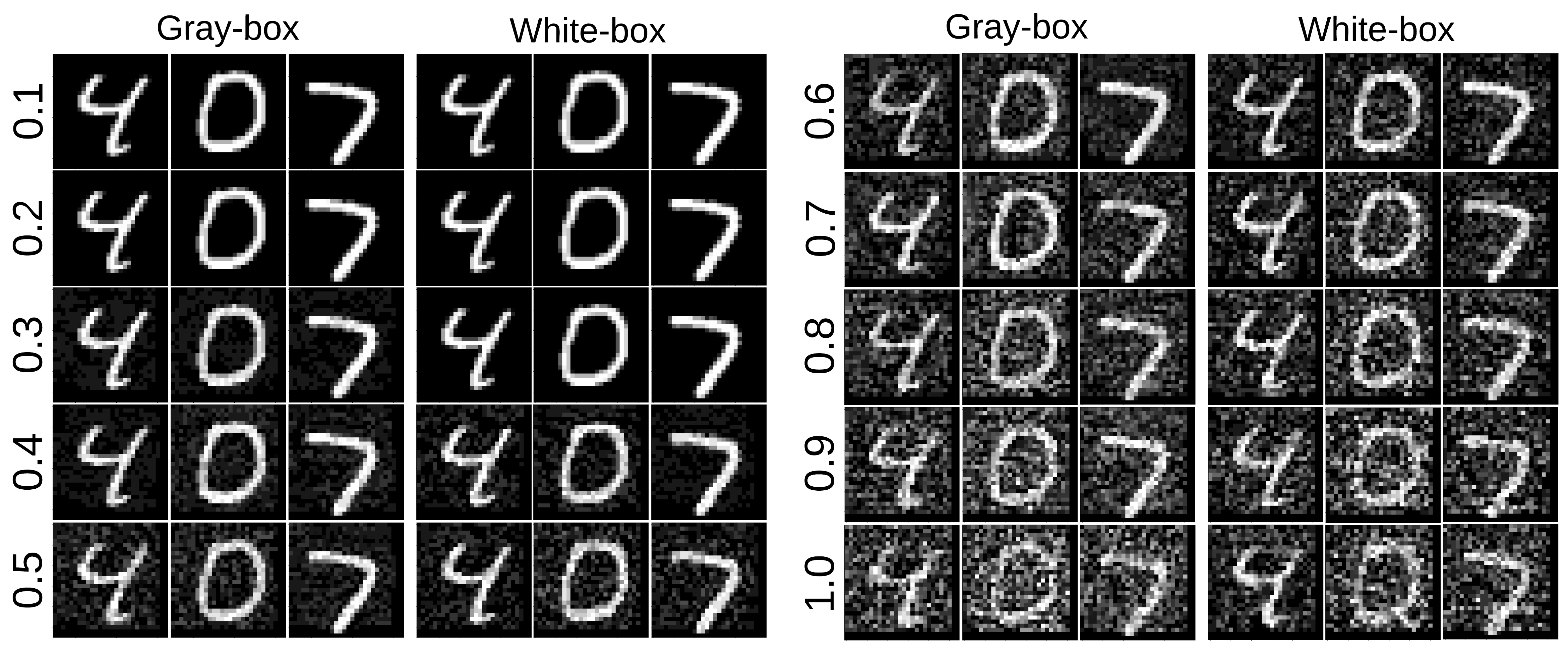}
    \caption{MNIST sample adversarial examples generated in and white-box threat models with various constraints on maximum normalized dissimilarity.}
    \label{fig:mnist_wb_samples}
\end{figure}

----------------------------
\begin{figure}[h]
    \scriptsize
    \centering
    \includegraphics[width=0.9\linewidth]{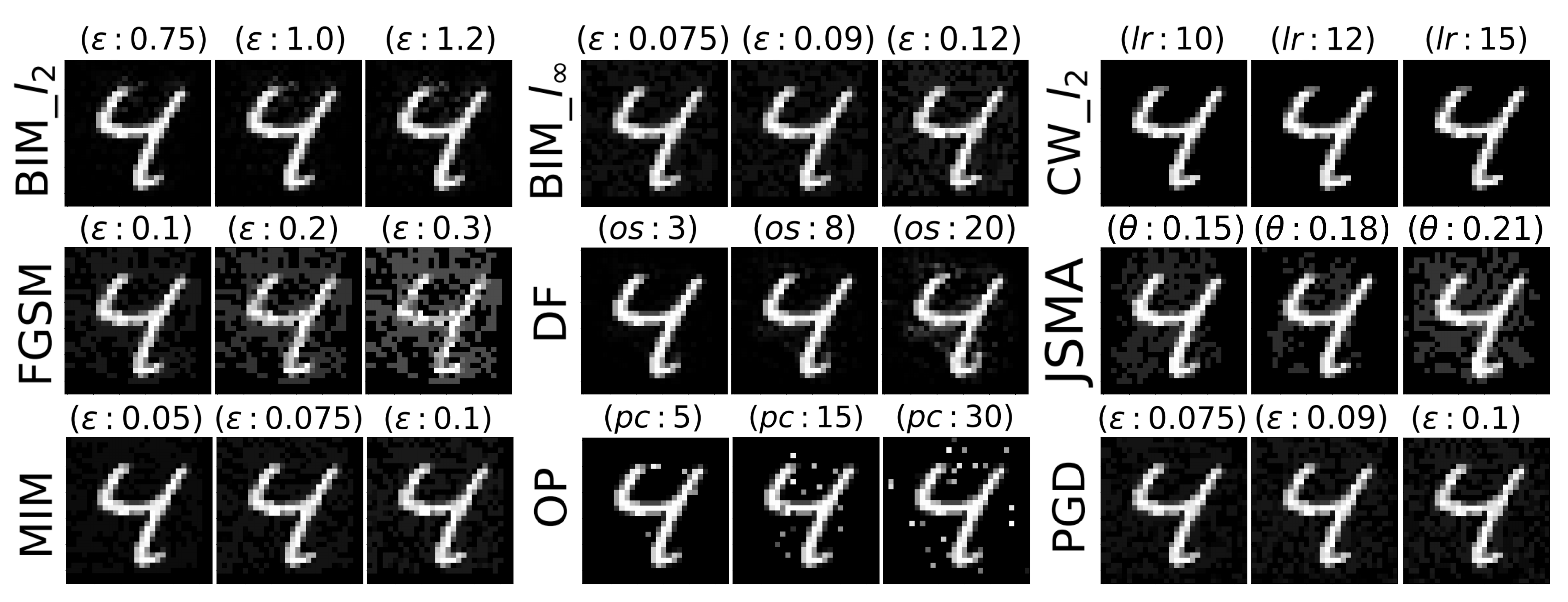}
    \caption{MNSIT sample adversarial examples generated by various adversaries.}
    \label{fig:mnist_zk_samples}
\end{figure}

\begin{figure}[hp]
    \tiny
    \centering
    \includegraphics[width=\linewidth]{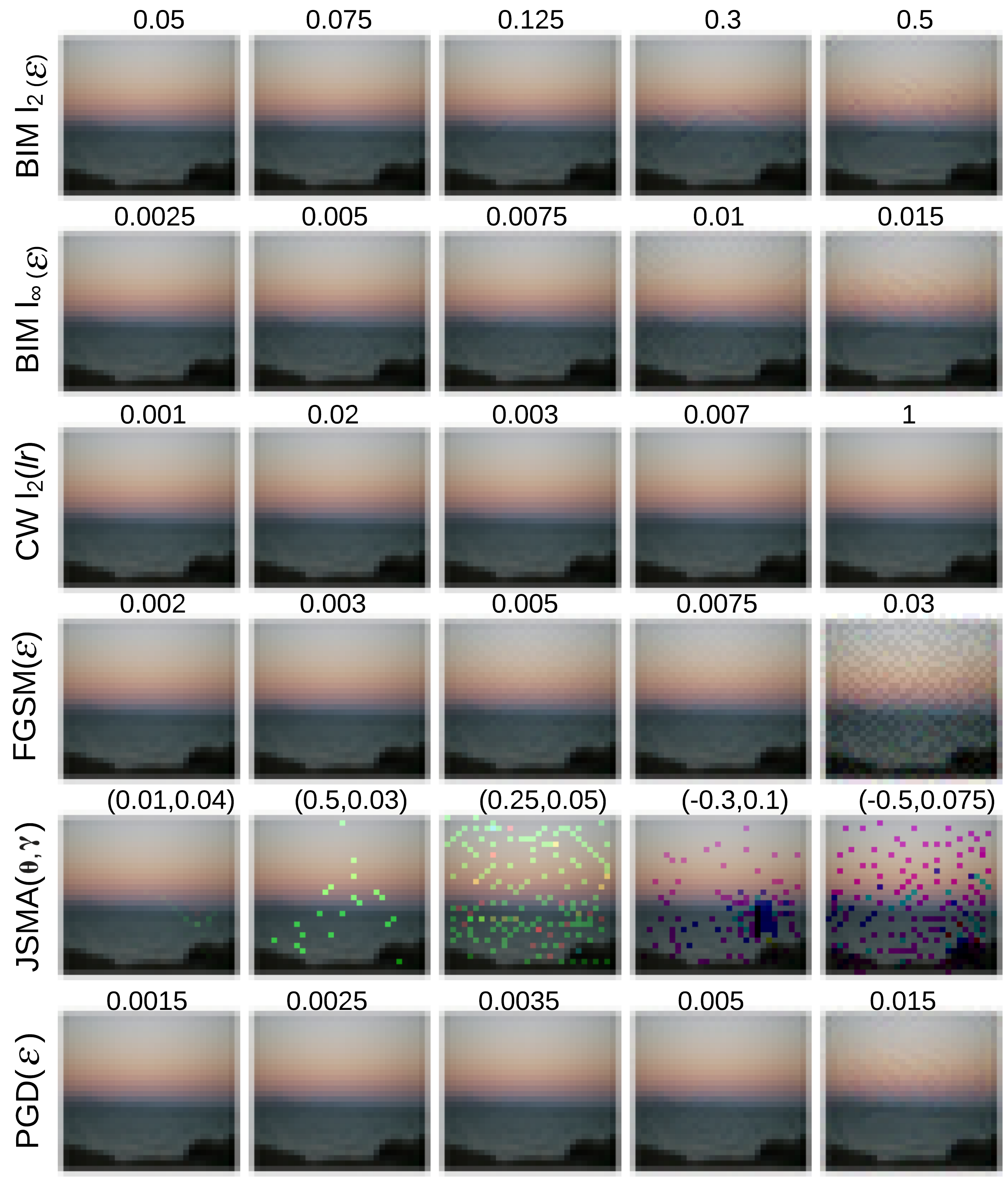}
    \caption{CIFAR-100 sample adversarial examples generated by various adversaries.}
    \label{fig:cifar100_zk_samples}
\end{figure}

\begin{figure}[h]
    \tiny
    \centering
    \includegraphics[width=\linewidth]{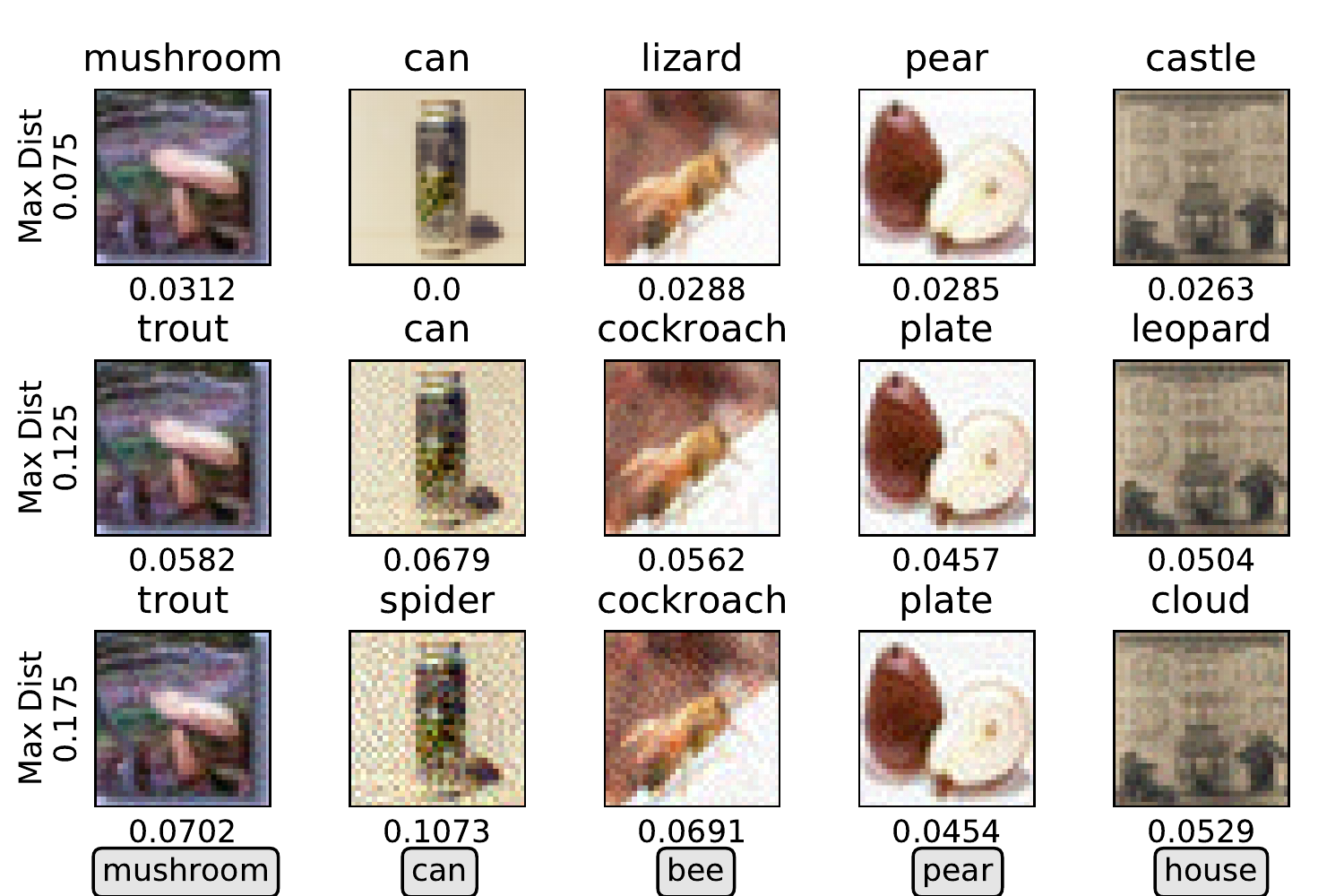}
    \caption{The adversarial examples (CIFAR-100) generated by greedily aggregating perturbations under gray/white-box threat models with various  constraints on maximum normalized dissimilarity. Note: 1) the text above each subfigure is its predicted label by \ourframework; 2) the number under each subfigure indicates its l2 distance from the original benign sample; 3) true labels are displayed in round rectangles.}
    \label{fig:cifar100_wb_greedAE}
\end{figure}

\begin{figure}
    \footnotesize
    \centering
    \includegraphics[width=\linewidth]{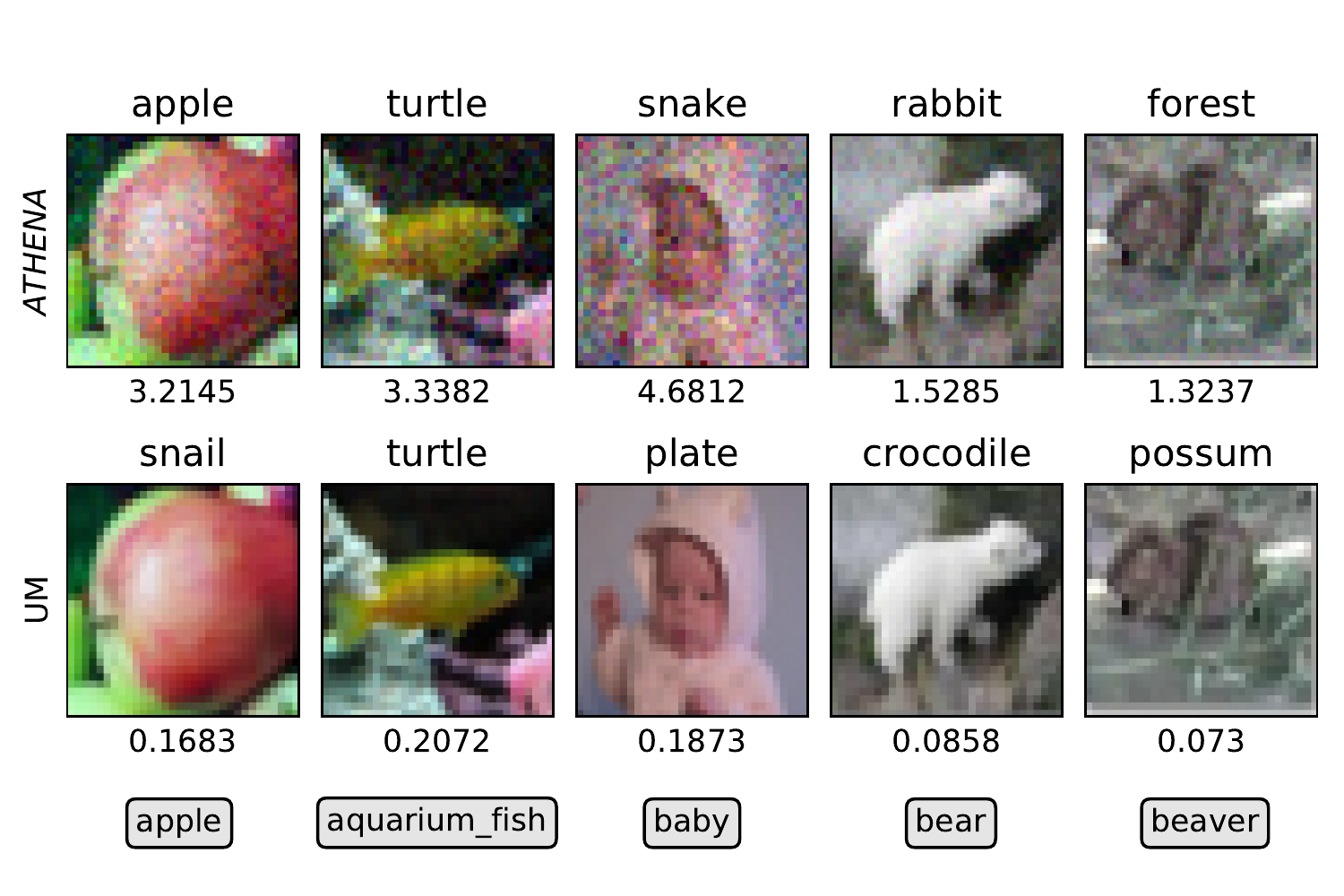}
    \caption{The adversarial examples (CIFAR-100) generated by HSJA with $l_2$ distance and a budget of 5000 queries against the undefended model (UM) and \ourframework. Note: 1) the text above each subfigure is its predicted label by the corresponding target model; 2) the number under each subfigure indicates its $l_2$ distance from the original benign sample; 3) true labels are displayed in round rectangles.}
    \label{fig:hsja_l2-um_vs_avep}
\end{figure}


\begin{figure}[t]
    \centering
    \includegraphics[width=0.9\linewidth]{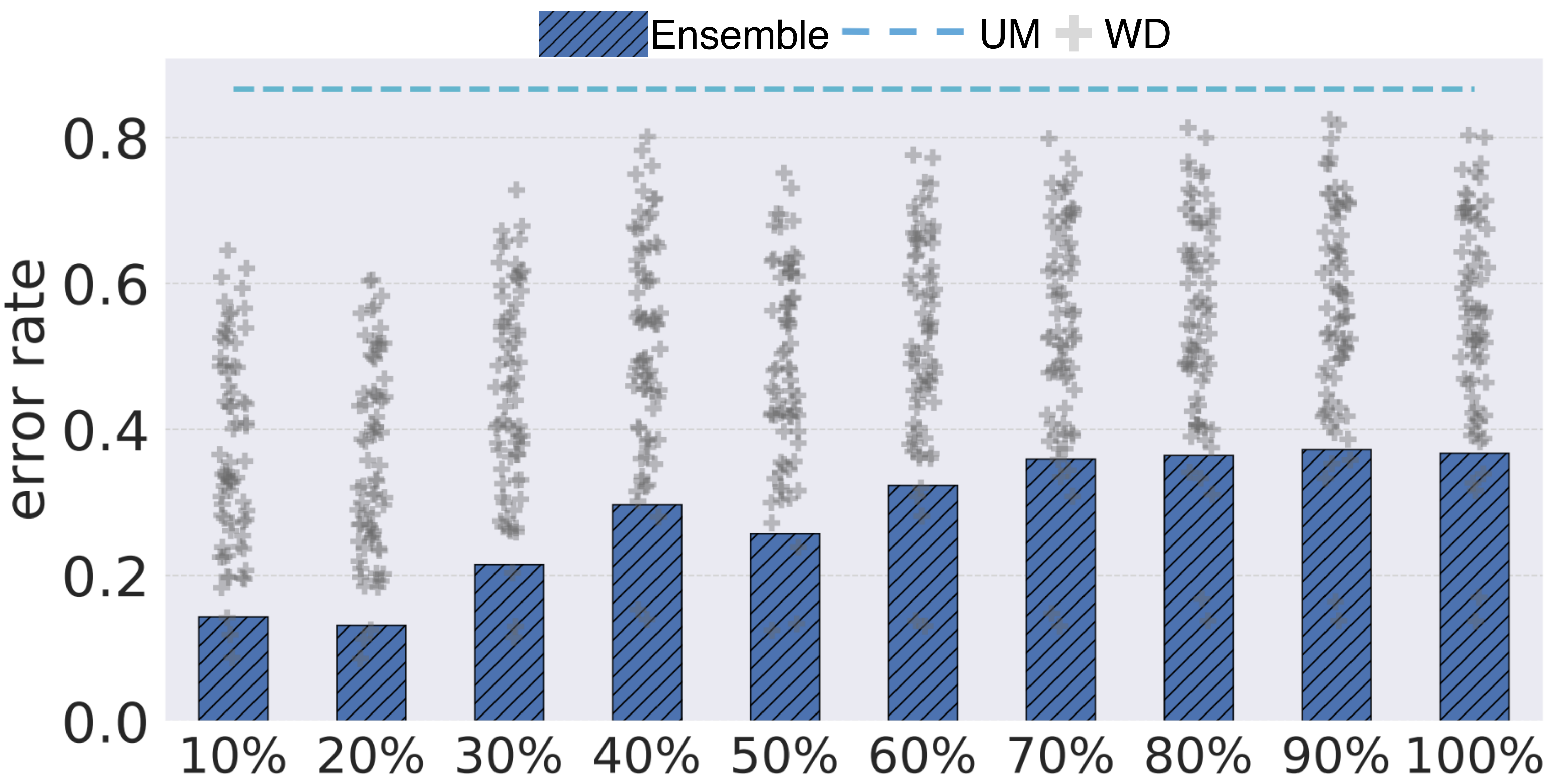}
    \caption{Evaluations of an AVEP ensemble against the optimization-based attack on MNIST.} 
    \label{fig:err_synthesis_mnist}
\end{figure}

\begin{figure}[t]
    \captionsetup[subfigure]{font=scriptsize,labelfont=scriptsize}
    \scriptsize
    \centering
    \subfloat[Normalized $l_2$ dissimilarity]{
        \includegraphics[width=0.49\linewidth]{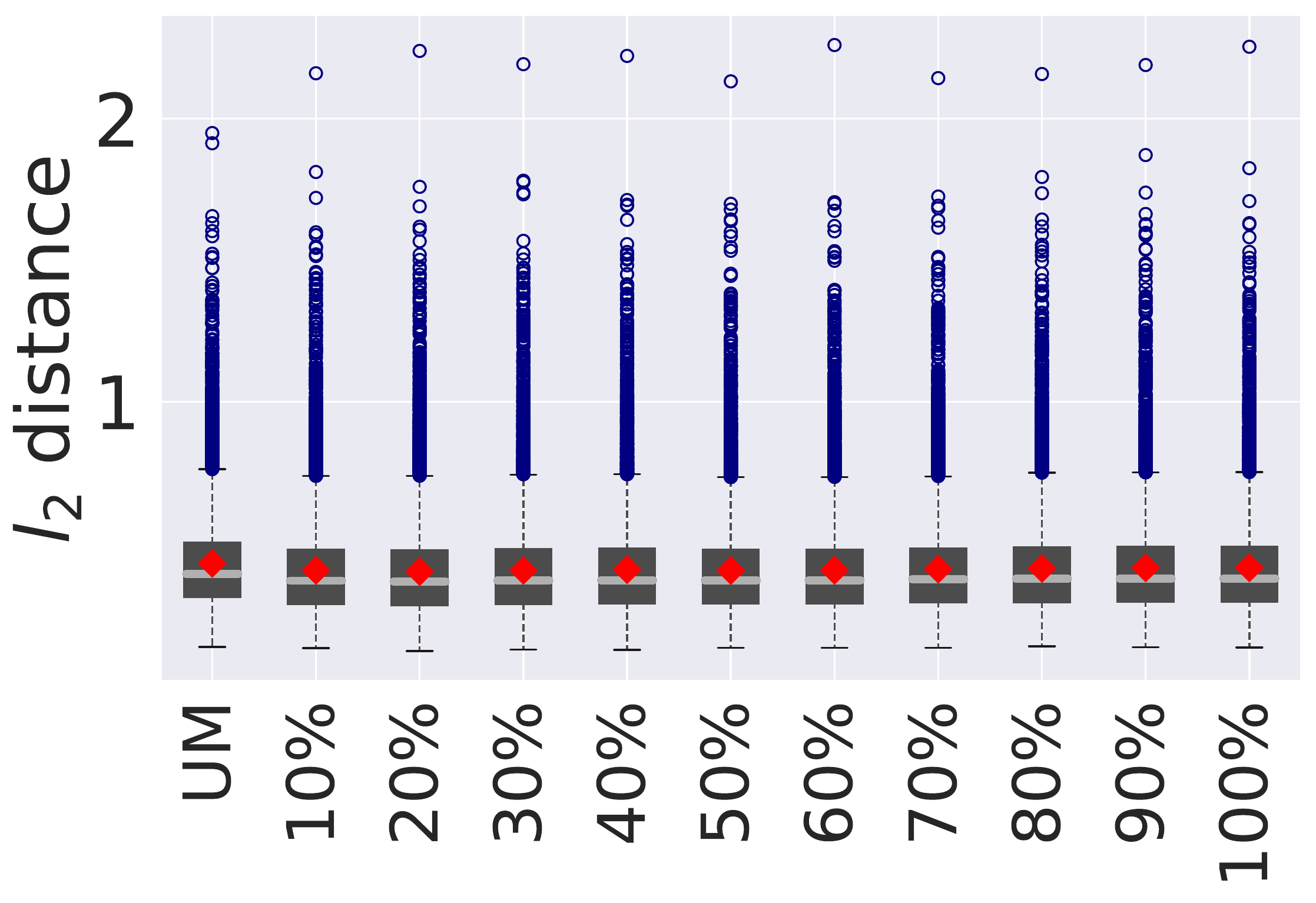}
        \label{subfig:synthesis_cifar100_dist}
    } 
    \subfloat[Computational cost]{
        \includegraphics[width=0.49\linewidth]{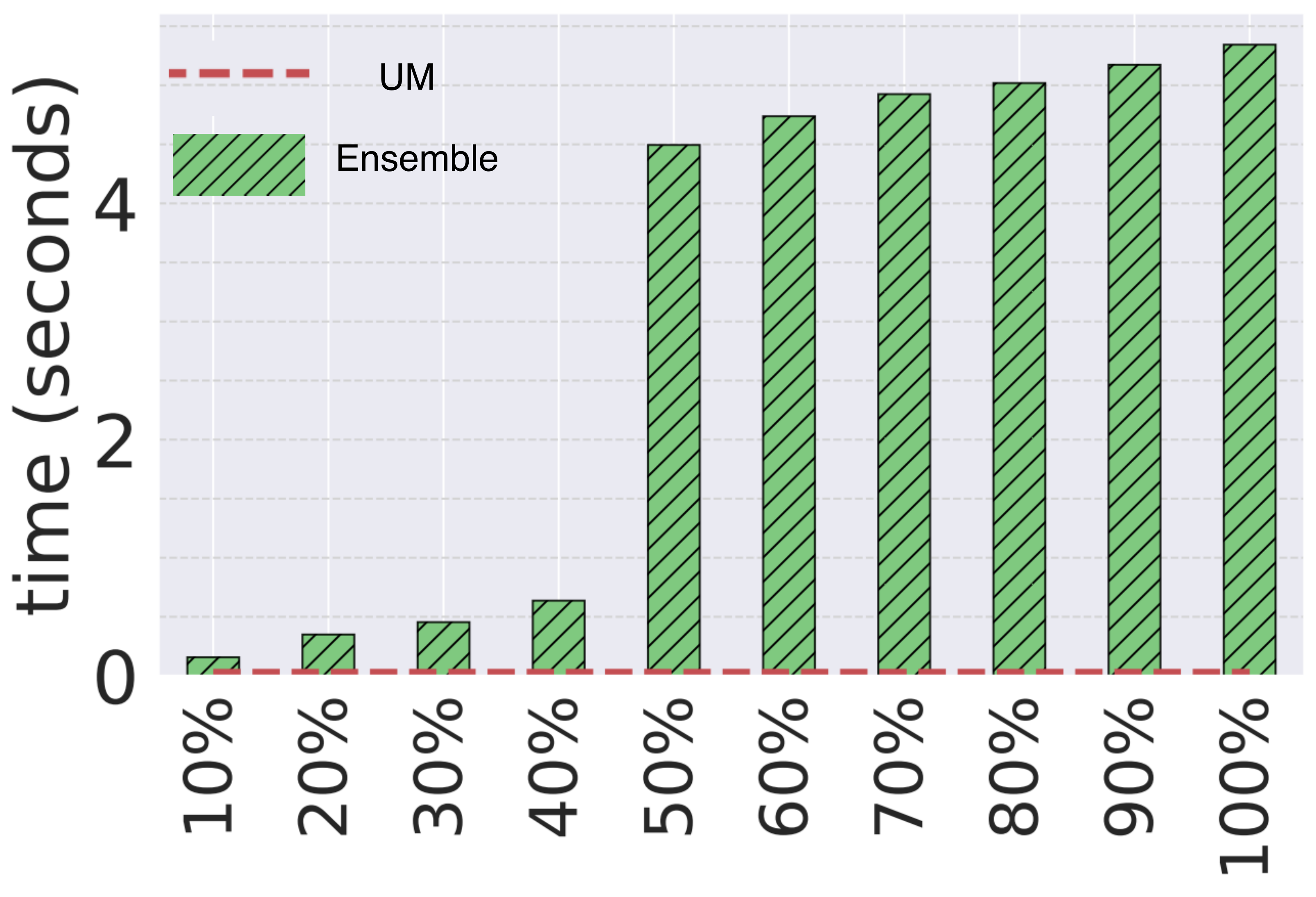}
        \label{subfig:synthesis_cifar100_cost}
    }
    \caption{The normalized $l_2$ dissimilarity of AEs and computational time of the optimization-based attack on MNIST.}
    \label{fig:eval_synthesized_ae_mnist}
\end{figure}

\begin{figure}[h]
    \captionsetup[subfigure]{font=scriptsize,labelfont=scriptsize}
    \scriptsize
    \centering
    \subfloat[Time cost (seconds)]{
        \includegraphics[width=0.47\linewidth]{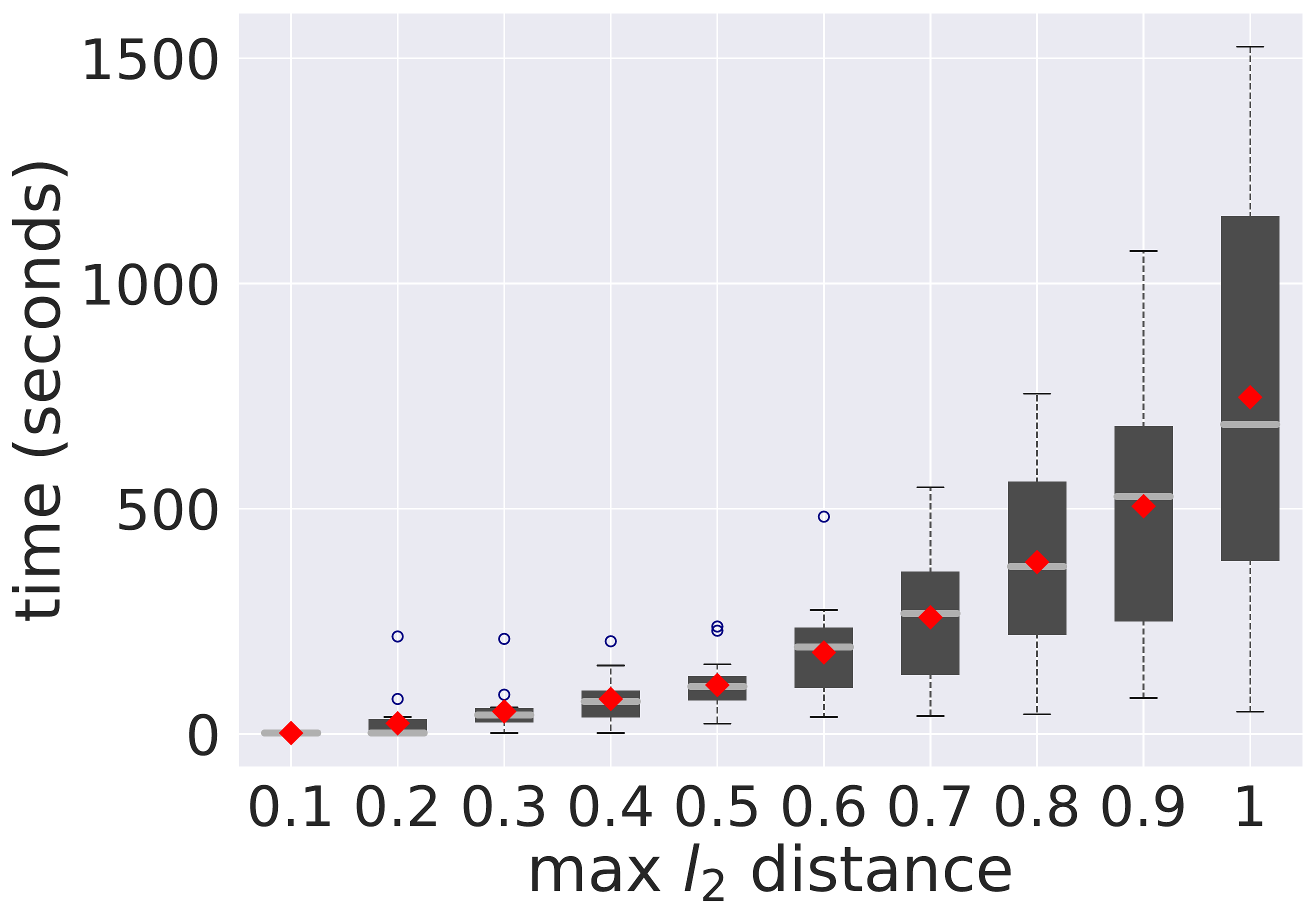}
    } 
    \subfloat[Normalized $l_2$ dissimilarity]{
        \includegraphics[width=0.45\linewidth]{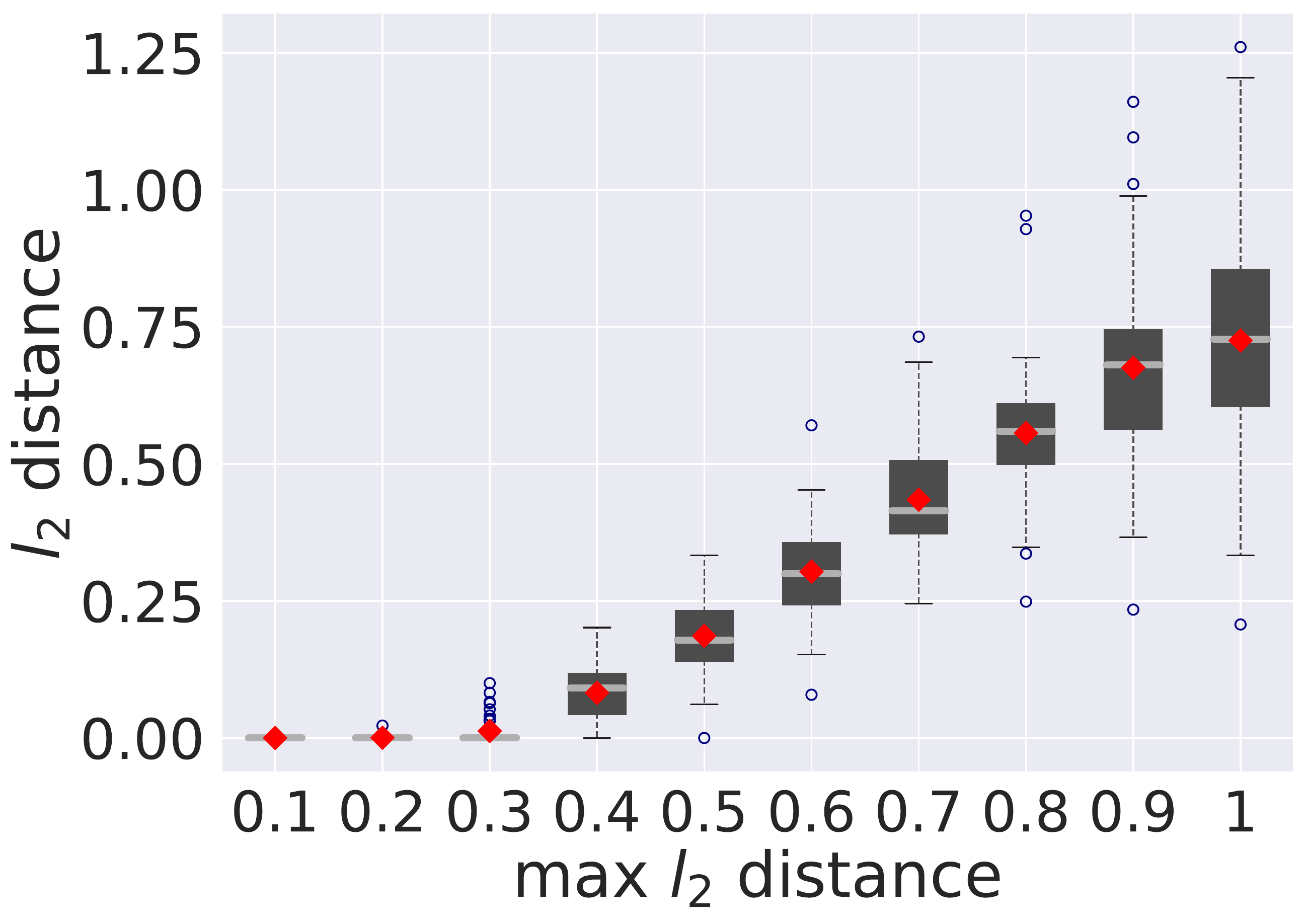}
    } \\
    \subfloat[Number of iterations (Algorithm~\ref{alg:attack_white-box})]{
        \includegraphics[width=0.47\linewidth]{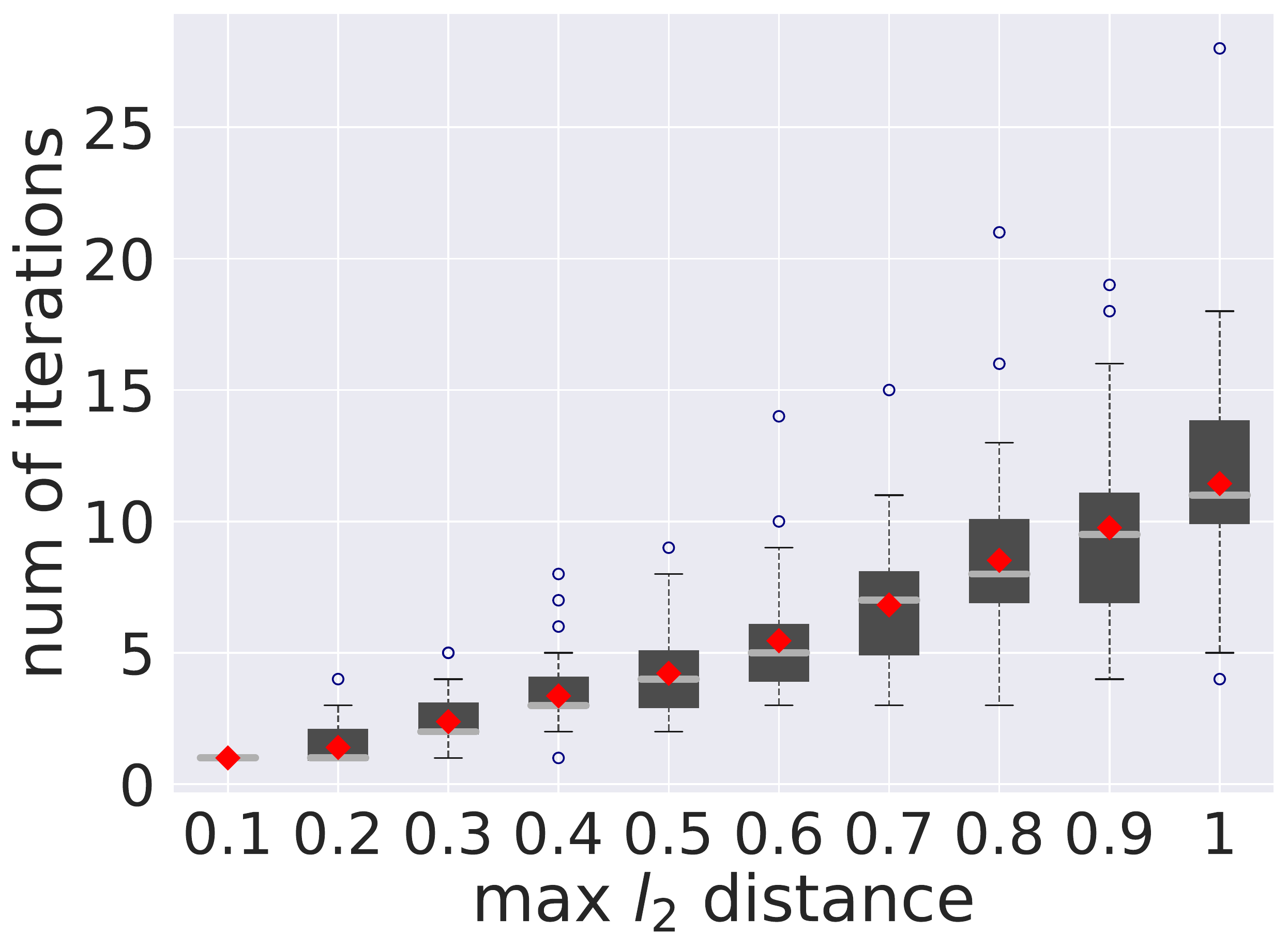}
    }
    \subfloat[Number of WDs being fooled]{
        \includegraphics[width=0.45\linewidth]{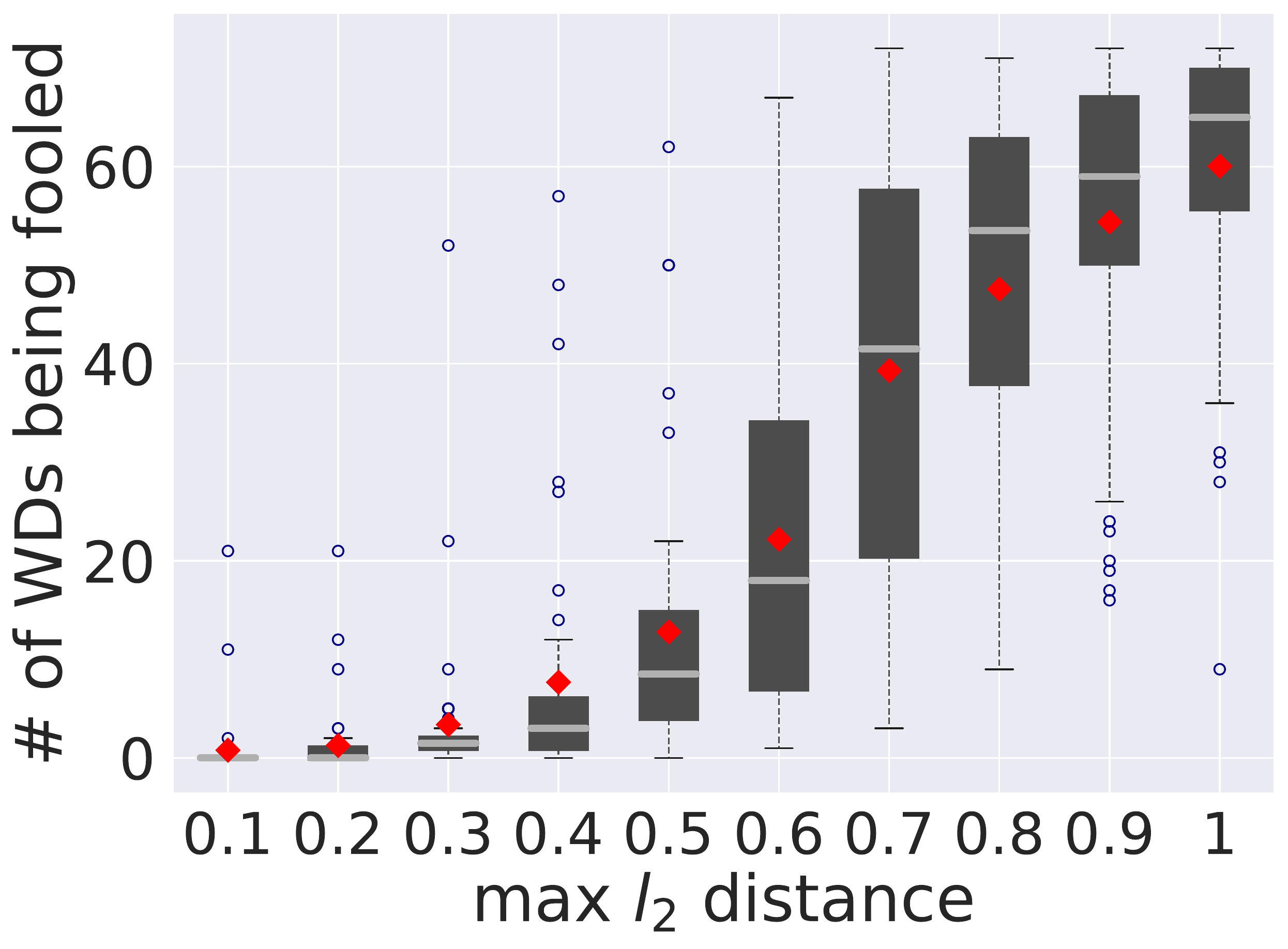}
    }
    \caption{Cost evaluation of AEs generated based on white-box model (MV strategy) of the greedy attack for MNIST.}
    \label{fig:cost_wb_mnist}
\end{figure}

\begin{figure}[hp]
    \tiny
    \centering
    \subfloat[][\FGSM]{
        \includegraphics[width=0.49\linewidth, trim={0.8cm 0.7cm 0.8cm 1.5cm}]{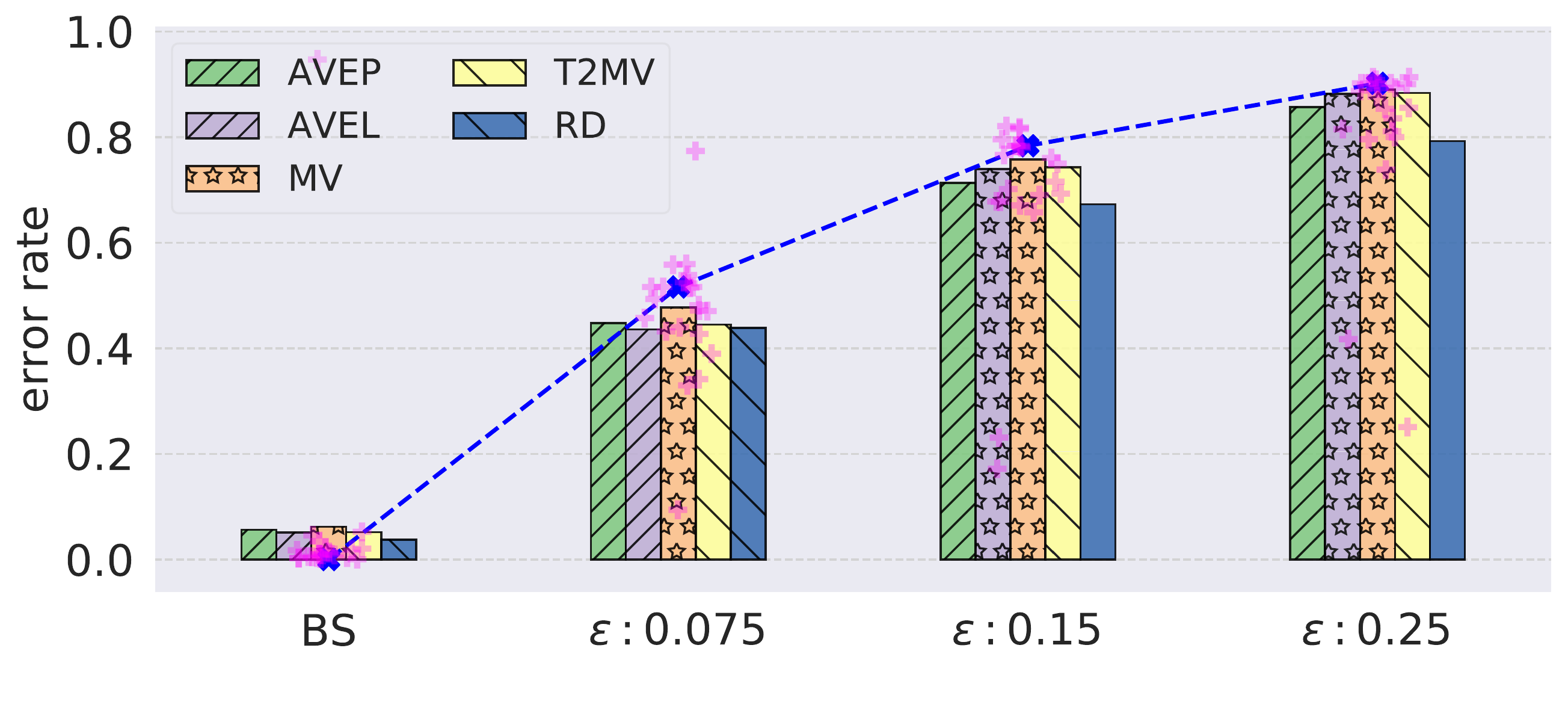}
    }
    \subfloat[][\BIM\_$l_2$]{
        \includegraphics[width=0.49\linewidth, trim={0.8cm 0.7cm 0.8cm 1.5cm}]{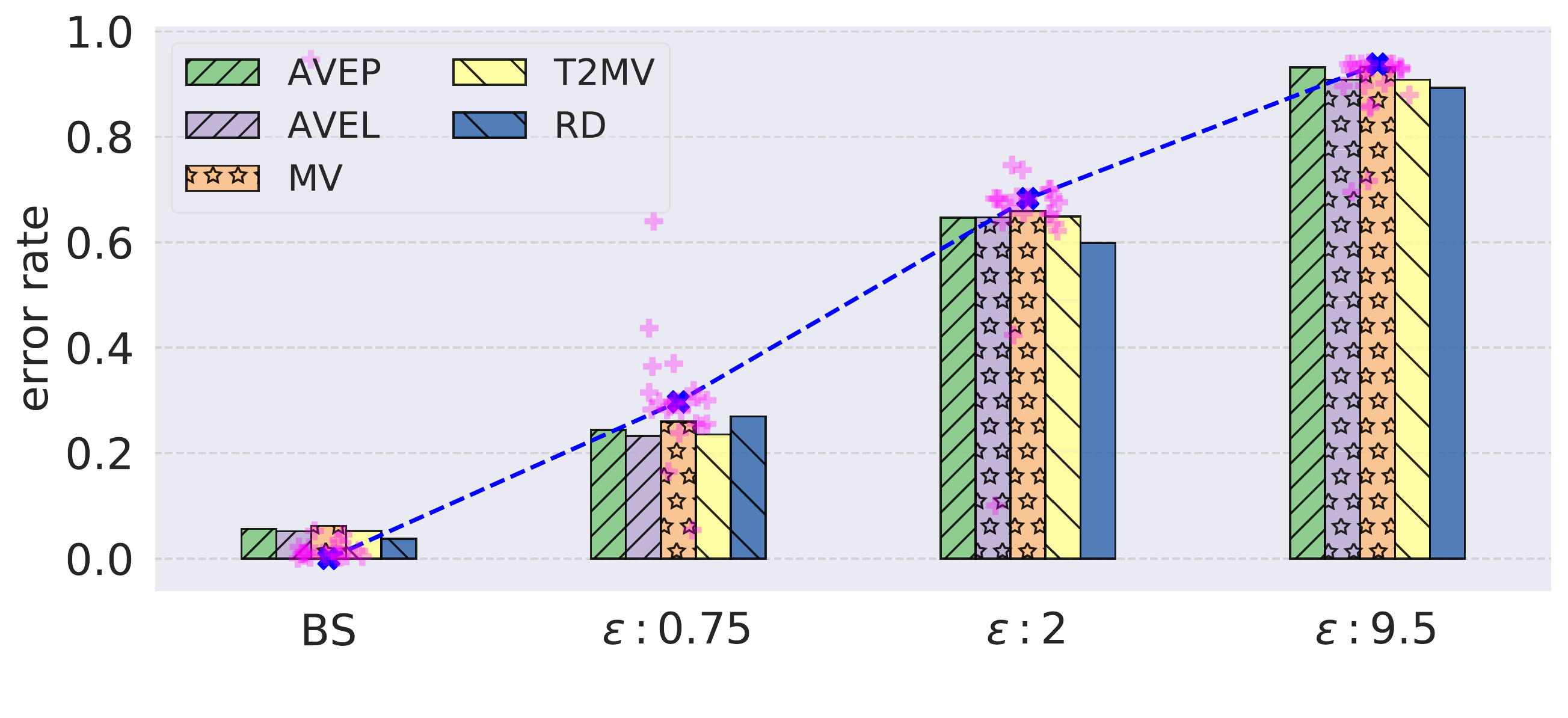}
    } \\
    \subfloat[][\BIM\_$l_{\infty}$]{
        \includegraphics[width=0.49\linewidth, trim={0.8cm 0.7cm 0.8cm 1.5cm}]{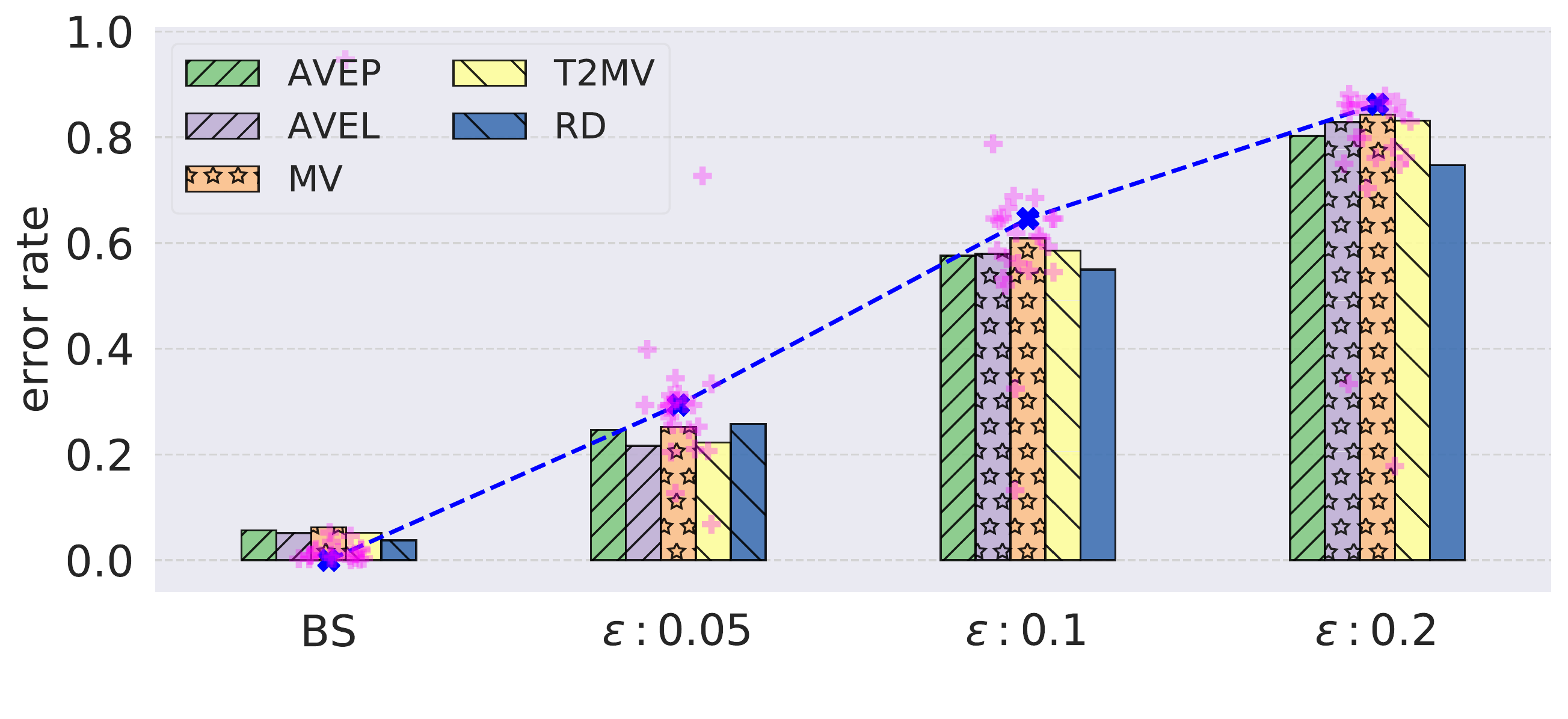}
    } 
    \subfloat[][\CW\_$l_2$]{
        \includegraphics[width=0.49\linewidth, trim={0.8cm 0.7cm 0.8cm 1.5cm}]{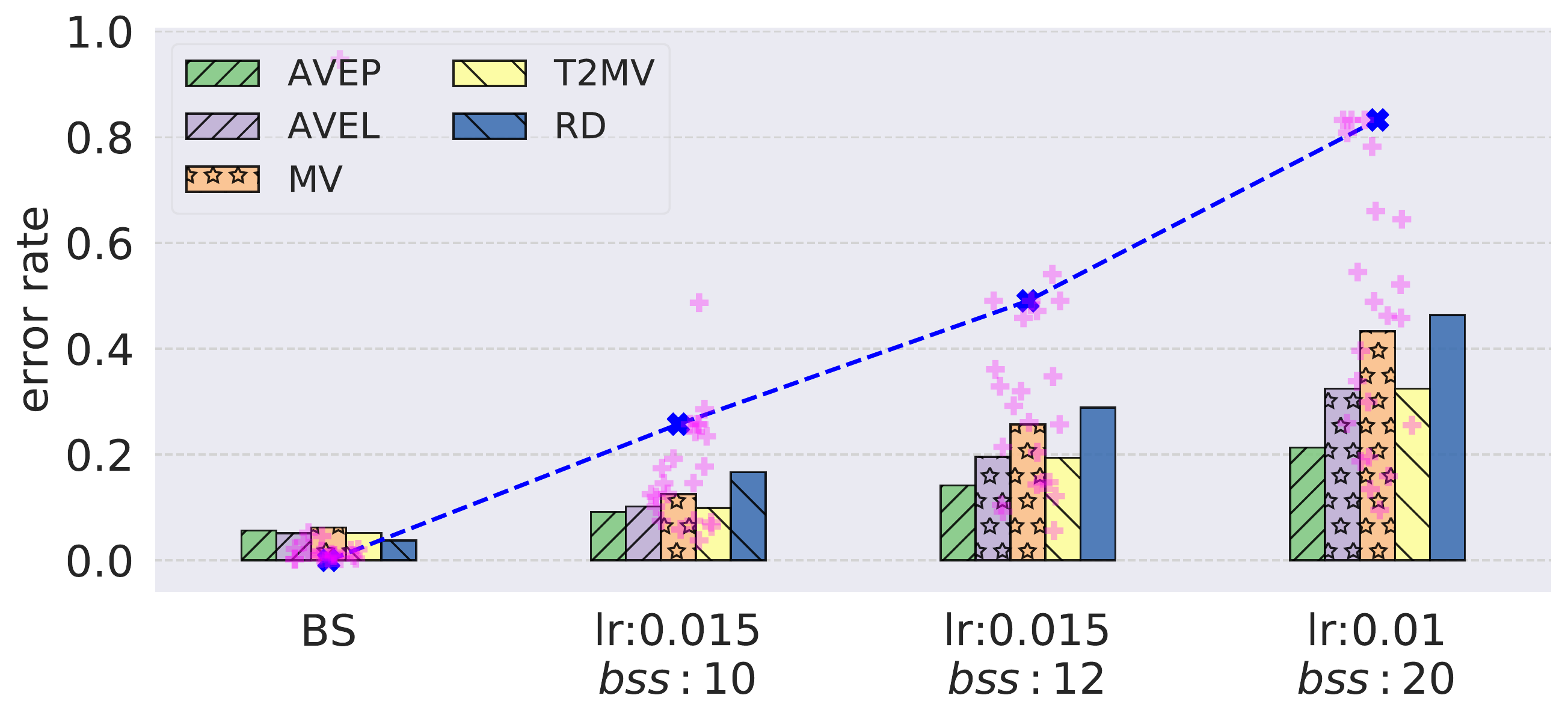}
    } \\
    \subfloat[][\JSMA]{
        \includegraphics[width=0.49\linewidth, trim={0.8cm 0.7cm 0.8cm 1.5cm}]{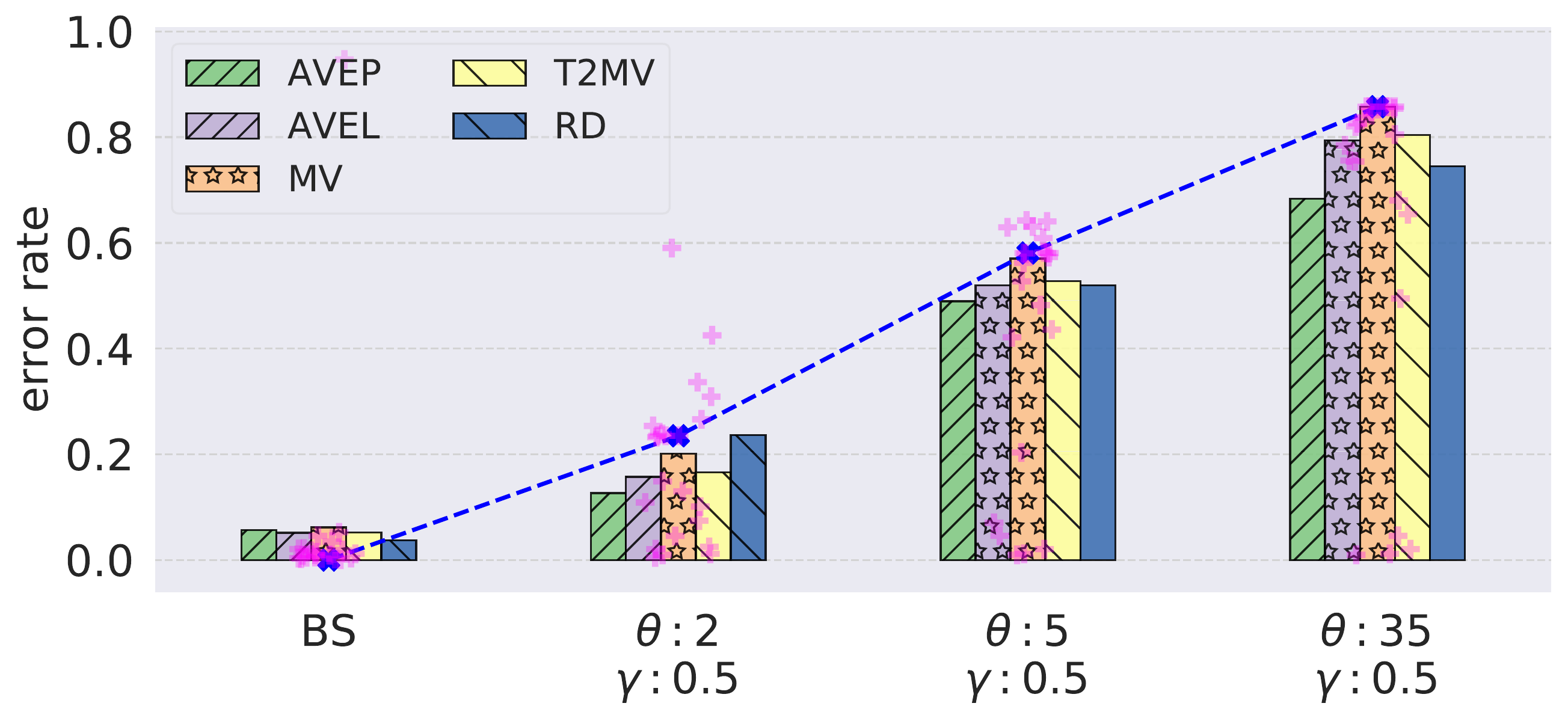}
    } 
    \subfloat[][\PGD]{
        \includegraphics[width=0.49\linewidth, trim={0.8cm 0.7cm 0.8cm 1.5cm}]{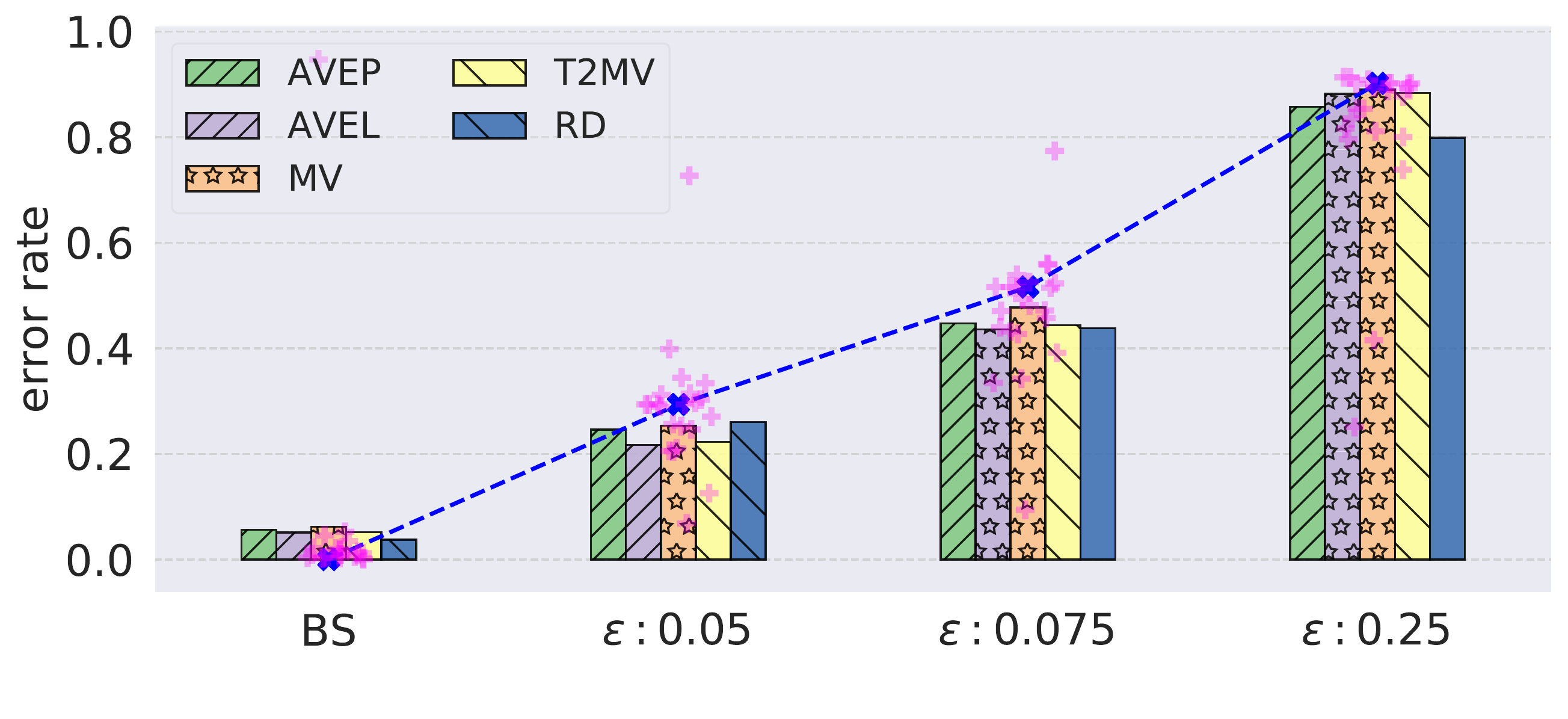}
    }
    \caption{Evaluation results for \ourframework~using linear SVM as weak defenses on MNIST.}
    \label{fig:eval_zk_svm}
\end{figure}

\begin{figure}[t]
    \footnotesize
    \centering
    \includegraphics[width=1.0\linewidth]{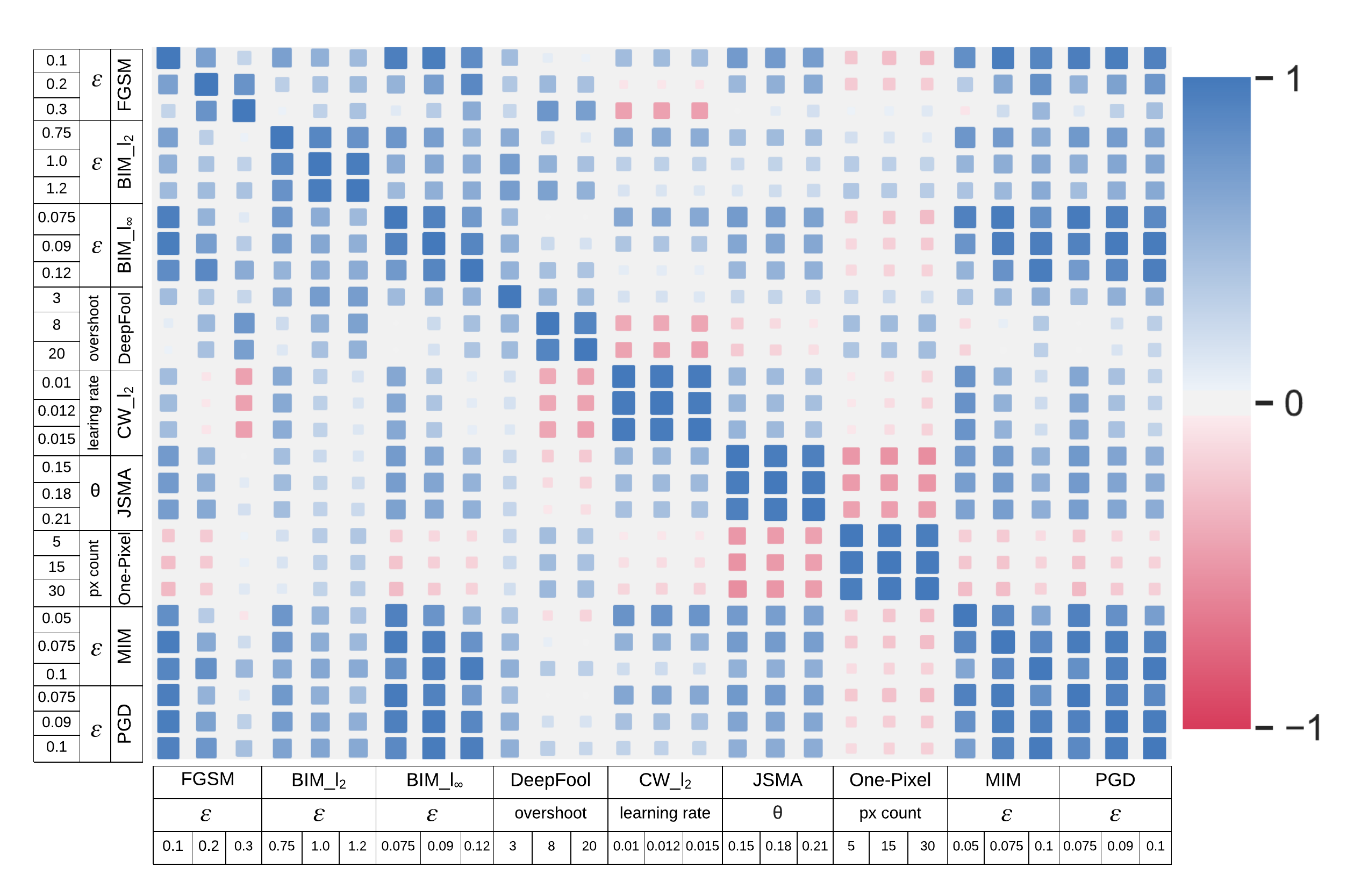}
    \caption{Rank correlation between the accuracy of WDs (CNN architectures) in different AEs crafted by from MNIST.}
    \label{fig:corr_trans_model} 
\end{figure}

\begin{figure}[h]
    \tiny
    \centering
    \includegraphics[width=0.76\linewidth]{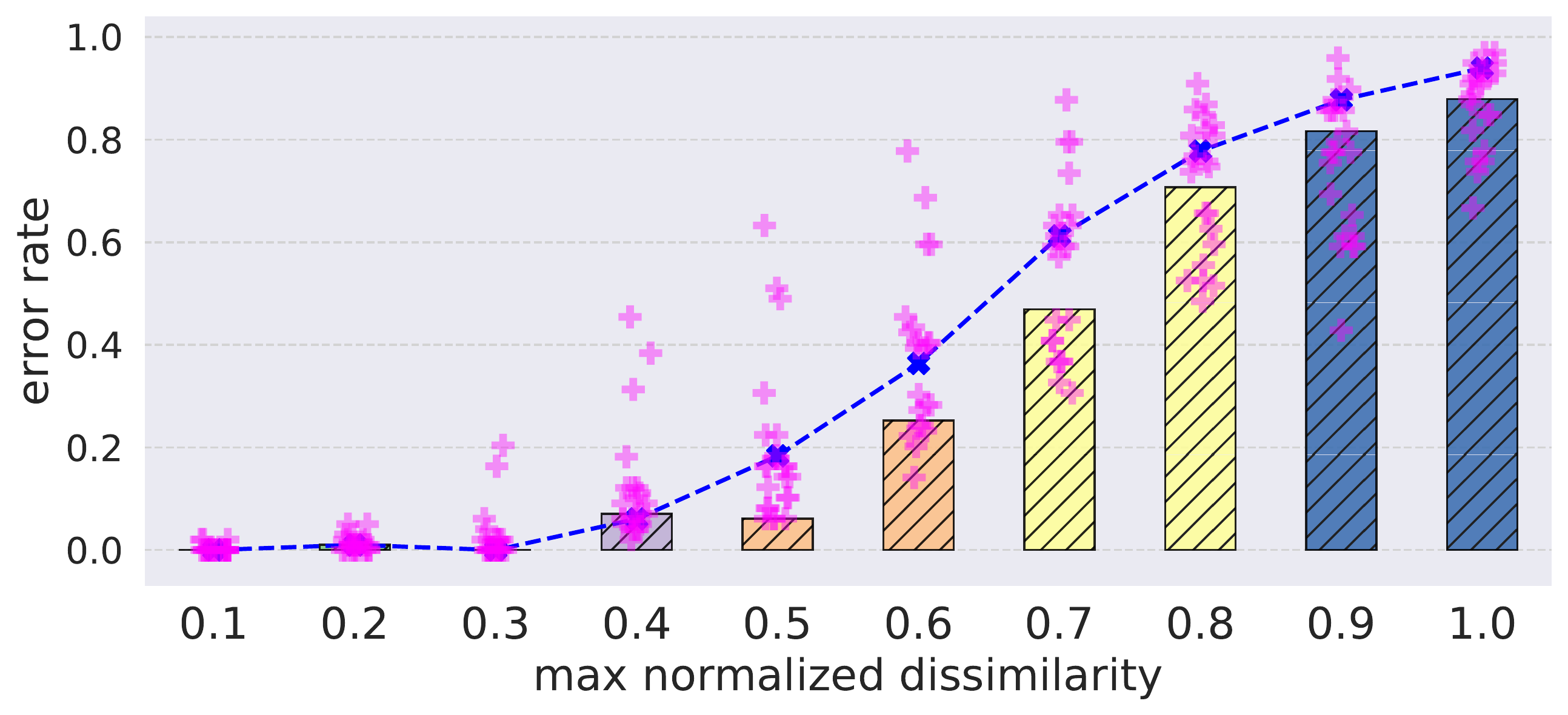}
    \caption{Evaluation of UM and MV ensemble against the greedy attack for MNIST. 10 AE variants were generated using \FGSM. From left to right, the AEs were generated with a larger constraint of max dissimilarity with respect to the corresponding benign samples.}
    \label{fig:errrate_wb_mnist}
\end{figure}

\begin{figure}
    \centering
    \includegraphics[width=0.9\linewidth]{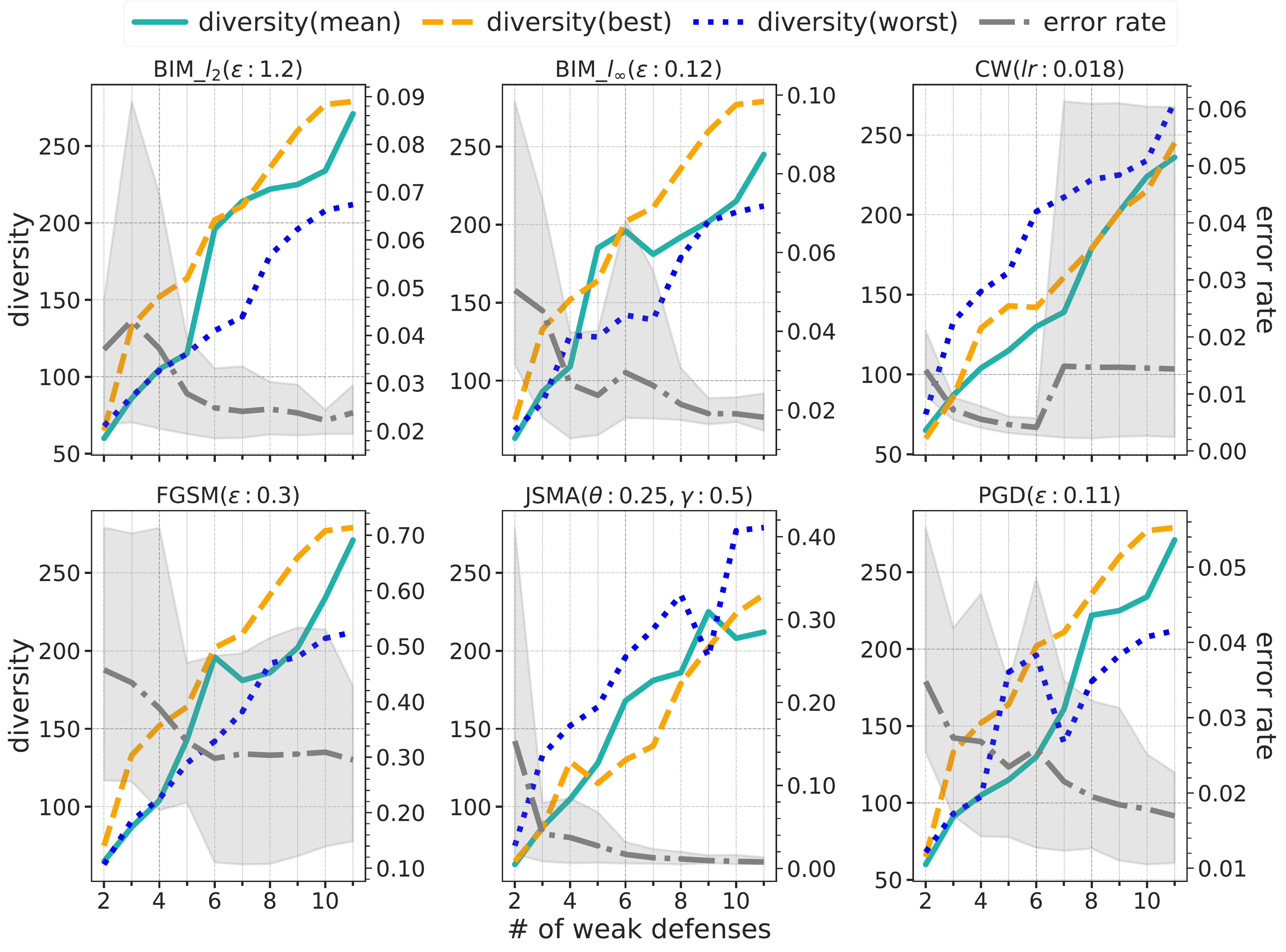}
    \caption{Ensemble diversities against various attacks on MNIST.}
    \label{fig:ensemble_diversity_mnist}
\end{figure}

\begin{figure}
    \footnotesize
    \centering
    \includegraphics[width=0.85\linewidth]{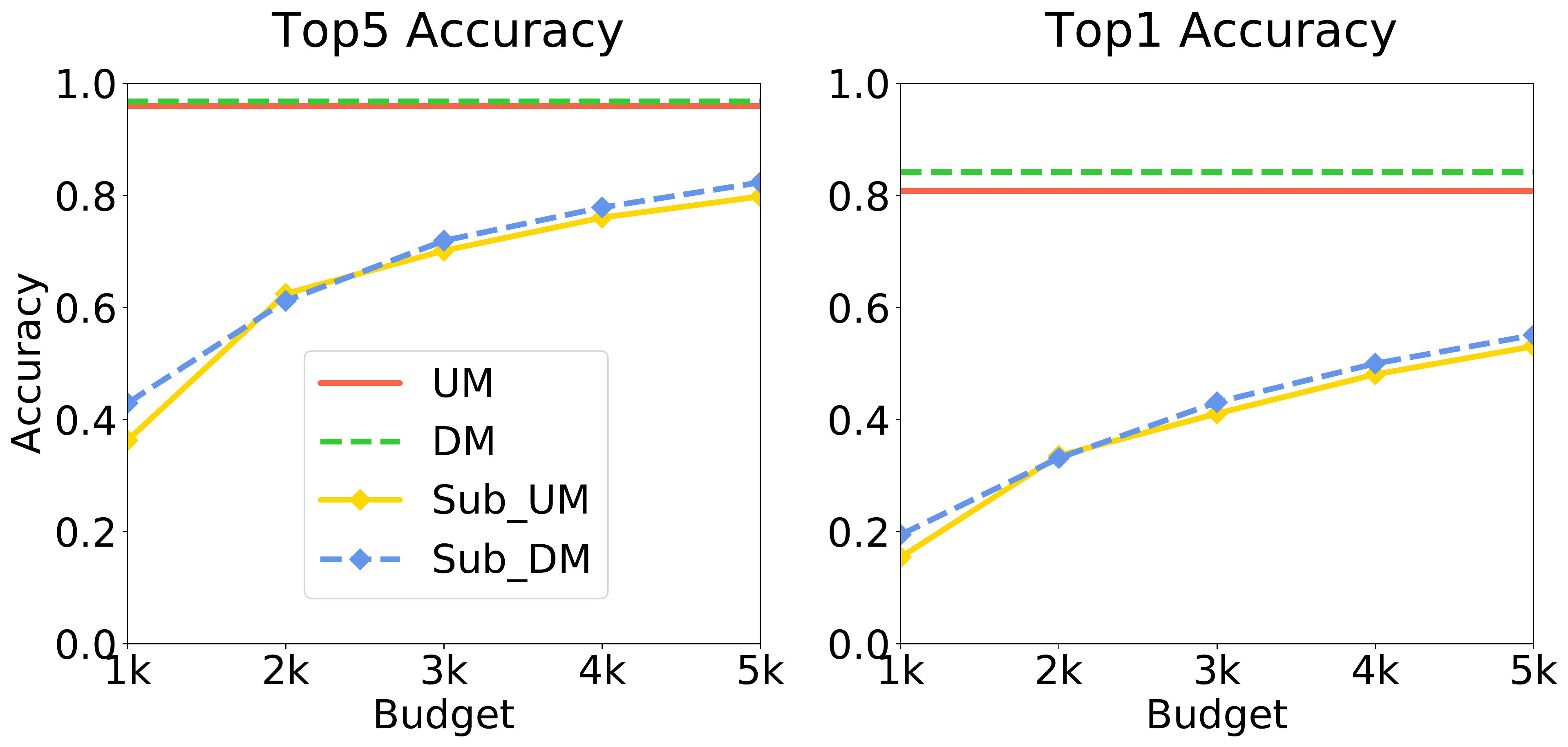}
    \caption{Performance of surrogate model trained with different budgets. Note: UM and DM stand for the undefended model and the defended model with \ourframework~respectively. 2) Sub\_UM and Sub\_DM stand for substitute models of UM and DM respectively.}
    \label{fig:perf_surrogates} 
\end{figure}

\begin{figure*}
    \footnotesize
    \centering
    \includegraphics[width=0.85\linewidth]{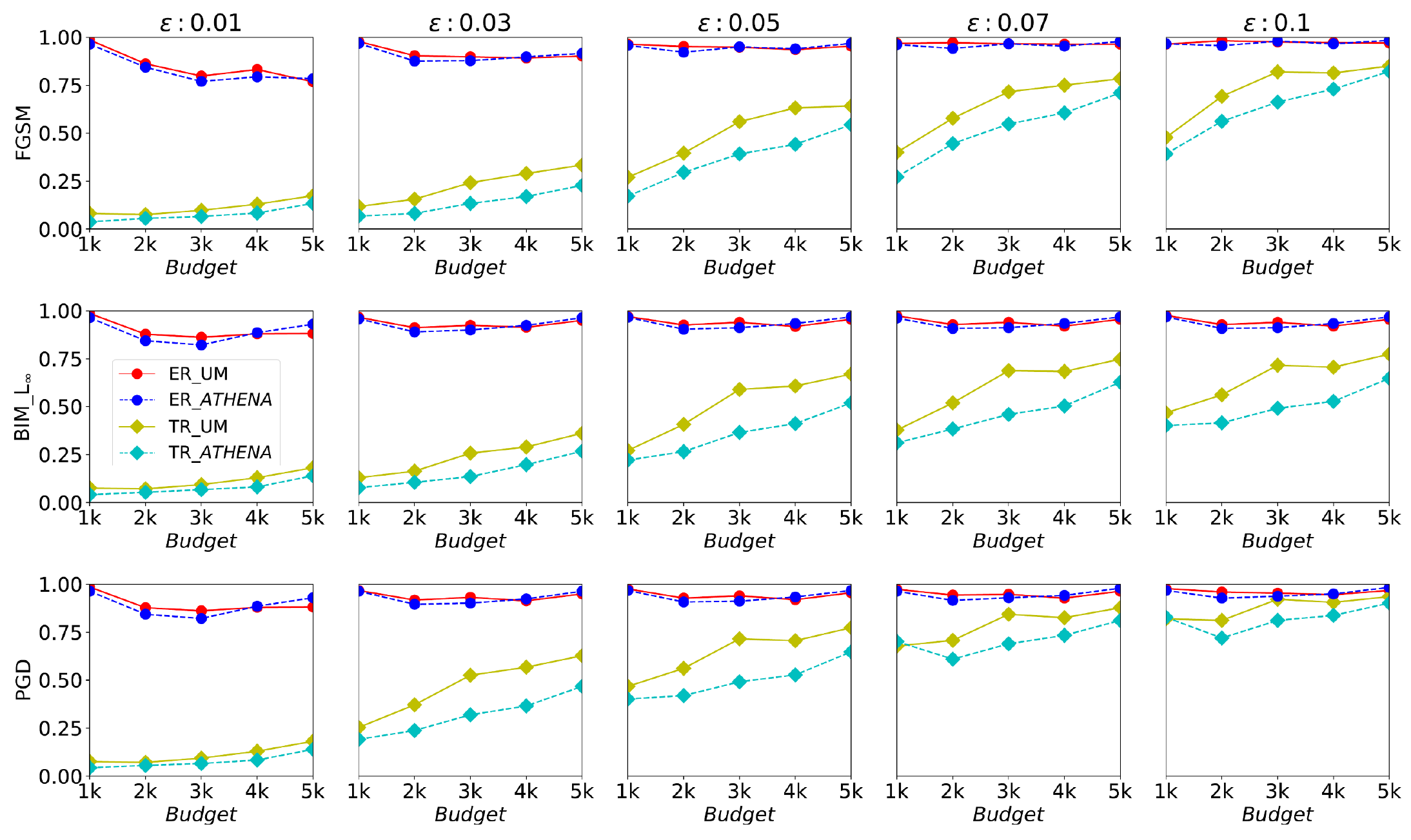}
    \caption{Performance of transfer-based black-box attack with different budgets. Note: (1) ER, TR and UM stands for error rate, transferability rate and Undefended Model respectively; (2) \ourframework~here is realized by AVEP ensemble strategy.}
    \label{fig:perf_transfer_bb_attack}
\end{figure*}

\begin{figure*}[t]
    \scriptsize
    \centering
    \subfloat[][\FGSM]{
        \includegraphics[width=0.32\linewidth, trim={0.8cm 0.7cm 0.8cm 1.5cm}]{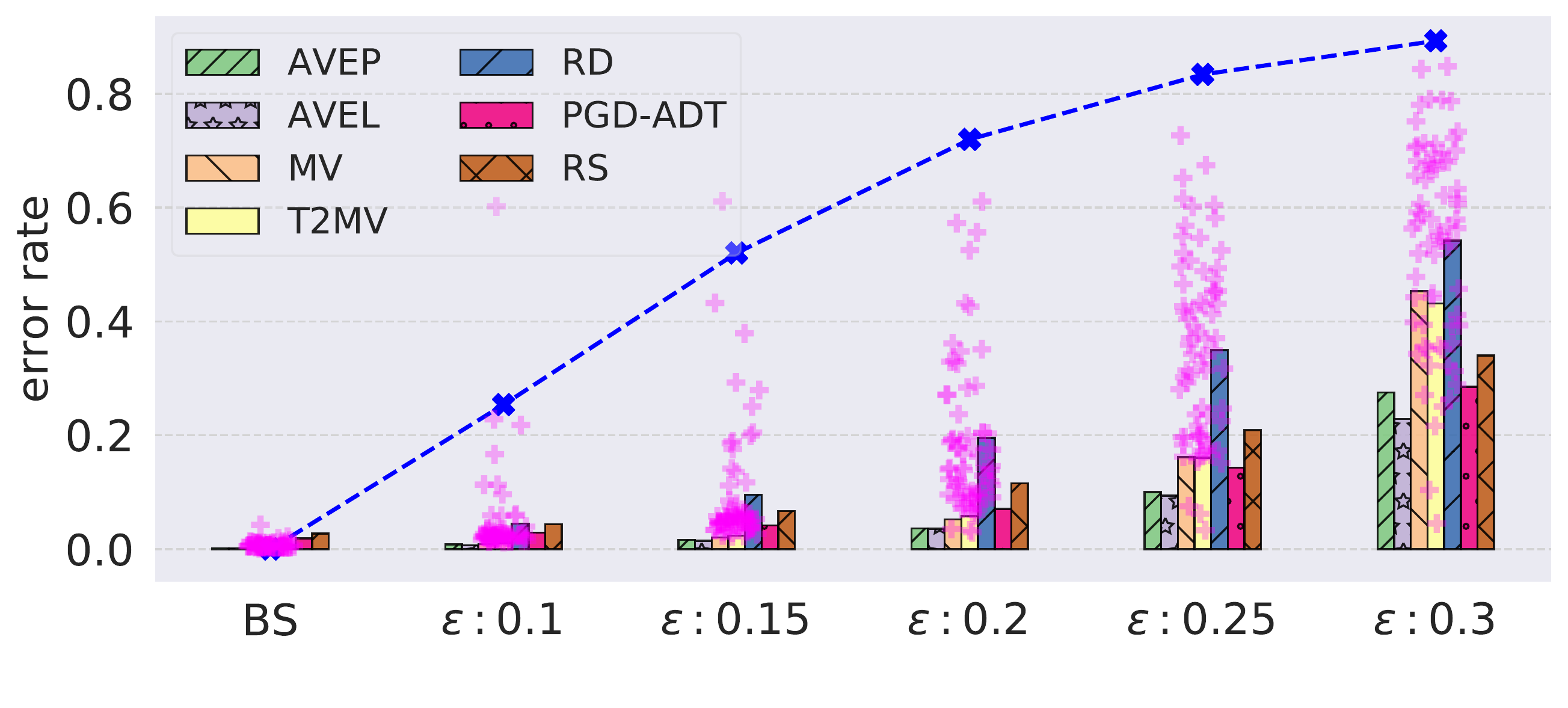}
    }
    \subfloat[][\BIM\_$l_2$]{
        \includegraphics[width=0.32\linewidth, trim={0.8cm 0.7cm 0.8cm 1.5cm}]{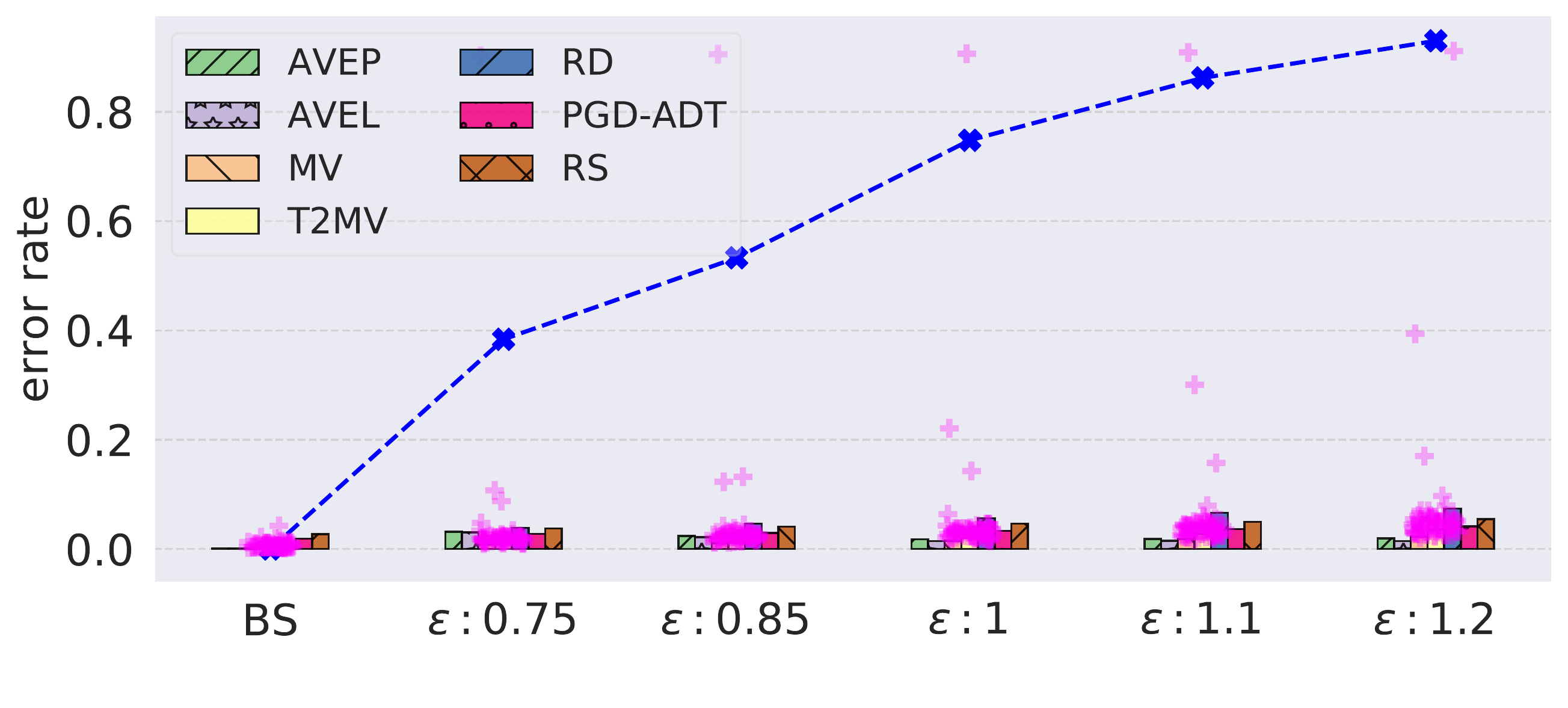}
    }
    \subfloat[][\BIM\_$l_{\infty}$]{
        \includegraphics[width=0.32\linewidth, trim={0.8cm 0.7cm 0.8cm 1.5cm}]{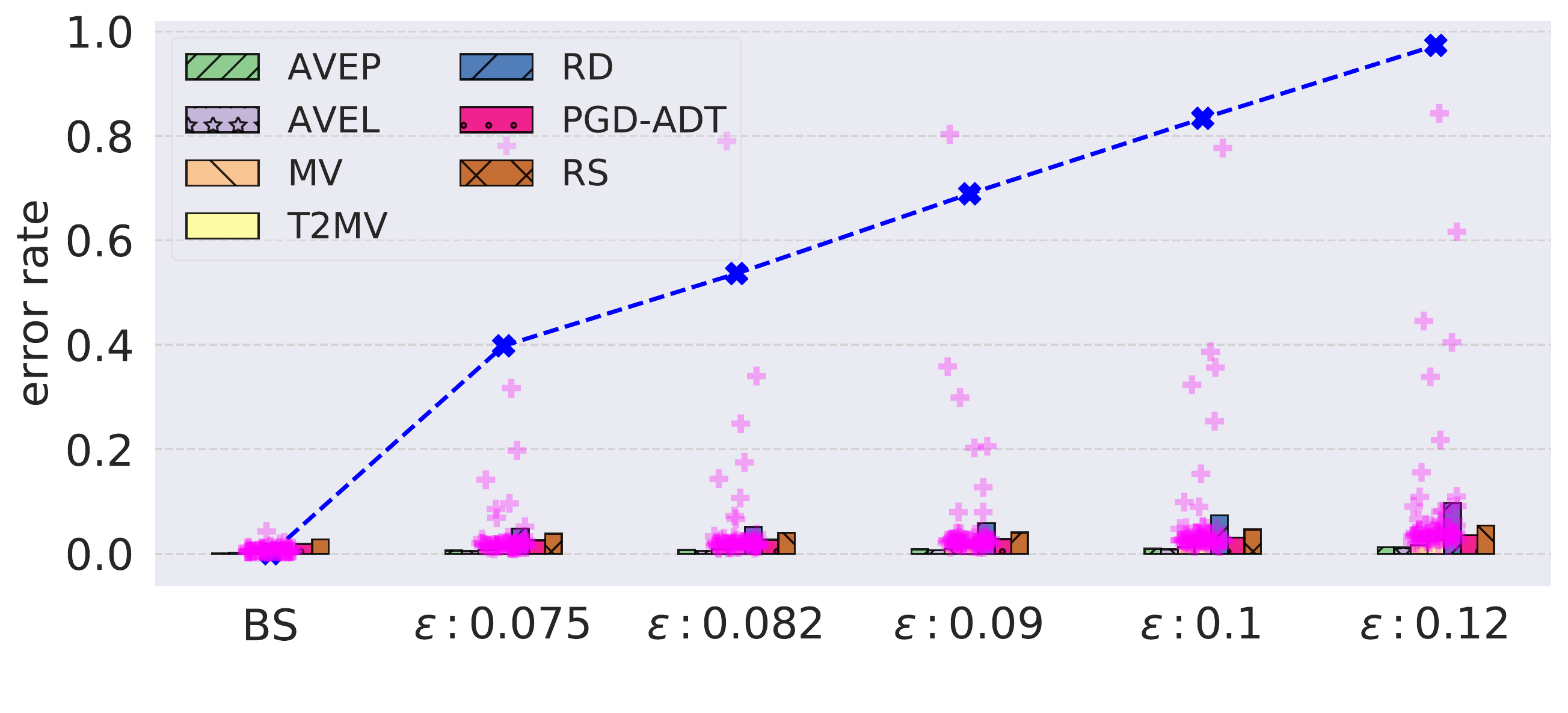}
    } \\
    \subfloat[][\CW\_$l_2$]{
        \includegraphics[width=0.32\linewidth, trim={0.8cm 0.7cm 0.8cm 1.5cm}]{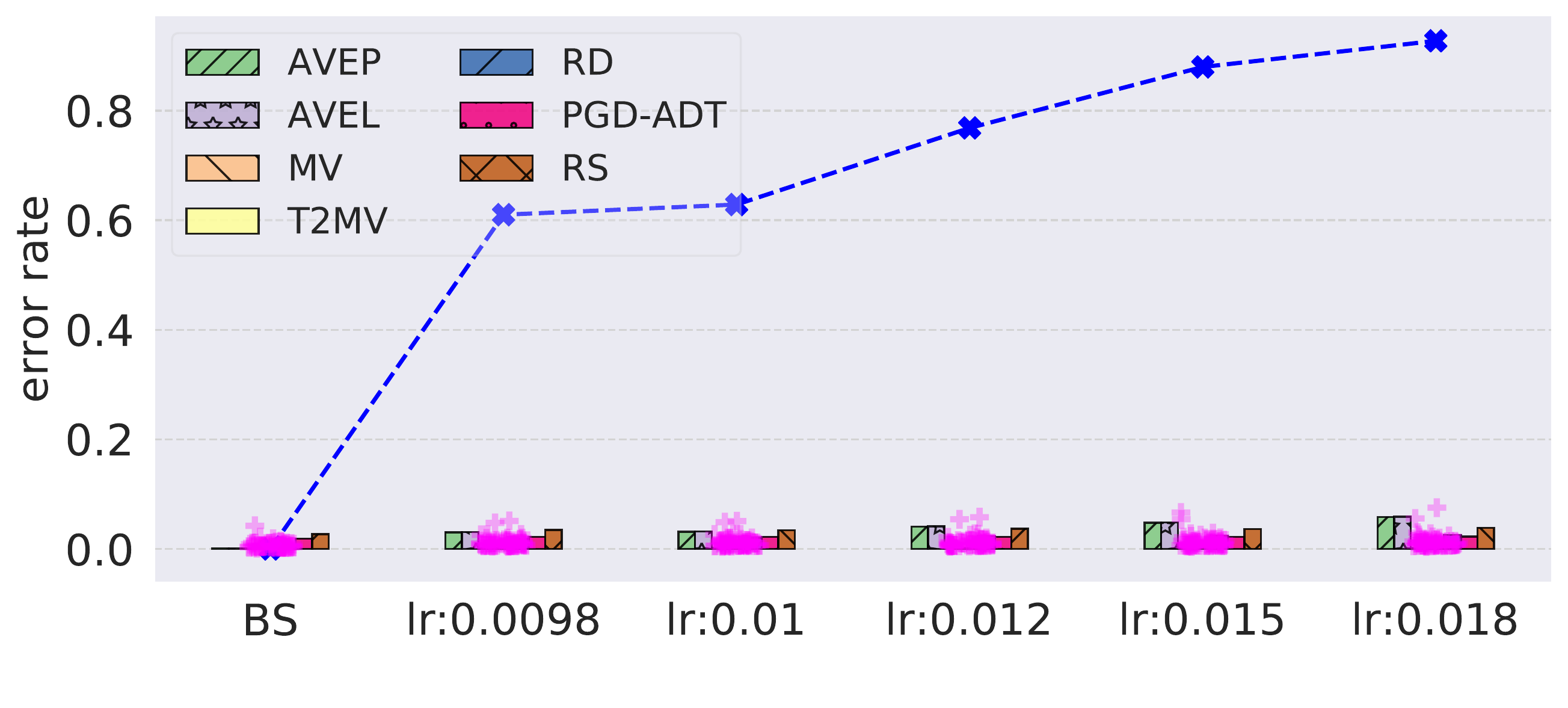}
    }
    \subfloat[][\DF]{
        \includegraphics[width=0.32\linewidth, trim={0.8cm 0.7cm 0.8cm 1.5cm}]{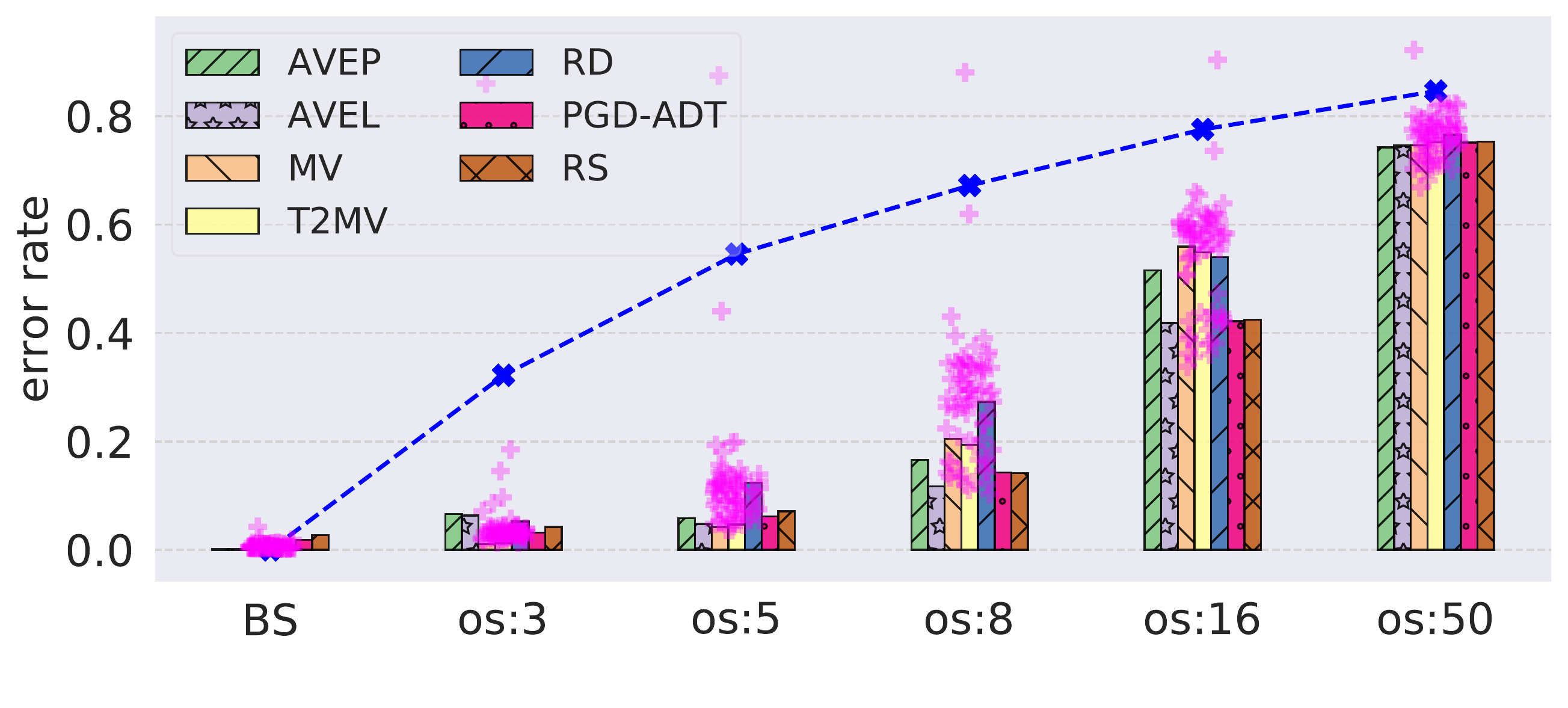}
    }
    \subfloat[][\JSMA]{
        \includegraphics[width=0.32\linewidth, trim={0.8cm 0.7cm 0.8cm 1.5cm}]{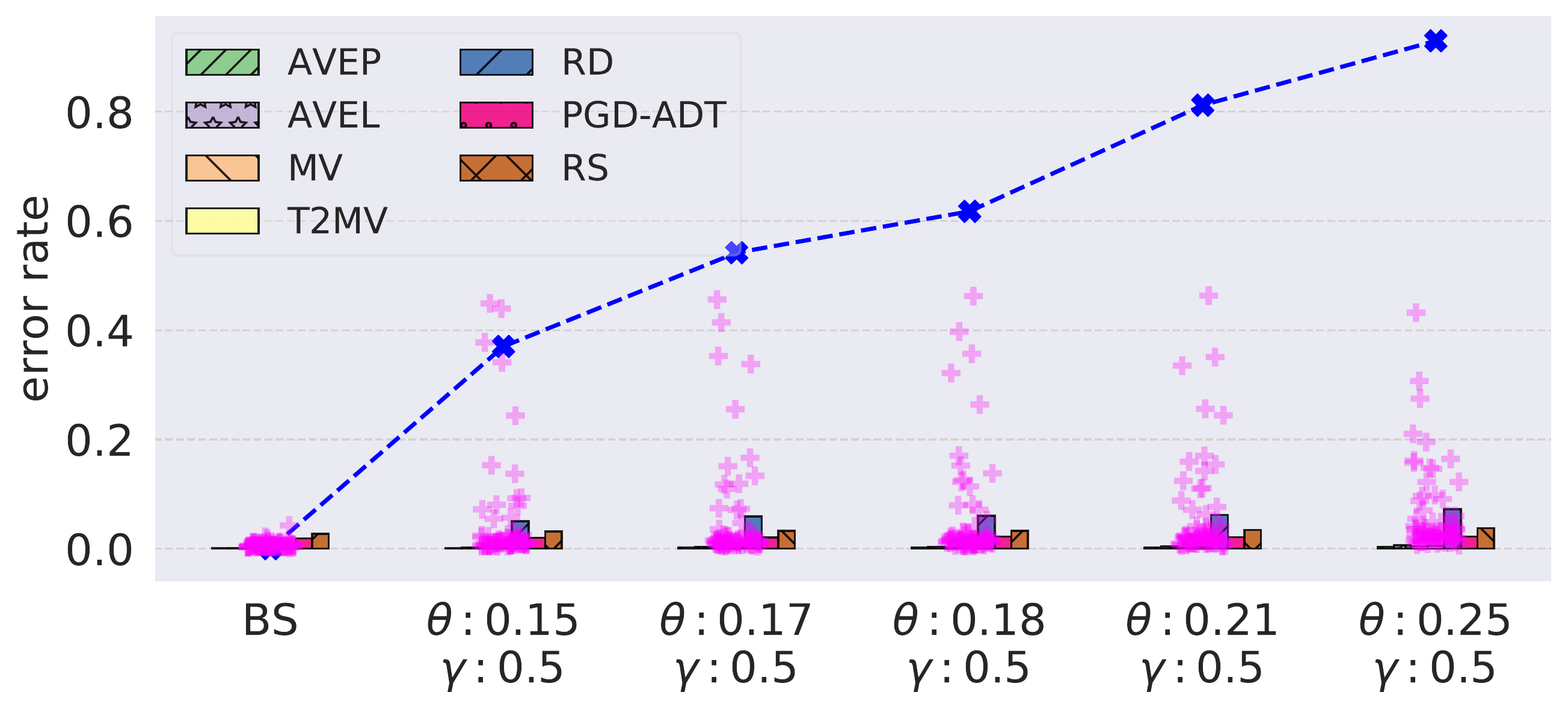}
    } \\
    \subfloat[][\OP]{
        \includegraphics[width=0.32\linewidth, trim={0.8cm 0.7cm 0.8cm 1.5cm}]{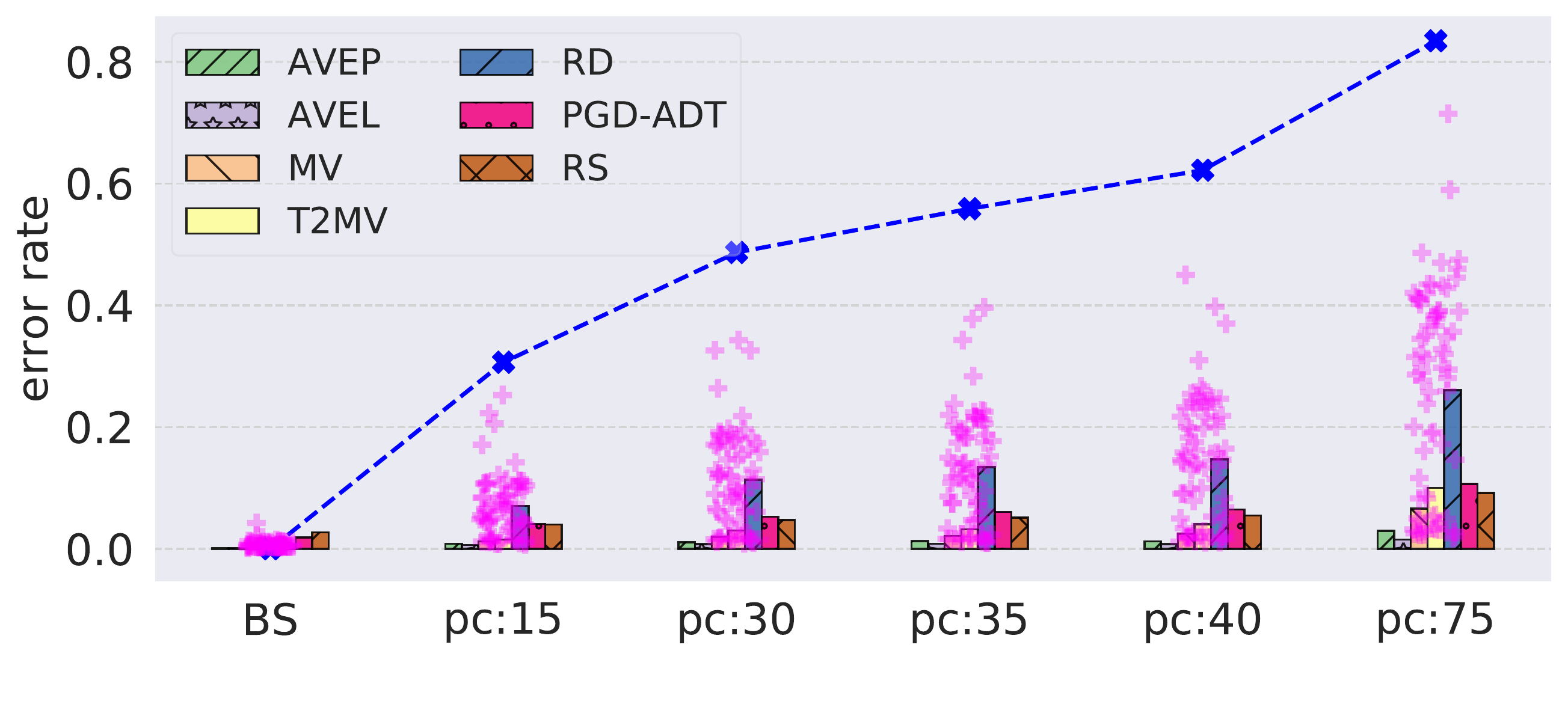}
    }
    \subfloat[][\MIM]{
        \includegraphics[width=0.32\linewidth, trim={0.8cm 0.7cm 0.8cm 1.5cm}]{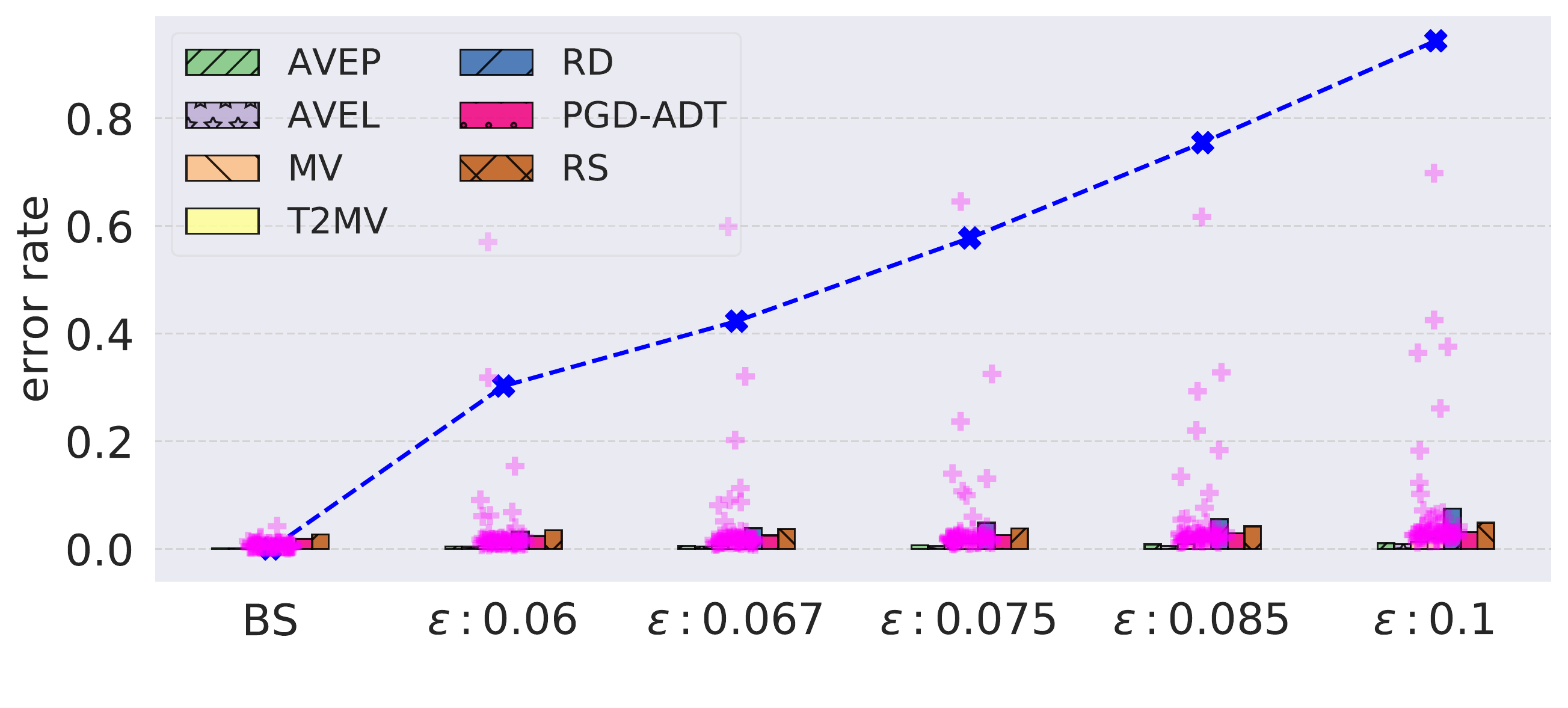}
    }
    \subfloat[][\PGD]{
        \includegraphics[width=0.32\linewidth, trim={0.8cm 0.7cm 0.8cm 1.5cm}]{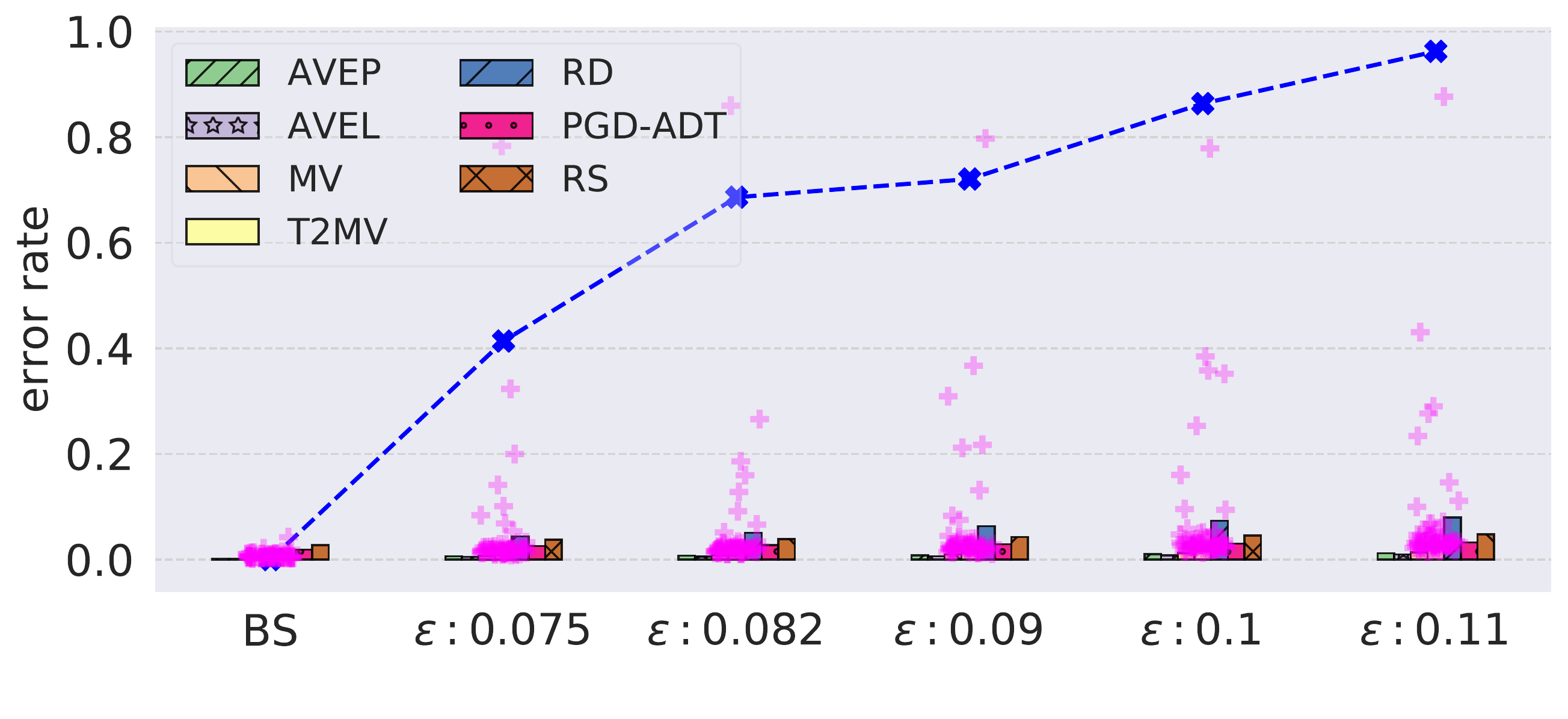}
    }
    \caption{Evaluation results for \ourframework~using CNN as weak defenses on MNIST.} 
    \label{fig:eval_zk_mnist-cnn}
\end{figure*}

\end{document}